\documentclass[twoside]{article}
\usepackage[accepted]{aistats2018}
\usepackage[]{amsmath,amssymb,epsfig}
\usepackage{amsthm}
\usepackage{amsmath}
\usepackage{amssymb}
\usepackage{graphicx}
\usepackage{epstopdf}
\usepackage{comment}
\usepackage{array}
\usepackage{algorithm}
\usepackage{url}

\usepackage{lscape}
\usepackage{algpseudocode}
\usepackage{setspace}
\usepackage{multicol}
\usepackage{multirow}
\usepackage{color}
\usepackage{colortbl}
\usepackage{xcolor}
\usepackage{hyperref}

\usepackage[%
    font={small,sf},
    labelfont=bf,
    format=hang,    
    format=plain,
    margin=0pt,
    width=0.8\textwidth,
]{caption}
\usepackage[list=true]{subcaption}

\newtheorem{theorem}{Theorem}

\newtheorem{corollary}{Corollary}

\newtheorem{lemma}{Lemma}

\newtheorem{remark}{Remark}

\numberwithin{equation}{section}

\newenvironment{rcases}
  {\left.\begin{aligned}}
  {\end{aligned}\right\rbrace}

\newcommand{\calC}{\ensuremath{\mathcal{C}}}

\newcommand{\calS}{\ensuremath{\mathcal{S}}}

\newcommand{\calN}{\ensuremath{\mathcal{N}}}

\newcommand{\norm}[1]{\parallel{#1}\parallel}
\newcommand{\abs}[1]{|{#1}|}

\newcommand{\set}[1]{\left\{{#1}\right\}}
\newcommand{\dotprod}[2]{\langle#1,#2\rangle}
\newcommand{\est}[1]{\widehat{#1}}

\newcommand{\matR}{\ensuremath{\mathbb{R}}}

\newcommand{\argmin}[1]{\underset{#1}{\operatorname{argmin}}}

\newcommand{\vecx}{\mathbf x}

\newcommand{\vecq}{\mathbf q}

\newcommand{\vecv}{\mathbf v}

\newcommand{\vecb}{\mathbf b}

\newcommand{\vecz}{\mathbf z}

\newcommand{\vecg}{\mathbf g}

\newcommand{\matA}{\ensuremath{\mathbf{A}}}

\newcommand{\matI}{\ensuremath{I}} 

\newcommand{\matP}{\ensuremath{\mathbf{P}}}

\newcommand{\real}{\text{Re}}
\newcommand{\imag}{\text{Im}}
\newcommand{\vecgbar}{\bar{\mathbf g}}
\newcommand{\vecgbarest}{\est{\vecgbar}}
\newcommand{\Hbar}{H}
\newcommand{\veczbar}{\bar{\mathbf z}}

\newcommand{\vecgest}{\est{\vecg}}
\newcommand{\gest}{\est{g}}
\newcommand{\qtathresh}{\zeta}
\newcommand{\sign}{\text{sign}}

\newcommand{\htil}{h}

\newcommand{\vechtil}{\mathbf{h}}
\newcommand{\vechtilbar}{\bar{\mathbf{h}}}

\newcommand{\trseq}{\ensuremath{TSR_{=}}}
\newcommand{\trsineq}{\ensuremath{TSR_{\leq}}}
\newcommand{\mb}[1]{\mbox{\boldmath$#1$}} 
  
\usepackage{rotating}

\usepackage{caption}

 \usepackage{float}

\begin{document}

%

%

\twocolumn[
\aistatstitle{On denoising modulo $1$ samples of a function}
\aistatsauthor{Mihai Cucuringu \And Hemant Tyagi }
\aistatsaddress{Alan Turing Institute, London  \\ University of Oxford\\ mihai.cucuringu@stats.ox.ac.uk \And 
Alan Turing Institute, London \\ University of Edinburgh \\ htyagi@turing.ac.uk}]


\begin{abstract}

Consider an unknown smooth function  \hspace{5mm} $f: [0,1] \rightarrow \matR$, and say we are given 
$n$ noisy$\mod 1$ samples of $f$, i.e., $y_i = (f(x_i) + \eta_i)\mod 1$ for  $x_i \in [0,1]$,  
where $\eta_i$ denotes noise. Given the samples $(x_i,y_i)_{i=1}^{n}$ our goal is to recover smooth, robust estimates of the clean 
samples $f(x_i) \bmod 1$. We formulate a natural approach for solving this problem which works with representations of
mod  1 values over the unit circle. This amounts to solving a quadratically constrained quadratic program (QCQP) 
with non-convex constraints involving points lying on the unit circle.
Our proposed approach is based on solving its relaxation which is a \emph{trust region subproblem}, and hence solvable efficiently.
We demonstrate its robustness to noise  
via extensive simulations on several synthetic examples, and provide a detailed theoretical 
analysis. 
\end{abstract}

 \vspace{-3mm}
\section{Introduction} \label{sec:Intro}
\vspace{-1mm}

The problem of recovering a function $f$ from noisy samples of its $\mod 1$ values has received recent 
interest both in the literature and the media (MIT News \cite{MIT_News_link}).
%
%
This recent surge of interest was motivated by a new family of analog-to-digital converters (ADCs). Traditional ADCs have voltage limits in place that cut off the signal at the maximum allowed voltage, whenever it exceeds the limit.  In very recent work, the authors of \cite{bhandari17} introduced a technique, denoted as \textit{unlimited sampling}  that is able to accurately digitize signals whose voltage peaks are much larger than the voltage limits of an ADC.  Their work was inspired by a new type of experimental ADC 
with a \textit{modulo architecture} (the so-called \textit{self-reset ADC}, that has already been prototyped) 
which captures not the voltage of a signal but its modulo, by having the voltage reset itself whenever 
it crosses a pre-specified threshold. In other words, the ADC captures the remainder obtained when the voltage of an analog signal is divided by the maximum voltage of the ADC.

\vspace{-1mm}
The multi-dimensional version of this problem has a long history in the geosciences literature, often dubbed as the \textit{phase unwrapping} problem. Phase unwrapping refers to the process of recovering unambiguous phase values from phase data that are measured modulo $2 \pi$ rad (wrapped data). Instances of this problem arise in many applications, with an initial spike of interest in early 1990s spurred by the  synthetic aperture radar interferometry (InSAR) technology for determining the surface topography and deformation of the Earth, which motivated the development of two-dimensional phase unwrapping algorithms. 
Most of the commonly used phase unwrapping algorithms relate the phase values by first differentiating the
phase field and subsequently reintegrating, adding back the missing integral cycles with the end goal of obtaining a more continuous result \cite{Lu98phaseunwrapping}. 
Other approaches explored in the literature include combinations of least-squares techniques \cite{pritt1996}, 
methods exploiting measures of data integrity to guide the unwrapping process \cite{Bone_91}, and several techniques 
employing neural network or genetic algorithms \cite{Collaro_98}. 
The three-dimensional version of the problem \cite{Hooper_2007_PhaseUnwrap3D} has received relatively little attention, 
 a recent line of work in this direction being \cite{Osmanoglu_3D_2014}.




\vspace{-1mm}
As a word of caution, 
 note that this problem is different from \textit{phase retrieval}, a classical problem in optics that has attracted a surge of interest in recent years \cite{phase_retrieval_survery_candes, phase_retrieval_survery}, which attempts to recover an unknown signal from the magnitude 
of its Fourier transform.  Just like phase retrieval, the recovery of a function from mod 1 measurements is, by its very nature, an ill-posed problem, and one needs to incorporate prior structure on the signal, which in our case is smoothness of $f$ (
analogously to how enforcing sparsity renders the phase retrieval problem well-posed). 


\vspace{-1mm}
At a high level, one would like to recover denoised samples (i.e., smooth, robust estimates) of $f$ from 
its noisy mod  1 versions. A natural mode of attack for this problem is the following two-stage approach. 
In the first stage, one recovers denoised  mod 1 samples of $f$, and then in the (unwrapping) second stage, 
one uses these samples to recover the original real-valued samples of $f$. 
In this paper, we mainly focus on the first stage, which is a challenging problem in itself.  To the best of our knowledge, we provide the first algorithm for denoising mod $1$ samples of a function, which comes with  robustness guarantees. In particular, we make the following contributions.

\vspace{-3mm}
\begin{enumerate} 
\item We formulate a general framework for denoising the mod 1  samples of $f$; it involves 
mapping the noisy  mod 1 values (in $[0,1)$) to the angular domain (i.e. in $[0,2\pi)$), 
and leads to a QCQP formulation with non-convex constraints. We consider solving a relaxation of this QCQP which is a trust region 
subproblem, and hence solvable efficiently.

\vspace{-1mm}
\item We provide a detailed theoretical analysis for the above approach, which demonstrates its robustness 
to noise for the arbitrary bounded noise model (see \eqref{eq:arb_bd_noise_model},\eqref{eq:BoundedNoiseModel}).

\vspace{-1mm}
\item We test the above method on several synthetic examples which demonstrate that it performs well for reasonably high noise levels.
To complete the picture, we also implement the second stage with a simple recovery method for recovering the (real valued) 
samples of $f$, and show that it performs surprisingly well via extensive simulations.
\end{enumerate}

\vspace{-6mm}
\paragraph{Outline of paper.} Section \ref{sec:ProblemSetup} formulates the problem formally, and introduces notation.
Section \ref{sec:AngularLS} sets up the  mod 1 denoising problem as a smoothness regularized least-squares problem 
in the angular domain; this is a QCQP with non-convex constraints.
Section \ref{sec:trust_reg_relax} describes 
its relaxation to a \textit{trust-region subproblem}, and some possible approaches for recovering the samples of $f$, along with our complete two-stage algorithm. 
Section \ref{sec:bound_noise_analysis} contains  approximation guarantees for our algorithm for recovering the  denoised   mod 1 samples of $f$. 
Section \ref{sec:num_exps} contains numerical experiments on different synthetic examples. 
Finally, 
Section \ref{sec:conclusion} summarizes our results and contains a discussion of  possible future research directions.  
\vspace{-2mm}


\vspace{-2mm}

\section{Problem setup} \label{sec:ProblemSetup}
\vspace{-3mm}
Consider a smooth, unknown function $f : [0,1] \rightarrow \mathbb{R}$, and a uniform grid on $[0,1]$, 
\vspace{-3mm}
\begin{equation}
0 = x_1 < x_2 < \cdots < x_n = 1 \ \text{with} \ x_i = \frac{i-1}{n-1}.
\end{equation}
\vspace{-6mm}

We assume that we are given \emph{mod 1} samples of $f$ on the above grid. Note that for each  sample 

\vspace{-5mm}
\begin{equation}   \label{eq:fmod1}
 f(x_i) = q_i + r_i \in \mathbb{R}, 
\end{equation}
\vspace{-6mm}

with $q_i \in \mathbb{Z}$ and $r_i \in [0,1)$, we have 
$r_i = f(x_i) \bmod 1$. The modulus is fixed to 1 without loss of generality since 
$\frac{f \bmod s}{s} = \frac{f}{s} \bmod 1$. This is easily seen by writing $ f = sq + r $, with $q \in \mathbb{Z}$, 
and observing that $ \frac{f}{s} \bmod 1 =  \frac{sq + r}{s} \bmod 1 = \frac{r}{s} = \frac{f \bmod  s}{s}$.
In particular, we assume that the mod 1 samples are \emph{noisy}, and consider the following noise models. 

\vspace{-3mm}
\begin{enumerate}
 \setlength\itemsep{0em}
\item \textbf{Arbitrary bounded noise}

\vspace{-6mm}
\begin{equation} \label{eq:arb_bd_noise_model}
y_i = (f(x_i) + \delta_i) \bmod 1; \ \abs{\delta_i} \in (0,1/2), \ \forall i.
\end{equation}
\vspace{-5mm}
%
\item \textbf{Gaussian noise}

\vspace{-6mm}
\begin{equation} \label{eq:gauss_noise_model}
y_i = (f(x_i) + \eta_i) \mod 1;   \forall i  
\end{equation}
\vspace{-2mm}
where $\eta_i \sim \mathcal{N}(0,\sigma^2)$ i.i.d.
\end{enumerate}
\vspace{-2mm}

We will denote $f(x_i)$ by $f_i$ for convenience. Our aim is to recover smooth, robust estimates (up to a global shift) 
of the original samples $(f_i)_{i=1}^{n}$ from the measurements $(x_i, y_i)_{i=1}^{n}$. 
We will assume $f$ to be H\"older continuous, meaning that for constants $M > 0$, $\alpha \in (0,1]$, 

\vspace{-6mm}
\begin{equation} \label{eq:f_smooth_hold}
\abs{f(x) - f(y)} \leq M \abs{x - y}^{\alpha}; \quad \forall \ x,y \in [0,1].
\end{equation}  
\vspace{-6mm}

The above assumption is quite general and reduces to Lipschitz continuity when $\alpha = 1$. 
%

\vspace{-2mm}
\paragraph{Notation.} Scalars and matrices are denoted by lower case and upper cases symbols respectively, 
while vectors are denoted by lower bold face symbols. Sets are denoted by calligraphic symbols (eg., $\calN$), 
with the exception of $[n] = \set{1,\dots,n}$ for $n \in \mathbb{N}$. The imaginary unit is denoted by $\iota = \sqrt{-1}$. 
\vspace{-3mm}


\section{Smoothness regularized least-squares in the angular domain}  \label{sec:AngularLS}
\vspace{-3mm}

Our algorithm essentially works in two stages.

\vspace{-4mm}
\begin{enumerate}
\item \textbf{Denoising stage}. Our goal here is to denoise the mod 1 samples, 
which is also the main focus of this paper. In a nutshell, we map the given noisy mod 1 samples 
to  points on the unit complex circle, 
and solve a smoothness regularized, constrained least-squares problem. 
The solution to this problem, followed by a simple post-processing step, gives us  denoised mod 1 samples  of $f$.

\item \textbf{Unwrapping stage}. The second stage takes as input the above denoised  mod 1 samples, and recovers an estimate 
to the original real-valued samples of $f$ (up to a global shift).
\end{enumerate}

\vspace{-4mm}
We start the denoising stage by mapping the  mod 1 samples to the angular domain as follows. Let

\vspace{-7mm}
\begin{equation} \label{eq:unit_complex_circ_rep}
\hspace{-3mm} \htil_i := \exp(2 \pi \iota f_i) = \exp(2 \pi \iota r_i), \; z_i  := \exp(2 \pi \iota y_i)
\end{equation}
\vspace{-7mm}

be the respective representations of the clean and noisy mod 1 samples on the unit 
circle in $\mathbb{C}$, where the first equality is due to the fact that $f_i = q_i + r_i$, with $q_i \in \mathbb{Z}$.    
\begin{figure}
\subcaptionbox[Short Subcaption]{ Clean $ f $ mod 1 
}[ 0.22\textwidth ]
{\includegraphics[width=0.22\textwidth] {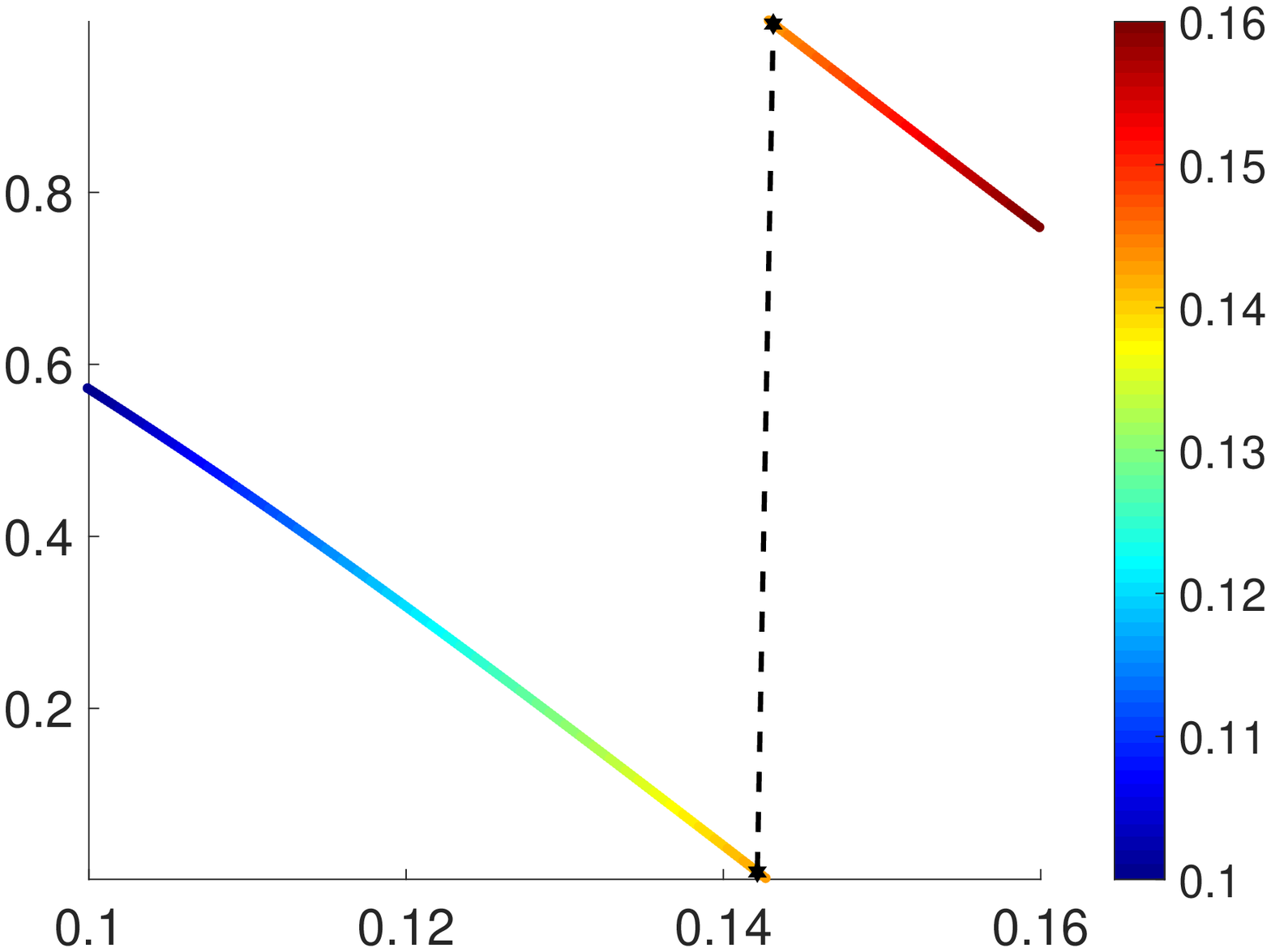} }
\subcaptionbox[Short Subcaption]{ Angular embedding 
}[ 0.19\textwidth ]
{\includegraphics[width=0.19\textwidth] {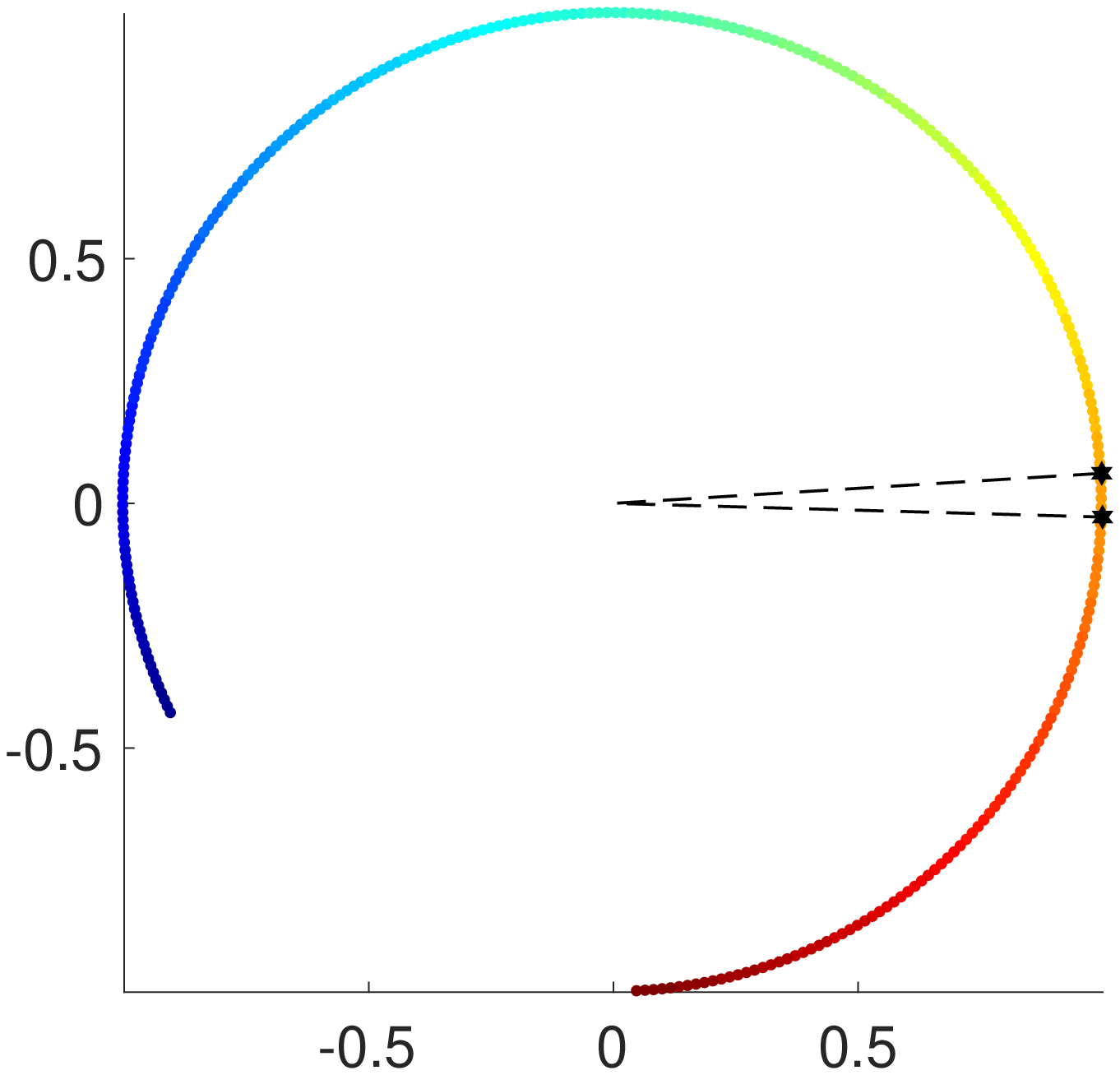} }
%
\vspace{-1mm} 
\caption[Short Caption]{Motivation for the angular embedding approach.}
\label{fig:toyExampleFarNear}
\end{figure}
\vspace{-2mm} 

The choice of representing the mod 1 samples in \eqref{eq:unit_complex_circ_rep} is very natural for the following reason. 
For points $x_i,x_j$ sufficiently close, the samples $f_i, f_j$ will also be close (by H\"older continuity of $f$). 
While the corresponding wrapped samples $f_i \mod 1, f_j \mod 1$ can still be far apart,  the 
complex numbers $\exp(\iota 2\pi f_i)$ and $\exp(\iota 2\pi f_j)$ will necessarily be close 
to each other\footnote{Indeed, $\abs{\exp(\iota 2\pi f_i) - \exp(\iota 2\pi f_j)}$ $= \abs{1-\exp(\iota 2\pi (f_j - f_i))}$ 
$= 2\abs{\sin (\pi (f_j - f_i))} \leq 2\pi\abs{f_j - f_i}$ (since $\abs{\sin x} \leq \abs{x}$ $\forall x \in \matR$).}.
This is illustrated in the toy example in Figure \ref{fig:toyExampleFarNear}.  

Consider the graph $G = (V,E)$ with $V = \{1,2,\ldots,n\}$ where index $i$ corresponds to the point $x_i$ on our grid, 
and $E = \{(i,j) \in {[n] \choose 2} : |i-j| \leq k\}$ denotes the set of edges for a suitable parameter $k \in \mathbb{N}$. 
A natural approach for recovering smooth estimates of $(h_i)_{i=1}^n$ would be to solve the following optimization problem.

\vspace{-8mm} 
\begin{eqnarray}
\min_{ g_1,\ldots,g_n \in \mathbb{C}; |g_i|=1 } \sum_{i=1}^{n} |g_i - z_i|^2 + \lambda  \sum_{(i,j) \in E} |g_i - g_{j}|^2. 
\label{eq:orig_denoise_hard}
\end{eqnarray}
\vspace{-6mm} 

Here, $\lambda > 0$ is a regularization parameter, which along with $k$, controls the smoothness of the solution. 
Let us denote $L \in \mathbb{R}^{n \times n}$ to be the Laplacian matrix associated with $G$, defined as

\vspace{-5mm} 
\begin{equation} \label{eq:laplacian_def}
L_{i,j} = \left\{
\begin{array}{rl}
\text{deg}(i) \quad ; & i = j \\
-1 \quad ; & (i,j) \in E \ \text{or} \ (j,i) \in E   \\
0  \quad ; & \text{otherwise}  
\end{array} \right.
\end{equation} 
\vspace{-4mm} 

Denoting $\vecg = [g_1 \ g_2 \ \dots \ g_n]^T \in \mathbb{C}^n$,  the second term in \eqref{eq:orig_denoise_hard} can be simplified to 

\vspace{-6mm} 
\begin{align}
\hspace{-5mm}  \lambda \left( \sum_{i \in V} \text{deg}(i) \abs{g_i}^2  -  \hspace{-2mm} \sum_{(i,j) \in E} (g_i g_j^{*} + g_i^{*} g_{j}) \right)  = \lambda \vecg^{*}  L \vecg. \label{eq:quadform_exp_compl}
\end{align}

\vspace{-5mm} 
Next, denoting $\vecz = [z_1 \ z_2 \ \dots \ z_n]^T \in \mathbb{C}^n$, we can further simplify the first 
term in \eqref{eq:orig_denoise_hard} as follows.

\vspace{-6mm} 
\begin{align}
\sum_{i=1}^{n} |g_i - z_i|^2 
&= \sum_{i=1}^{n} (\abs{g_i}^2 + \abs{z_i}^2 - g_i z^{*}_i - g^{*}_i z_i) \\
&= 2n - 2Re(\vecg^{*}\vecz).
\end{align}
\vspace{-6mm} 

This gives us the following equivalent form of \eqref{eq:orig_denoise_hard}

\vspace{-7mm} 
\begin{eqnarray} \label{eq:orig_denoise_hard_1}
\min_{ \vecg \in \mathbb{C}^n: \abs{g_i} = 1} \lambda \vecg^{*}  L \vecg - 2Re(\vecg^{*}\vecz).
\end{eqnarray}



\vspace{-3mm}

\section{A trust region based relaxation for denoising modulo 1 samples} \label{sec:trust_reg_relax} 
\vspace{-3mm}
The optimization problem in \eqref{eq:orig_denoise_hard_1} is over a non-convex set $\calC_n := \set{\vecg \in \mathbb{C}^n: \abs{g_i} = 1}$. 
In general, the problem $\min_{\vecg \in \calC_n} \vecg^{*} \matA \vecg$, where $\matA \in \mathbb{C}^{n\times n}$ is positive semidefinite, 
is NP-hard \cite[Proposition 3.3]{zhang06}. The quadratic term in \eqref{eq:orig_denoise_hard_1} involves the Laplacian of a nearest neighbour graph, 
and of course has more structure, however the precise complexity of \eqref{eq:orig_denoise_hard_1} is unclear. As pointed out by a reviewer, 
one possible approach is to discretize the angular domain, and solve \eqref{eq:orig_denoise_hard_1} approximately via dynamic programming. 
Since the graph $G$ has tree width $k$, the computational cost for this approach may be exponential in $k$. 

The approach we adopt involves relaxing the constraints in \eqref{eq:orig_denoise_hard_1} to one where the points lie 
on a sphere of radius $n$, resulting in the following optimization 
\vspace{-1mm}
\begin{eqnarray} \label{eq:qcqp_denoise_complex}
\min_{\vecg \in \mathbb{C}^n: \norm{\vecg}^2 = n} \lambda \vecg^{*} L \vecg  - 2 Re (\vecg^{*} \vecz).
\end{eqnarray}
%
%
It is straightforward to reformulate \eqref{eq:qcqp_denoise_complex} in terms of real variables. 
We do so by introducing the following notation 
for the real-valued versions of the variables $\vechtil$ (clean signal), $\vecz$ (noisy signal), and  $\vecg$ (free variable) 
%
%
\begin{eqnarray} \label{eq:real_notat}
\hspace{-2mm}
\vechtilbar = \begin{pmatrix}
    \real(\vechtil)    \\    \imag(\vechtil) 
   \end{pmatrix}, \    
\veczbar = \begin{pmatrix}
  \real(\vecz) \\ \imag(\vecz) 
 \end{pmatrix}, \ 
\vecgbar = \begin{pmatrix}
  \real(\vecg) \\ \imag(\vecg) 
 \end{pmatrix} \in \matR^{2n}, \;
\end{eqnarray}
and the corresponding block-diagonal Laplacian 
 
\vspace{-6mm}
\begin{eqnarray} \label{eq:def_H}
\Hbar = \begin{pmatrix}
  \lambda L \quad & 0 \\ 0 \quad & \lambda L  
 \end{pmatrix}  =  \lambda \begin{pmatrix}
   1 & 0 \\ 
   0 & 1   
 \end{pmatrix} \otimes L
 \;\;    \in    \matR^{2n \times 2n}.
\end{eqnarray} 
\vspace{-6mm}

In light of this, the optimization problem \eqref{eq:qcqp_denoise_complex} can be 
equivalently formulated as 

\vspace{-7mm}
\begin{eqnarray} \label{eq:qcqp_denoise_real}
\min_{ \vecgbar \in \mathbb{R}^{2n}: \norm{\vecgbar}^2 = n} \vecgbar^{T} \Hbar \vecgbar  - 2 \vecgbar^{T} \veczbar,  
\end{eqnarray}
\vspace{-6mm}

which is formally shown in the appendix for completeness. Let us note that 
the Laplacian matrix $L$ is positive semi-definite (p.s.d),  with its smallest eigenvalue $ \lambda_1(L) = 0$ with multiplicity $1$ (since $G$ is connected). 
Therefore,  $\Hbar$ is also p.s.d, with smallest eigenvalue $ \lambda_1(H) = 0$ with 
multiplicity $2$. 

\eqref{eq:qcqp_denoise_real} is actually an instance of the so-called trust region subproblem (TRS) with equality 
constraint (which we denote by $\trseq$ from now on),  where one minimizes a general quadratic function (not necessarily convex), subject to a sphere constraint. 
For completeness, we also mention the closely related trust region subproblem with inequality constraint (denoted by  $\trsineq$), 
where we have a $\ell_2$ ball constraint. There exist several algorithms that efficiently 
solve $\trsineq$ (cf., \cite{Sorensen82,MoreSoren83,Rojas01,Rendl97,Gould99,Naka17}) 
and also some which explicitly solve $\trseq$ (cf., \cite{Hager01,Naka17}). In particular, we note the recent work in \cite{Naka17} which 
showed that trust region subproblems can be solved to high accuracy via a single generalized eigenvalue problem\footnote{The computational 
complexity is $O(n^3)$ in the worst case, but improves when the matrices involved are sparse. In our case, the Laplacian  
is sparse when $k$ is not large.}. In our experiments, we employ 
their algorithm for solving \eqref{eq:qcqp_denoise_real}.
  
Rather surprisingly, one can fully characterize\footnote{Discussed in detail in the appendix for completeness.} 
the solutions to $\trseq$ and  $\trsineq$.  
The following Lemma \ref{lemma:qcqp_denoise_real} characterizes the solution for \eqref{eq:qcqp_denoise_real}; 
it follows directly from \cite[Lemma 2.4, 2.8]{Sorensen82} (also \cite[Lemma 1]{Hager01}).
%
%
\begin{lemma} \label{lemma:qcqp_denoise_real}
$\vecgbarest$ is a solution to \eqref{eq:qcqp_denoise_real} iff $\norm{\vecgbarest}^2 = n$ and $\exists \mu^{*}$ such that
(a) $2\Hbar + \mu^{*}\matI \succeq 0$ and (b) $(2\Hbar + \mu^{*}\matI) \vecgbarest = 2\veczbar$. 
Moreover, the solution is unique if $2\Hbar + \mu^{*}\matI \succ 0$. 
\end{lemma}
We analyze the solution of \eqref{eq:qcqp_denoise_real} with the help of Lemma \ref{lemma:qcqp_denoise_real} 
in the appendix.

\begin{remark}
Note that \eqref{eq:orig_denoise_hard_1} is similar to  
angular synchronization \cite{sync} as they 
both optimize a quadratic form subject to entries lying on the unit circle.
The fundamental difference is that
the matrix in the quadratic term in synchronization is formed
using the given noisy pairwise angle offsets (embedded on the
unit circle), and thus depends on the data. In our setup, the
quadratic term is formed using the Laplacian of the smoothness
regularization graph, and thus is independent of the data (noisy mod 1 samples). 
\end{remark}


\vspace{-2mm}
\subsection{Recovering the denoised mod 1 samples} \label{subsec:rec_denoised_mod1} 
\vspace{-2mm}

The solution to \eqref{eq:qcqp_denoise_real} is 
a vector $\vecgbarest \in \matR^{2n}$. Let $\vecgest \in \mathbb{C}^n$ be the complex representation of $\vecgbarest$ 
as per \eqref{eq:real_notat} so that $\vecgbarest = [\real(\vecgest)^T \ \imag(\vecgest)^T]^T$. Denoting $\gest_i \in \mathbb{C}$ to 
be the $i^{th}$ component of $\vecgest$, note that $\abs{\gest_i}$ is not necessarily equal to one. On the other hand, recall that 
$h_i  =\exp(\iota 2\pi f_i \bmod 1),$ $\forall i=i,\dots,n$ for the ground truth $\vechtil \in \mathbb{C}^n$. 
We obtain our final estimate $\widehat{f_i} \bmod 1$ to $f_i \bmod 1$ by projecting $\gest_i$ onto 
the unit complex disk 
%
%

\vspace{-7mm}
\begin{equation} \label{eq:extr_mod1_vals}
\exp(\iota 2\pi (\widehat{f_i} \bmod 1)) = \frac{\gest_i}{\abs{\gest_i}}; \quad i=1,\dots,n.
\end{equation}
\vspace{-5mm}

%
%
In order to measure the distance between $\widehat{f_i} \bmod 1$ and $f_i \bmod 1$, we will use 
the so called  \emph{wrap-around} distance on $[0,1]$ denoted by $d_w : [0,1]^2 \rightarrow [0,1/2]$, where

\vspace{-6mm}
\begin{align}
d_w(t_1,t_2) := \min\set{\abs{t_1-t_2}, 1-\abs{t_1-t_2}}
\end{align}
\vspace{-6mm}

for $t_1,t_2 \in [0,1]$. We will now show that if $\gest_i$ is sufficiently close to $h_i$ for each $i=1,\dots,n$, then 
each $d_w(\widehat{f_i} \bmod 1,f_i\bmod 1)$ will be correspondingly small. This is stated precisely in the 
following lemma, its proof being deferred to the appendix.
%
\begin{lemma} \label{lem:wrap_dist_fin_bd}
For $0 < \epsilon < 1/2$, let $\abs{\gest_i - h_i} \leq \epsilon$ hold for each $i=1,\dots,n$. Then,  for each $i=1,\dots,n$

\vspace{-7mm}
\begin{equation} \label{eq:wrap_dist_fin_bd}
d_w(\widehat{f_i} \bmod 1,f_i\bmod 1) \leq \frac{1}{\pi} \sin^{-1}\left(\frac{\epsilon}{1-\epsilon}\right).
\end{equation}
\vspace{-6mm}
\end{lemma}

\vspace{-2mm}
\subsection{Unwrapping stage and main algorithm} \label{subsec:unwrap_stage_and_algo} 
\vspace{-2mm}

Having recovered the denoised mod 1 samples $\widehat{f_i} \bmod 1$ for $i=1,\dots,n$, 
we now move onto the next stage of our method where the goal is to recover the samples $f$, for which we discuss two possible approaches. 
%
%
%

\textbf{1. Quotient tracker (QT) method.} \label{item:qt_unwrap}
The first approach for unwrapping the mod 1 samples is perhaps 
the most natural one, we outline it below for the setting where $G$ is a line graph, i.e., $k=1$. 
It is based on the idea that provided the denoised mod 1 samples are very close estimates 
to the original clean mod 1 samples, then we can sequentially find the quotient terms, by checking whether 
$\abs{\widehat{f}_{i+1} \bmod 1 - \widehat{f_i} \bmod 1} \geq \qtathresh$, for a suitable threshold parameter $\qtathresh \in (0,1)$. 
More formally, by initializing $\widehat{q}_1 = 0$ consider the rule

\vspace{-8mm}
\begin{align} \label{eq:qta_recovery_rule}
\widehat{q}_{i+1} &= \widehat{q}_i + \sign_{\qtathresh}(\widehat{f}_{i+1} \bmod 1 - \widehat{f_i} \bmod 1); \nonumber \\  
\sign_{\qtathresh}(t) &= \left\{
\begin{array}{rl}
-1 ; &  t \geq \qtathresh \\
0 ; &  \abs{t} < \qtathresh \\ 
1; & t \leq -\qtathresh
\end{array}. \right.
\end{align}
\vspace{-5mm}

Clearly, if $\widehat{f_{i}} \bmod 1 \approx f_i \bmod 1$ for each $i$, then for $n$ sufficiently large,
the procedure \eqref{eq:qta_recovery_rule} will result in correct recovery of the quotients. 
However, it is also obviously sensitive to noise, and hence would not be a viable option when the noise level is high.
%
%

\begin{figure*}
\centering
\subcaptionbox[]{  $\gamma=0.15$
 \label{subfig:instances_f1_Bounded_delta_cors__Cor_gamma0p15}
}[ 0.17\textwidth ]
{\includegraphics[width=0.17\textwidth] {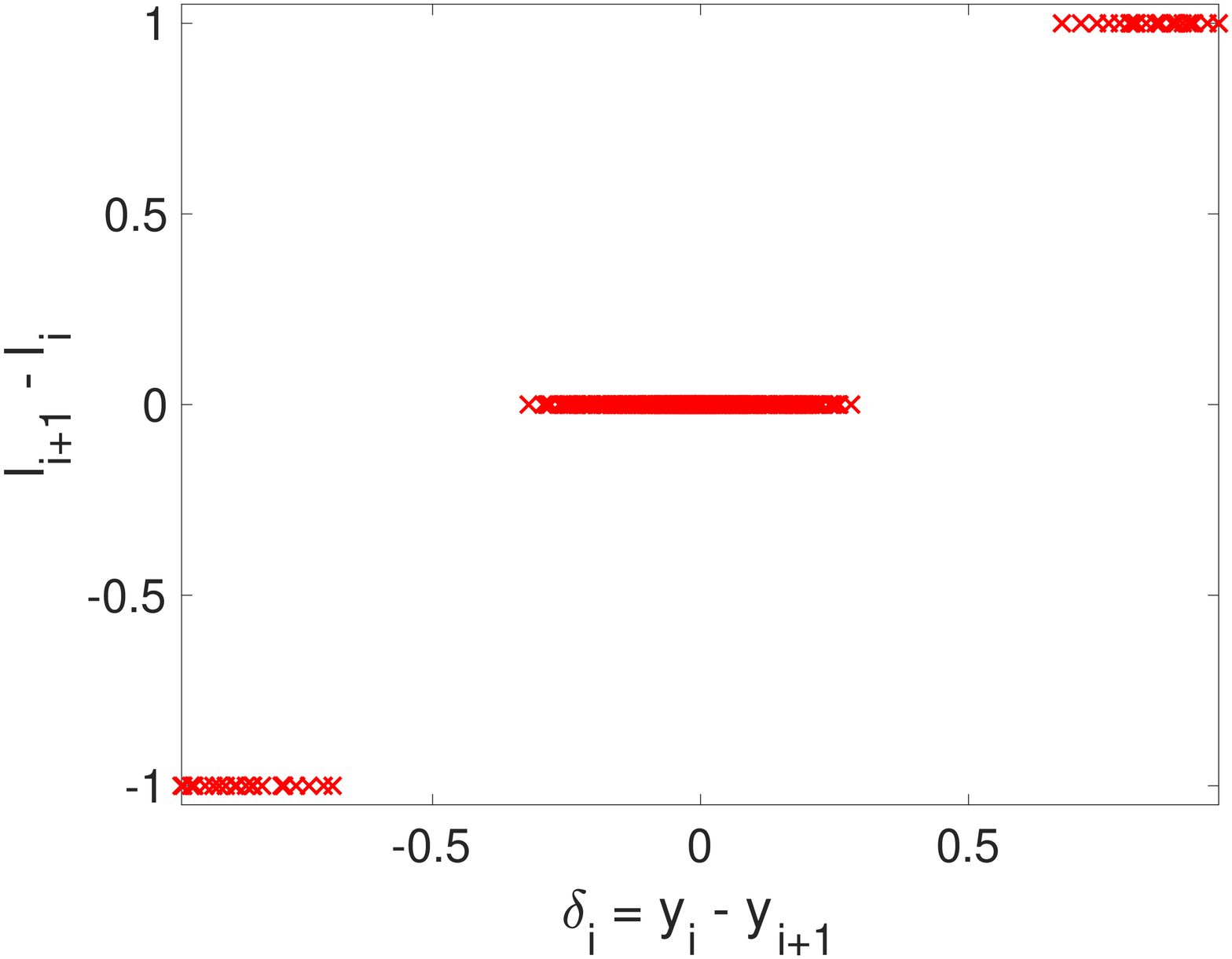} }
\hspace{0.01\textwidth} 
\subcaptionbox[]{  $\gamma=0.15$
}[ 0.67\textwidth ]
{\includegraphics[width=0.67\textwidth] {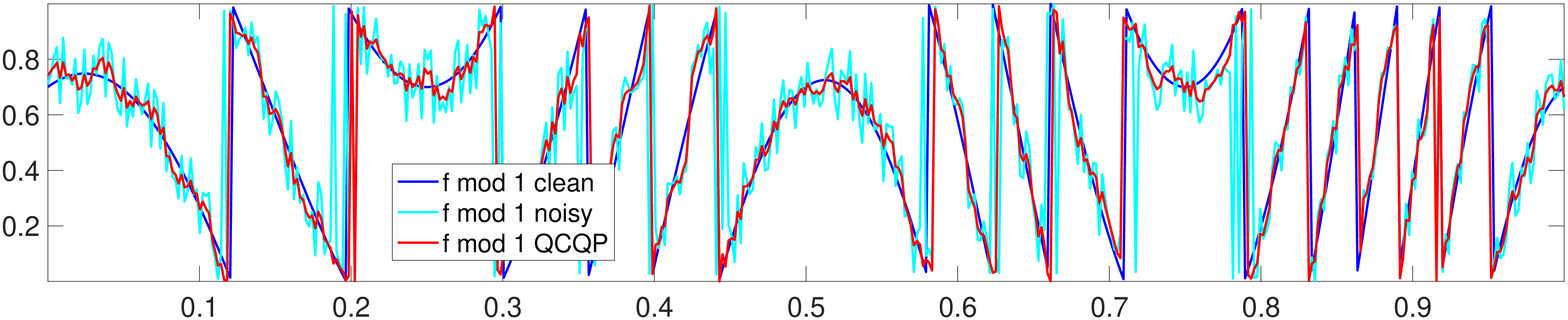} }
\hspace{0.01\textwidth} 
\subcaptionbox[]{  $\gamma=0.25$
}[ 0.17\textwidth ]
{\includegraphics[width=0.17\textwidth] {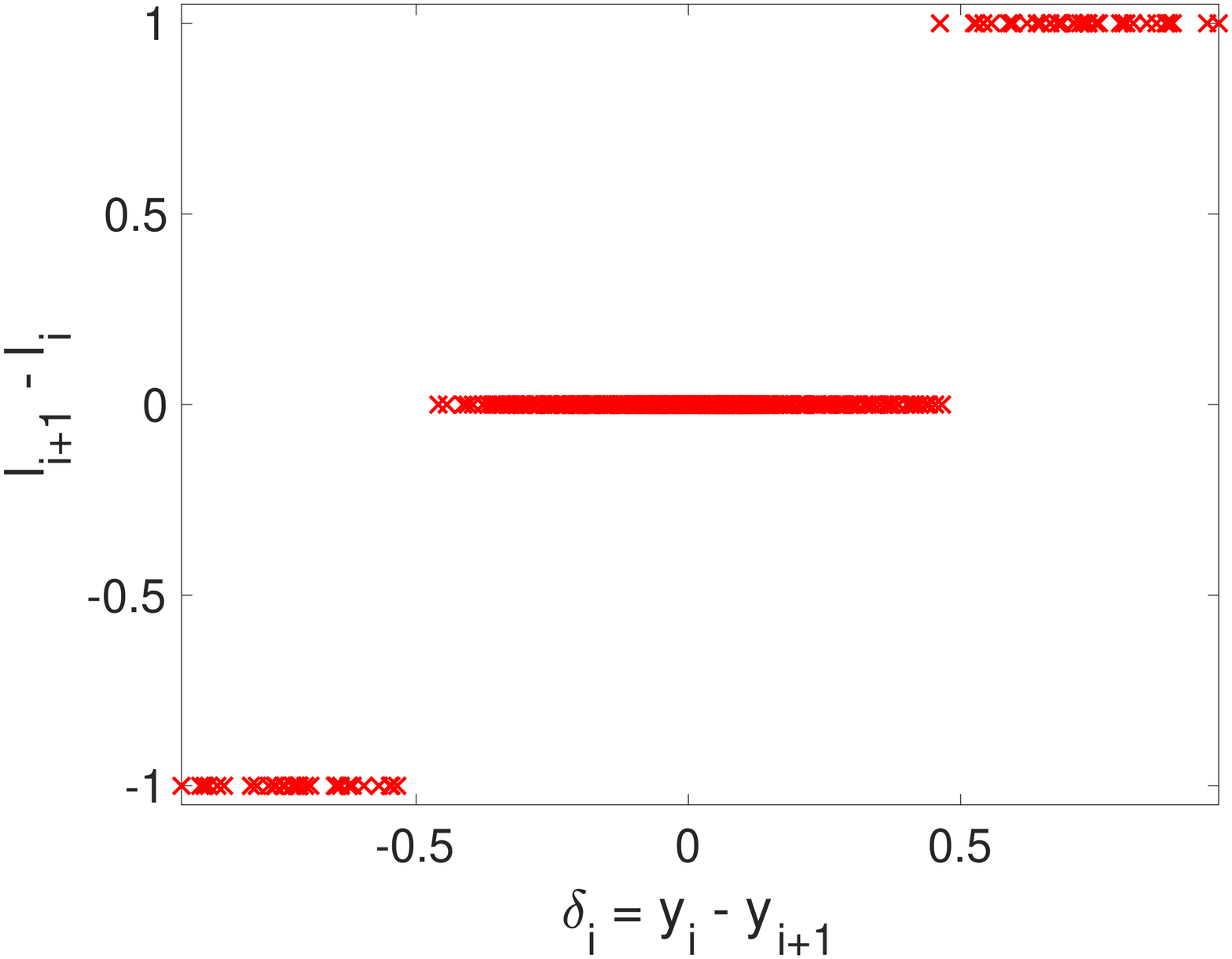} }
\hspace{0.01\textwidth} 
\subcaptionbox[]{  $\gamma=0.25$
}[ 0.67\textwidth ]
{\includegraphics[width=0.67\textwidth] {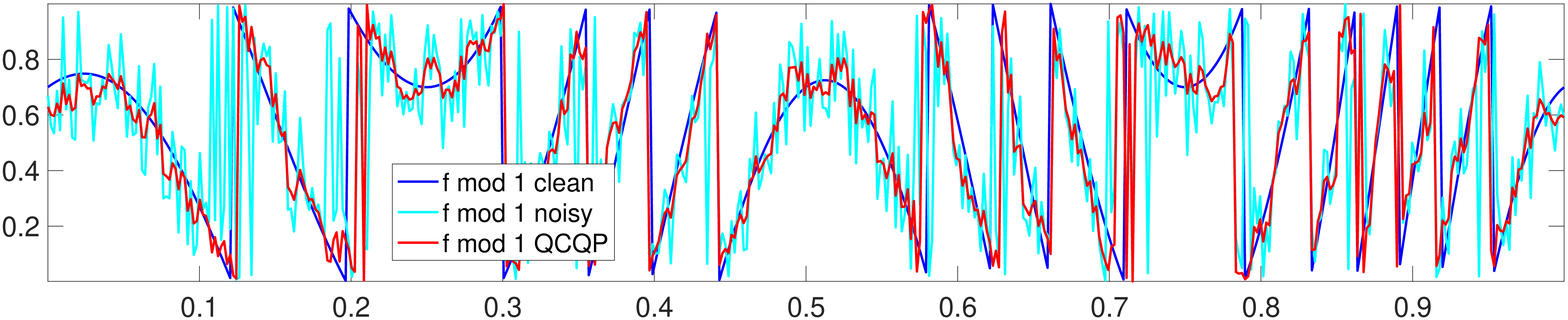} }
\hspace{0.01\textwidth} 
\subcaptionbox[]{  $\gamma=0.30$
}[ 0.17\textwidth ]
{\includegraphics[width=0.17\textwidth] {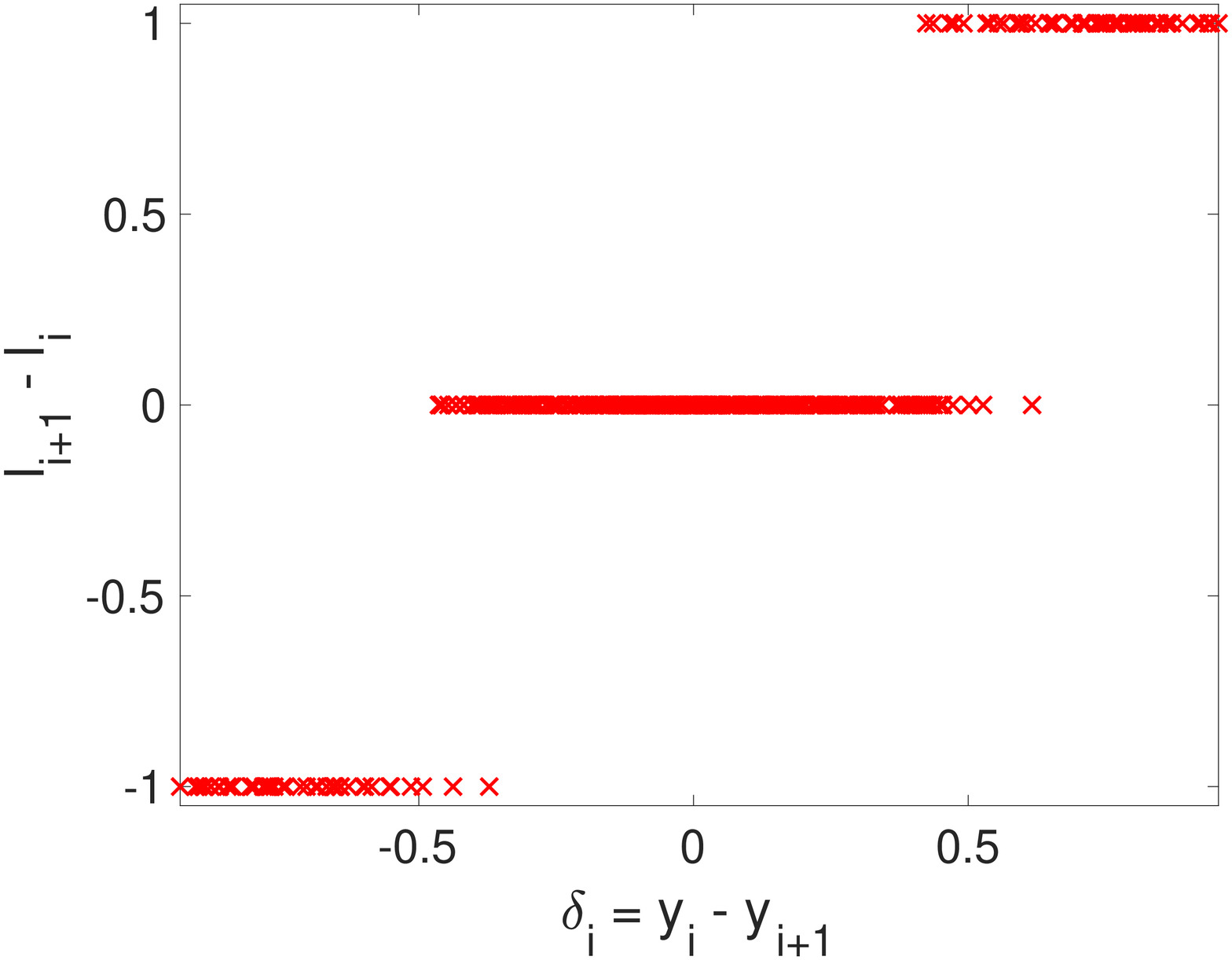} }
\hspace{0.01\textwidth} 
\subcaptionbox[]{  $\gamma=0.30$
}[ 0.67\textwidth ]
{\includegraphics[width=0.67\textwidth] {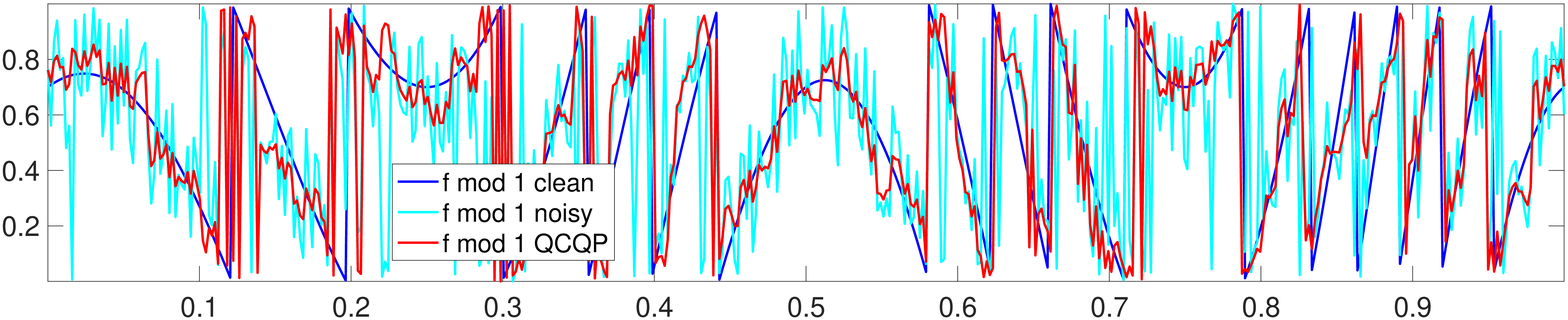} }
%
\vspace{-3mm}
\captionsetup{width=0.95\linewidth}
\caption[Short Caption]{Noisy instances of the Uniform noise model ($n=500$) for $f(x) = 4x \cos^2(2\pi x) - 2 \sin^2(2\pi x)$. Left: scatter plot of change in $y$ (the observed noisy f mod 1 values) versus change in $l$ (the noisy quotient). Right: plot of the clean f mod 1 values (blue), the noisy f mod 1 values (cyan) and the denoised (via QCQP) f mod 1 values. 
}
\label{fig:instances_f1_Bounded_delta_cors}
\end{figure*}

\textbf{2. Ordinary least-squares (OLS) based method.} \label{item:ols_unwrap}
A robust alternative to the aforementioned approach 
is based on directly recovering the function via a simple least squares problem. Recall that in the noise-free case, $f_i = q_i + r_i$, $ q_i \in \mathbb{Z}, r_i \in [0,1) $, and consider, for a pair of nearby points $(i,j)$, the difference $f_i - f_j = q_i - q_j  + r_i - r_j, \; i =1, \ldots, n$. The OLS formulation we solve stems from the observation that, if $| r_i - r_j | < \qtathresh$ for a small $\qtathresh$, then $q_i = q_j$. This intuition can be easily gained from the left plots of 
Figure \ref{fig:instances_f1_Bounded_delta_cors}, especially   \ref{subfig:instances_f1_Bounded_delta_cors__Cor_gamma0p15}, which pertains to the noisy case (but in the low noise regime $\gamma=0.15$), that plots   $l_{i+1} - l_i$  versus  $y_i - y_{i+1}$, where  $l_i$ denotes the noisy quotient of sample $i$, and $y_i$ the noisy remainder. For small enough $|y_i - y_{i+1}|$, we observe that  $| l_{i+1} - l_i | = 0$. Whenever  $y_i - y_{i+1} > \qtathresh$, we see that $ l_{i+1} - l_i = 1$, while $y_i - y_{i+1} < - \qtathresh$, indicates that $ l_{i+1} - l_i = -1$. Throughout  all our experiments we set $\qtathresh=0.5$. 
In Figure \ref{fig:instances_f1_Bounded_main} we also plot the true quotient $q$, which can be observed to be piecewise constant, in agreement with our above intuition.


For a graph $G = (V,E)$ with $k \in \mathbb{N}$, and for a suitable threshold parameter $\qtathresh \in (0,1)$, this intuition leads us to estimate the function values $f_i$ as the least-squares solution to the overdetermined system of linear equations \eqref{eq:fmod1}, 
without involving  the quotients $q_1, \ldots, q_n$. To this end, we consider a linear system of equations for the function differences $f_i - f_j,  \quad \forall (i,j) \in E$ 
%
%

\vspace{-8mm}
\begin{equation} \label{eq:ols_unwrap_lin_system}
	f_i - f_j = l_i - l_j  + y_i - y_j =  \sign_{\qtathresh}( y_{i}  - y_{j})   +   y_i - y_j,
\end{equation}
\vspace{-7mm}

and solve it in the least-squares sense. \eqref{eq:ols_unwrap_lin_system} is analogous to \eqref{eq:qta_recovery_rule}, except that we now recover 
$(\widehat{f}_{i})_{i=1}^n$ collectively as the least-squares solution to \eqref{eq:ols_unwrap_lin_system}. Denoting by $T$ the least-squares matrix associated with the overdetermined linear system \eqref{eq:ols_unwrap_lin_system}, and letting $b_{i,j} = \sign_{\qtathresh}( y_{i}  - y_{j})   +   y_i - y_j $, the system of equations can be written as $ T f = \mathbf{b}$ where $\mathbf{b} \in \mathbb{R}^{\abs{E}}$. Note that the matrix $T$ is sparse with only two non-zero entries per row, and that the all-ones vector $\mb{1} = ( 1, 1, \ldots, 1)^T$ lies in the null space of $T$, i.e., $T \mb{1} = 0 $. Therefore, we will find the minimum norm least-squares solution to \eqref{eq:ols_unwrap_lin_system}, and  recover $f$ only up to a global shift.

%
%
Algorithm \ref{algo:two_stage_denoise} summarizes our two-stage method for recovering the samples of $f$ (up to a global shift). 
Figure \ref{fig:instances_f1_Bounded_delta_cors} shows additional  noisy instances of the Uniform noise model. 
The scatter plots on the left show that, as the noise  level  
increases, the function \eqref{eq:qta_recovery_rule} will produce more and more errors in \eqref{eq:ols_unwrap_lin_system}. 
The right plots show the corresponding f mod 1 signal (clean, noisy, and denoised via Algorithm \ref{algo:two_stage_denoise}) for three levels of noise. 

%
%
\vspace{-3mm}
\begin{algorithm}[!ht]
\caption{Algorithm for recovering the samples $f_i$} \label{algo:two_stage_denoise} 
\begin{algorithmic}[1] 
\State \textbf{Input:} $(y_i)_{i=1}^n$ (noisy mod 1 samples), $k$, $\lambda$, $n$, $G=(V,E)$. 
\State \textbf{Output:} Denoised mod 1 samples $\widehat{f_i}\bmod 1$; $i=1,\dots,n$. 

\textsc{// Stage 1: Recovering denoised} mod 1 \textsc{samples of} $f$. 

\State Form $\Hbar \in R^{2n \times 2n}$ using $\lambda,L$ as in \eqref{eq:def_H}.

\State Form $\veczbar = [\real(\vecz)^T \imag(\vecz)^T]^T \in \matR^{2n}$ as in \eqref{eq:real_notat}.

\State Obtain $\vecgbarest \in \matR^{2n}$ as the solution to \eqref{eq:qcqp_denoise_real}, i.e., 
%
$$\vecgbarest = \argmin{\vecgbar \in \mathbb{R}^{2n}: \norm{\vecgbar}^2 = n} \vecgbar^{T} \Hbar \vecgbar  - 2 \vecgbar^{T} \veczbar.$$ 

\State Obtain $\vecgest \in \mathbb{C}^n$ from $\vecgbarest$ where $\vecgbarest = [\real(\vecgest)^T \imag(\vecgest)^T]^T$.

\State Recover $\est{f_i}\bmod 1 \in [0,1)$ from $\frac{\gest_i}{\abs{\gest_i}}$ for each $i=1,\dots,n$, as in \eqref{eq:extr_mod1_vals}.

\textsc{// Stage 2: Recovering denoised real valued samples of} $f$.

\State \textbf{Input:} $(\est{f_i}\bmod 1)_{i=1}^n$ (denoised mod 1 samples), $G=(V,E)$, $\qtathresh \in (0,1)$. 

\State \textbf{Output:} Denoised samples $\widehat{f_i}$; $i=1,\dots,n$. 

\State Obtain $(\est{f}_i)_{i=1}^n$ via the Quotient tracker (QT) or 
       OLS based method for suitable threshold $\qtathresh$.

\end{algorithmic}
\end{algorithm}
\vspace{0mm}

\section{Analysis for the arbitrary bounded noise model} \label{sec:bound_noise_analysis}
\vspace{-3mm}
We now provide some approximation guarantees for the solution $\vecgbarest \in \matR^{2n}$ 
to \eqref{eq:qcqp_denoise_real} for the arbitrary bounded noise model \eqref{eq:arb_bd_noise_model}. 
In particular, we consider a slightly modified version of this model, assuming 

\vspace{-5mm}
\begin{equation}  \label{eq:BoundedNoiseModel}
\norm{\veczbar - \vechtilbar}_2 \leq \delta \sqrt{n}
\end{equation}
\vspace{-7mm}

holds true for some $\delta \in [0,1]$. This is reasonable, since    
$\norm{\veczbar - \vechtilbar}_2 \leq 2 \sqrt{n}$ holds in general by triangle inequality. 
Also, note for \eqref{eq:arb_bd_noise_model} that $\abs{z_i - h_i} = 2\abs{\sin (\pi (\delta_i \bmod 1))} \leq 2\pi\abs{\delta_i}$, and thus 
$\norm{\veczbar - \vechtilbar}_2 = \norm{\vecz - \vechtil}_2 \leq  2 \pi \max_i ( |\delta_i| ) \sqrt{n}$.
Hence, while a small enough uniform bound on $\max_i (|\delta_i|)$ would of course imply \eqref{eq:BoundedNoiseModel}, however, clearly  
\eqref{eq:BoundedNoiseModel} can also hold even if some of the $\delta_i$'s are large.  
%
\begin{theorem} \label{thm:arb_noise_model} 
Under the above notation and assumptions,  
consider the arbitrary bounded
noise model in \eqref{eq:arb_bd_noise_model}, with $\veczbar$ satisfying 
$\norm{\veczbar - \vechtilbar}_2 \leq \delta \sqrt{n}$ for $\delta \in [0,1]$. 
Let $n \geq 2$, and let $\calN(\Hbar)$ denote the null space of $\Hbar$.

\vspace{-3mm}
\begin{enumerate}
\item If $\veczbar \not\perp \calN(\Hbar)$ then $\vecgbarest$ is the unique solution 
to \eqref{eq:qcqp_denoise_real} satisfying 
\vspace{-2mm}
\begin{align}
\hspace{-8mm} \frac{1}{n}\dotprod{\vechtilbar}{\vecgbarest} \geq 1 &- \frac{3\delta}{2} - \frac{\lambda \pi^2 M^2 (2k)^{2\alpha + 1}}{n^{2\alpha}}  \nonumber \\ 
&+ \frac{1}{(4\lambda k + 1)^2} \left(\frac{1}{2n} \veczbar^T \Hbar \veczbar\right).
\end{align}

\vspace{-4mm}
\item If $\veczbar \perp \calN(\Hbar)$ and $\lambda < \frac{1}{4k}$ then $\vecgbarest$ is the unique solution 
to \eqref{eq:qcqp_denoise_real} satisfying

\vspace{-5mm}
\begin{align}
\hspace{-6mm} \frac{1}{n}\dotprod{\vechtilbar}{\vecgbarest} &\geq 1 - \frac{3\delta}{2} - \frac{\lambda \pi^2 M^2 (2k)^{2\alpha + 1}}{n^{2\alpha}} \nonumber \\
 &+  \frac{1}{\left(1 + 4\lambda k - 4 \lambda k \sin^2\left(\frac{\pi}{2n}\right) \right)^2} \left(\frac{1}{2n} \veczbar^T \Hbar \veczbar\right).
\end{align}
\end{enumerate}
\end{theorem}

\vspace{-4mm}
The following useful Corollary of Theorem \ref{thm:arb_noise_model} is a direct consequence of the fact that 
$1/(2n) \veczbar^T \Hbar \veczbar \geq 0$ 
for all $\veczbar \in \matR^{2n}$, since $\Hbar$ is positive semi-definite.
%
\begin{corollary} \label{cor:main_thm_bounded_noise}
Consider the arbitrary bounded noise model in \eqref{eq:arb_bd_noise_model}, with $\veczbar$ satisfying 
$\norm{\veczbar - \vechtilbar}_2 \leq \delta \sqrt{n}$ for $\delta \in [0,1]$. 
Let $n \geq 2$. If $\lambda < \frac{1}{4k}$ then $\vecgbarest$ is the unique solution 
to \eqref{eq:qcqp_denoise_real} satisfying 
\begin{equation}
\frac{1}{n}\dotprod{\vechtilbar}{\vecgbarest} \geq 1 - \frac{3\delta}{2} - \frac{\lambda \pi^2 M^2 (2k)^{2\alpha + 1}}{n^{2\alpha}}.
\end{equation}
\end{corollary}

\vspace{-3mm}
Before presenting the proof of Theorem \ref{thm:arb_noise_model}, some remarks are in order.
\vspace{-1mm}

\textbf{1.} Theorem \ref{thm:arb_noise_model} give us a lower bound on the correlation between 
$\vechtilbar, \vecgbarest \in \matR^{2n}$, where clearly, $\frac{1}{n}\dotprod{\vechtilbar}{\vecgbarest} \in [-1,1]$. 
Note that the correlation improves when the noise term $\delta$ decreases, as one would expect.  
The term $\frac{\lambda \pi^2 M^2 (2k)^{2\alpha + 1}}{n^{2\alpha}}$ effectively arises on account of the smoothness 
of $f$, and is an upper bound on the term $\frac{1}{2n} \vechtilbar^T \Hbar \vechtilbar$ (made clear in Lemma \ref{lem:upbd_clean_quad_form}). 
Hence as the number of samples increases, $\frac{1}{2n} \vechtilbar^T \Hbar \vechtilbar$ goes to zero at the rate $n^{-2\alpha}$ (for fixed $k,\lambda$). 
Also note that the lower bound on $\frac{1}{n}\dotprod{\vechtilbar}{\vecgbarest}$ readily implies the $\ell_2$ norm bound 
$\norm{\vecgbarest-\vechtilbar}^2_2 = O(\delta n + n^{1-2\alpha})$.

\vspace{-1mm}
\textbf{2.} The term $\frac{1}{2n} \veczbar^T \Hbar \veczbar$ represents the smoothness of the observed noisy samples. 
While an increasing amount of noise would usually render 
$\veczbar$ to be more and more non-smooth, and thus typically increase $\frac{1}{2n} \veczbar^T \Hbar \veczbar$, note that this would be met by a corresponding 
increase in $\delta$, and hence the lower bound on the correlation would not necessarily improve.

\vspace{-1mm}
\textbf{3.} It is easy to verify that \eqref{eq:BoundedNoiseModel} implies $\dotprod{\veczbar}{\vechtilbar}/n \geq 1-(\delta/2)$.
Thus for $\veczbar$, which is feasible for $(P)$, we have a bound on correlation which is better than the bound in 
Corollary \ref{cor:main_thm_bounded_noise} by a $\delta + O(n^{-2\alpha})$ term. However, the solution $\vecgbarest$ 
to $(P)$ is a \emph{smooth} estimate of $\vechtilbar$ (and hence more interpretable), while $\veczbar$ is typically highly 
non-smooth. 
%

\vspace{-4mm}
\paragraph{Proof of Theorem \ref{thm:arb_noise_model}.} 
The proof of Theorem \ref{thm:arb_noise_model} relies heavily on Lemma \ref{lem:lowbd_feas_based} 
outlined below, whose proof is deferred to the appendix.
\begin{lemma} \label{lem:lowbd_feas_based}
Consider the arbitrary bounded noise model in \eqref{eq:arb_bd_noise_model}, with $\veczbar$ satisfying 
$\norm{\veczbar - \vechtilbar}_2 \leq \delta \sqrt{n}$ for $\delta \in [0,1]$. 
Any solution $\vecgbarest$ to \eqref{eq:qcqp_denoise_real} satisfies
\begin{equation} \label{eq:lowbd_ang_feas_form}
\frac{1}{n}\dotprod{\vechtilbar}{\vecgbarest} \geq  1 - \frac{3\delta}{2} 
 - \; \;   \frac{1}{2n} \vechtilbar^T \Hbar \vechtilbar + \; \; \frac{1}{2n} \vecgbarest^T \Hbar \vecgbarest. 
\end{equation}
\end{lemma}

We now upper bound the term $\frac{1}{2n} \vechtilbar^T \Hbar \vechtilbar$ 
in \eqref{eq:lowbd_ang_feas_form} using the H\"older continuity of $f$. 
This is formally shown below in the form of Lemma \ref{lem:upbd_clean_quad_form}, its proof 
is deferred to the appendix.
\begin{lemma} \label{lem:upbd_clean_quad_form}
For $n \geq 2$, the following is true.
\begin{equation}
\frac{1}{2n} \vechtilbar^T \Hbar \vechtilbar \leq \frac{\lambda \pi^2 M^2 (2k)^{2\alpha + 1}}{n^{2\alpha}}, 
\end{equation} 
where $\alpha \in (0,1]$ and $M > 0$ are related to the smoothness of $f$ and defined in \eqref{eq:f_smooth_hold}, and  
$\lambda \geq 0$ is the regularization parameter in \eqref{eq:orig_denoise_hard}.
\end{lemma}

\vspace{-3mm}
Lastly, we lower bound the term $\frac{1}{2n} \vecgbarest^T \Hbar \vecgbarest$ in \eqref{eq:lowbd_ang_feas_form}
using knowledge of the structure of the solution $\vecgbar$. This is outlined below as Lemma \ref{lem:lowbd_sol_quad_form}, 
its proof is deferred to the appendix. This completes the proof of Theorem \ref{thm:arb_noise_model}.
\begin{lemma} \label{lem:lowbd_sol_quad_form}
Denoting $\calN(\Hbar)$ to be the null space of $\Hbar$, the following holds 
for the solution $\vecgbarest$ to \eqref{eq:qcqp_denoise_real}.

\vspace{-3mm}
\begin{enumerate}
\item If $\veczbar \not\perp \calN(\Hbar)$ then $\vecgbarest$ is unique and 
\begin{equation}
\frac{1}{2n} \vecgbarest^T \Hbar \vecgbarest \geq \frac{1}{\left(1 + 4\lambda k \right)^2} \left(\frac{1}{2n} \veczbar^T \Hbar \veczbar\right).
\end{equation}

\vspace{-3mm}
\item If $\veczbar \perp \calN(\Hbar)$ and $\lambda < \frac{1}{4k}$, then $\vecgbarest$ is unique and 
{\small
\begin{equation}
\frac{1}{2n} \vecgbarest^T \Hbar \vecgbarest \geq 
\frac{1}{\left(1 + 4\lambda k - 4 \lambda k \sin^2\left(\frac{\pi}{2n}\right)\right)^2} \left(\frac{1}{2n} \veczbar^T \Hbar \veczbar\right).
\end{equation}}
\end{enumerate}
\end{lemma}
%
%


\vspace{-7mm}
\section{Numerical experiments} \label{sec:num_exps}
\vspace{-3mm}
This section contains numerical experiments for the two noise models 
discussed in Section \ref{sec:ProblemSetup}, 
 the Uniform model (with samples generated uniformly in $[-\gamma,\gamma]$ for bounded $\gamma$) 
and the Gaussian model. The function $f$ considered is $f(x) = 4x \cos^2(2\pi x) - 2 \sin^2(2\pi x)$.
For each experiment (averaged over 20 trials), we show the RMSE error on a log scale, for denoised samples of $f$ mod 1 
and $f$. For the latter, we compute the RMSE after we mod out the best global shift\footnote{Any algorithm that takes as input the $\mod 1$ samples will be able to recover $f$ only up to a global shift.}. We compare the performance of three algorithms. 

%
\textbf{OLS} denotes the algorithm based on the least squares formulation \eqref{item:ols_unwrap} used to recover samples of $f$, 
and works directly with noisy$\mod 1$ samples; the estimated $f$  mod 1 values are then obtained as the corresponding  
mod 1 versions.
%
\textbf{QCQP} denotes Algorithm \ref{algo:two_stage_denoise} where the unwrapping stage is performed via \textbf{OLS} \eqref{item:ols_unwrap}. 
We solve TRS via the generalized eigenvalue based 
solver\footnote{Available at \url{http://www.opt.mist.i.u-tokyo.ac.jp/~nakatsukasa/codes/TRSgep.m}} of \cite{Naka17}.
%
\textbf{iQCQP} denotes an iterated version of \textbf{QCQP}, wherein we repeatedly denoise
the noisy $f \mod 1$ estimates (via Stage $1$ of Algorithm \ref{algo:two_stage_denoise}) for $10$ iterations, 
and  perform the unwrapping stage via \textbf{OLS} \eqref{item:ols_unwrap}  to recover sample estimates of $f$.
%
%
%
%

\begin{figure}
\centering
\subcaptionbox[]{  $\gamma=0.27$, \textbf{OLS}
}[ 0.20\textwidth ]
{\includegraphics[width=0.20\textwidth] {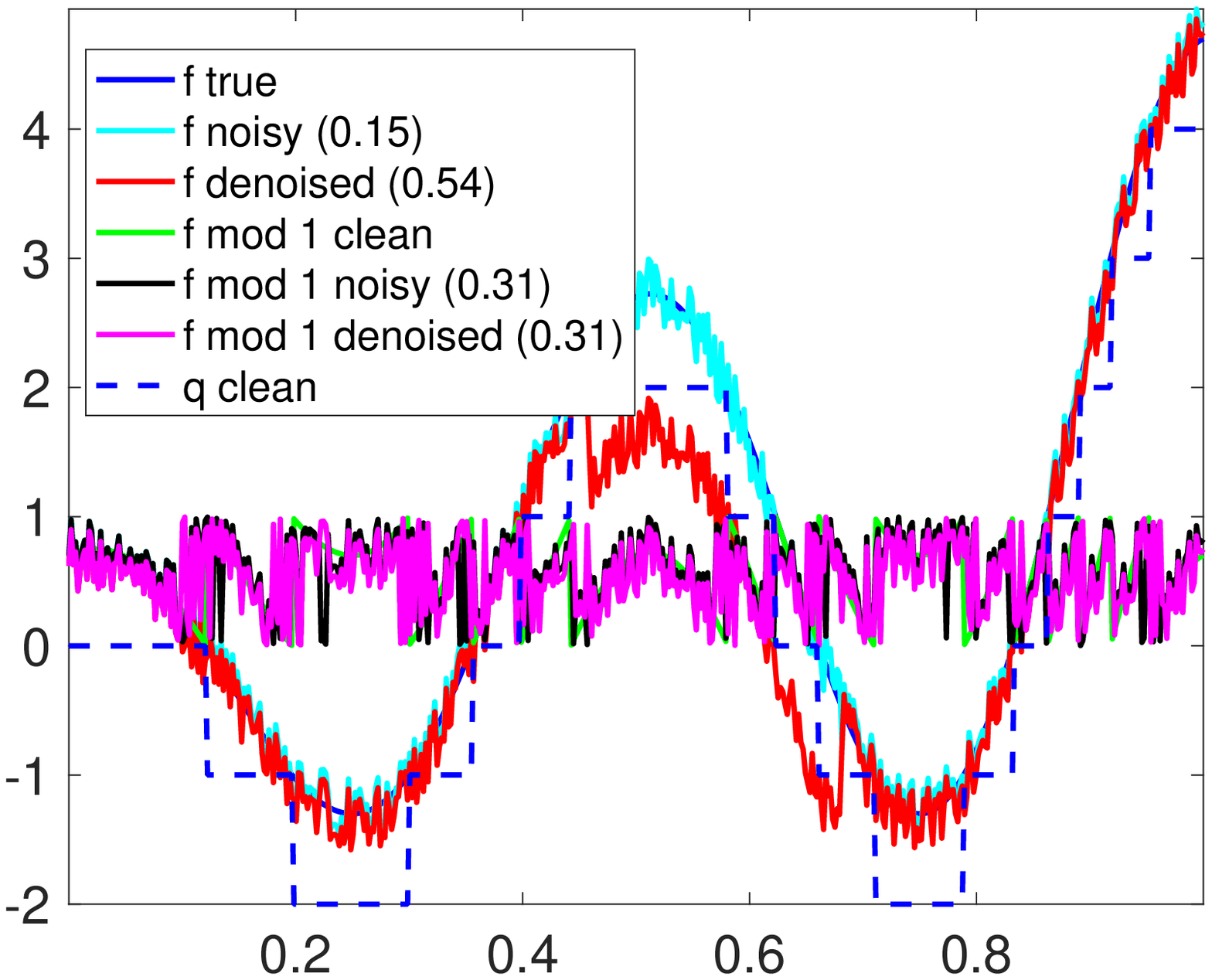} }
%
\subcaptionbox[]{  $\gamma=0.30$, \textbf{OLS}
}[ 0.20\textwidth ]
{\includegraphics[width=0.20\textwidth] {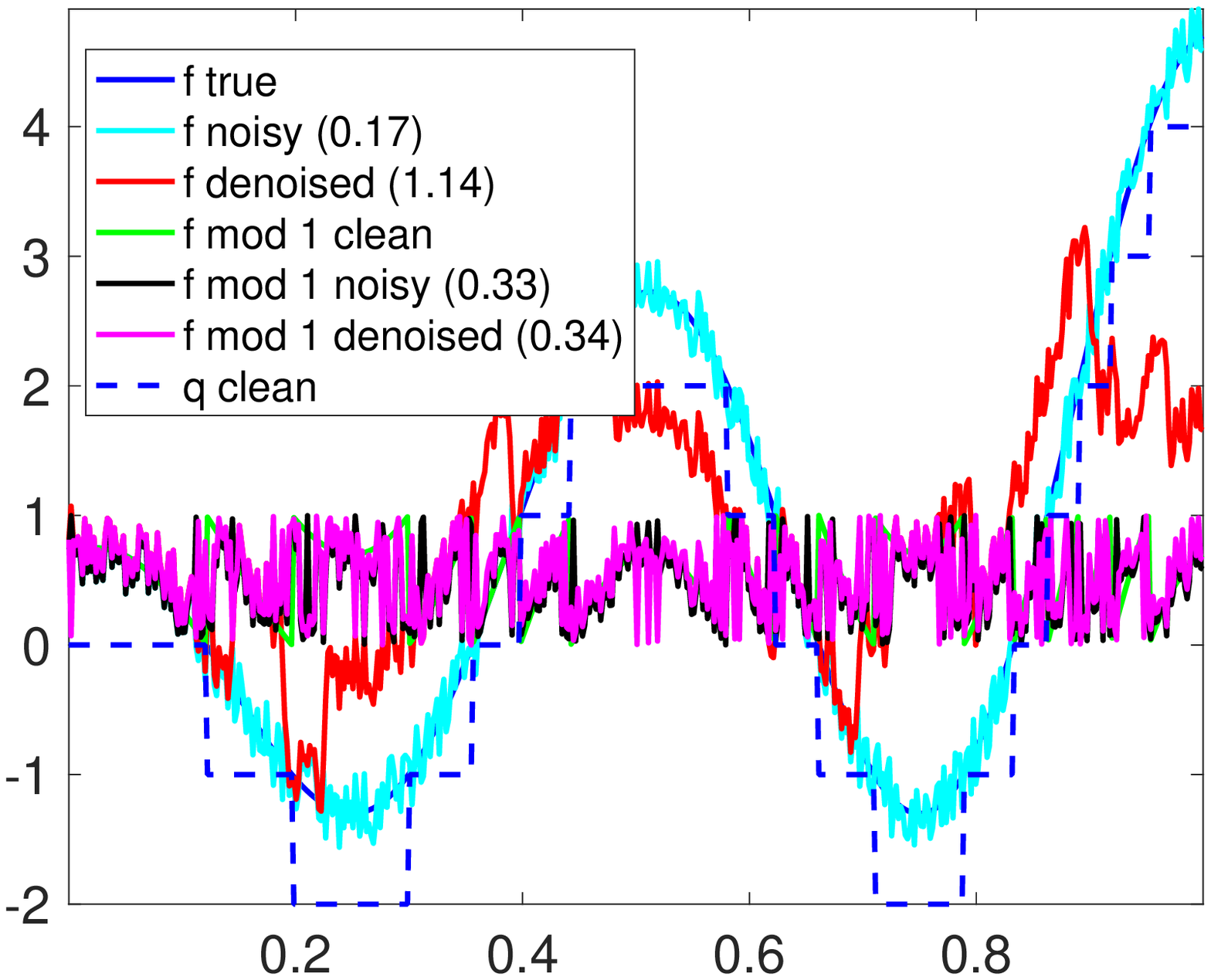} }
%
\subcaptionbox[]{  $\gamma=0.27$, \textbf{QCQP}
}[ 0.20\textwidth ]
{\includegraphics[width=0.20\textwidth] {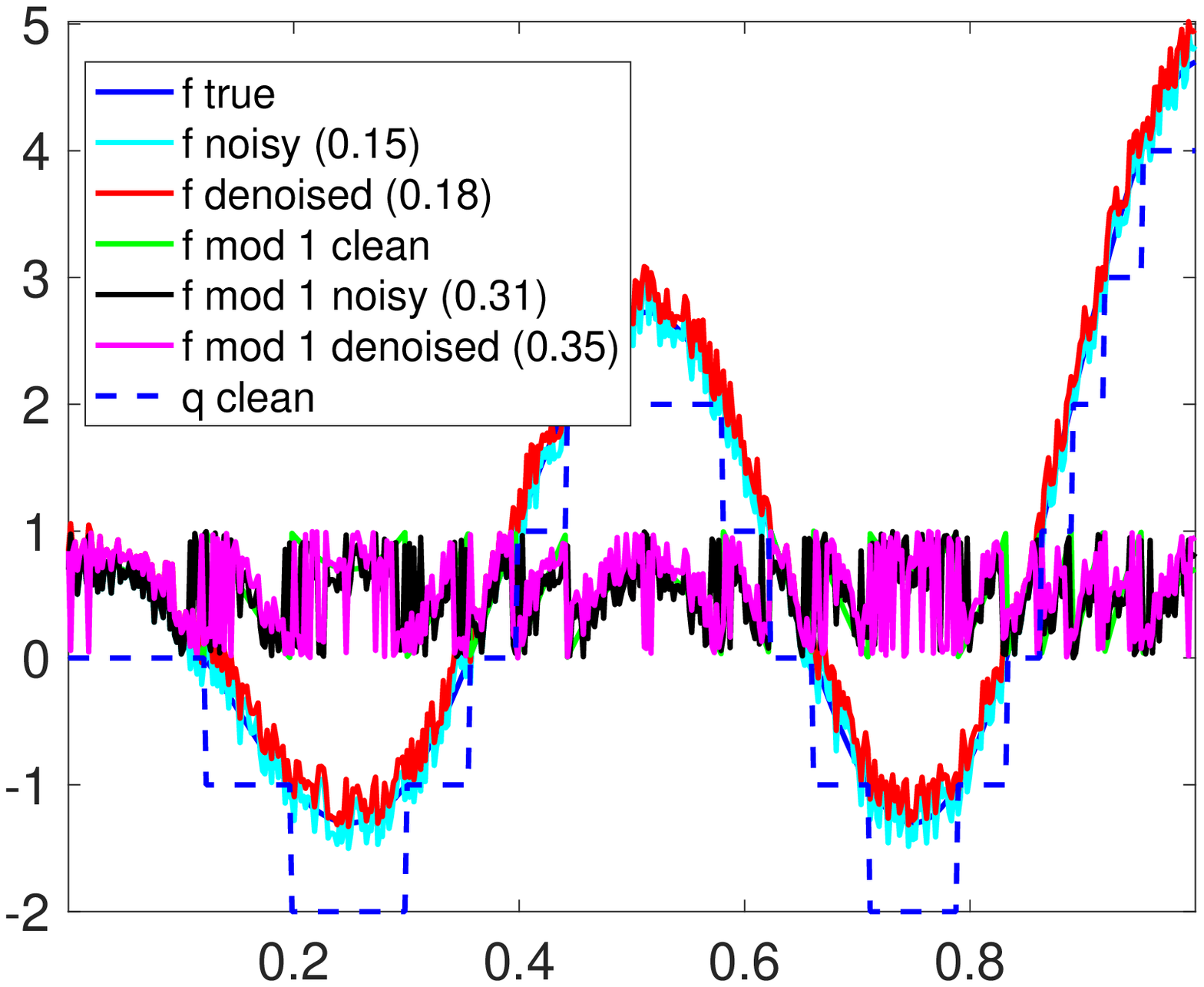} }
%
\subcaptionbox[]{  $\gamma=0.30$, \textbf{QCQP}
}[ 0.20\textwidth ]
{\includegraphics[width=0.20\textwidth] {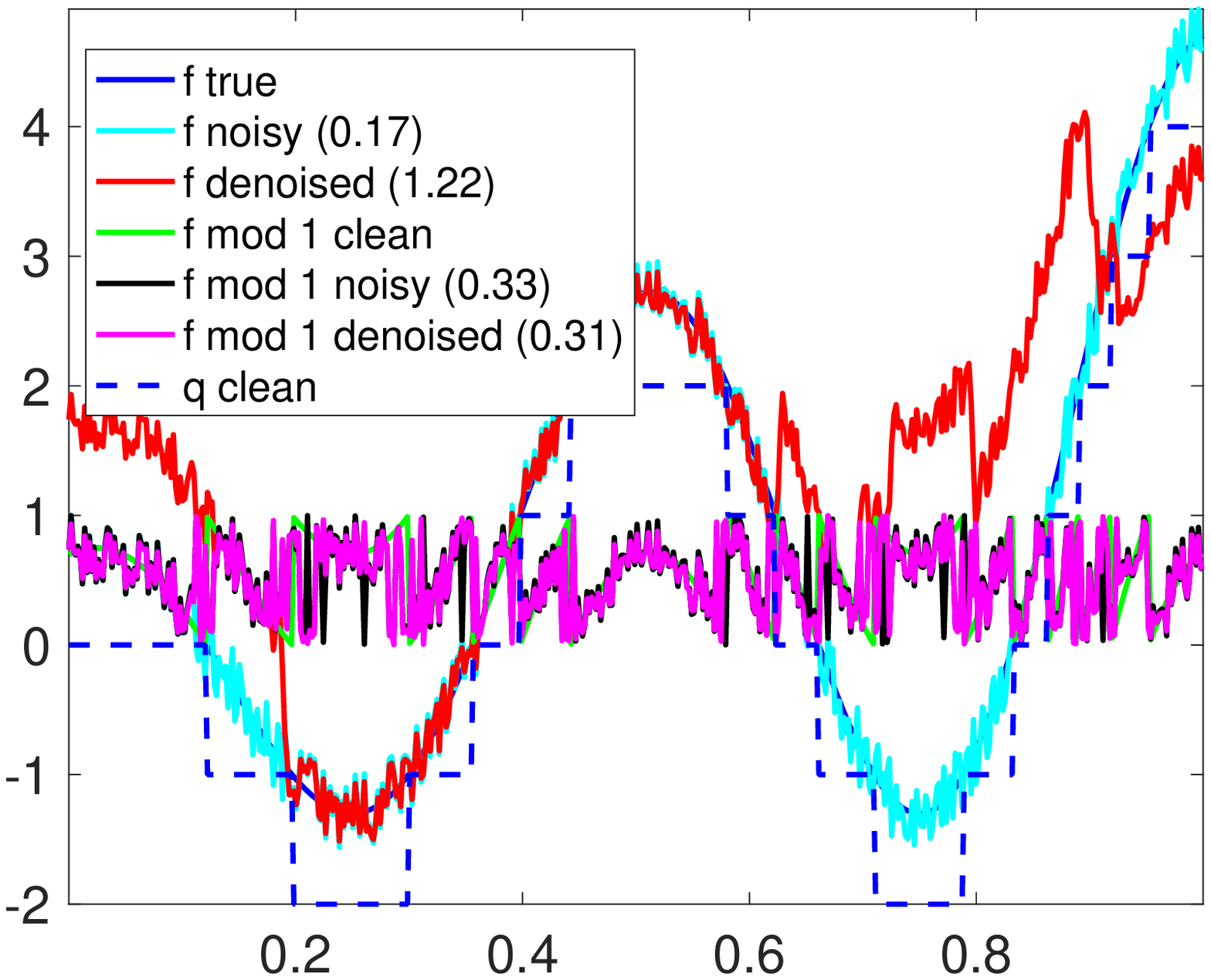} }
%
%
%
%
\subcaptionbox[]{  $\gamma=0.27$, \textbf{iQCQP} 
}[ 0.20\textwidth ]
{\includegraphics[width=0.20\textwidth] {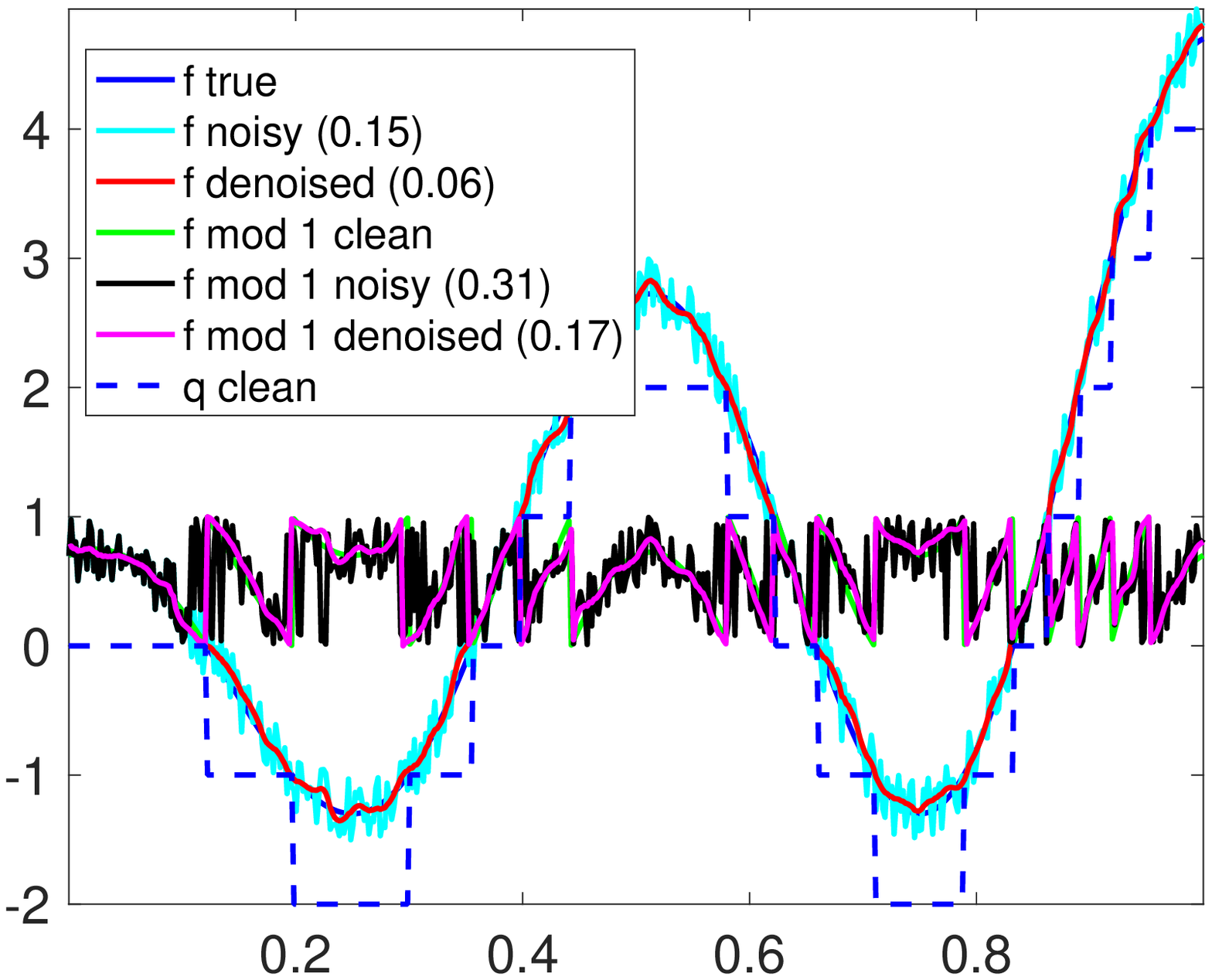} }
\subcaptionbox[]{  $\gamma=0.30$, \textbf{iQCQP}  
}[ 0.20\textwidth ]
{\includegraphics[width=0.20\textwidth] {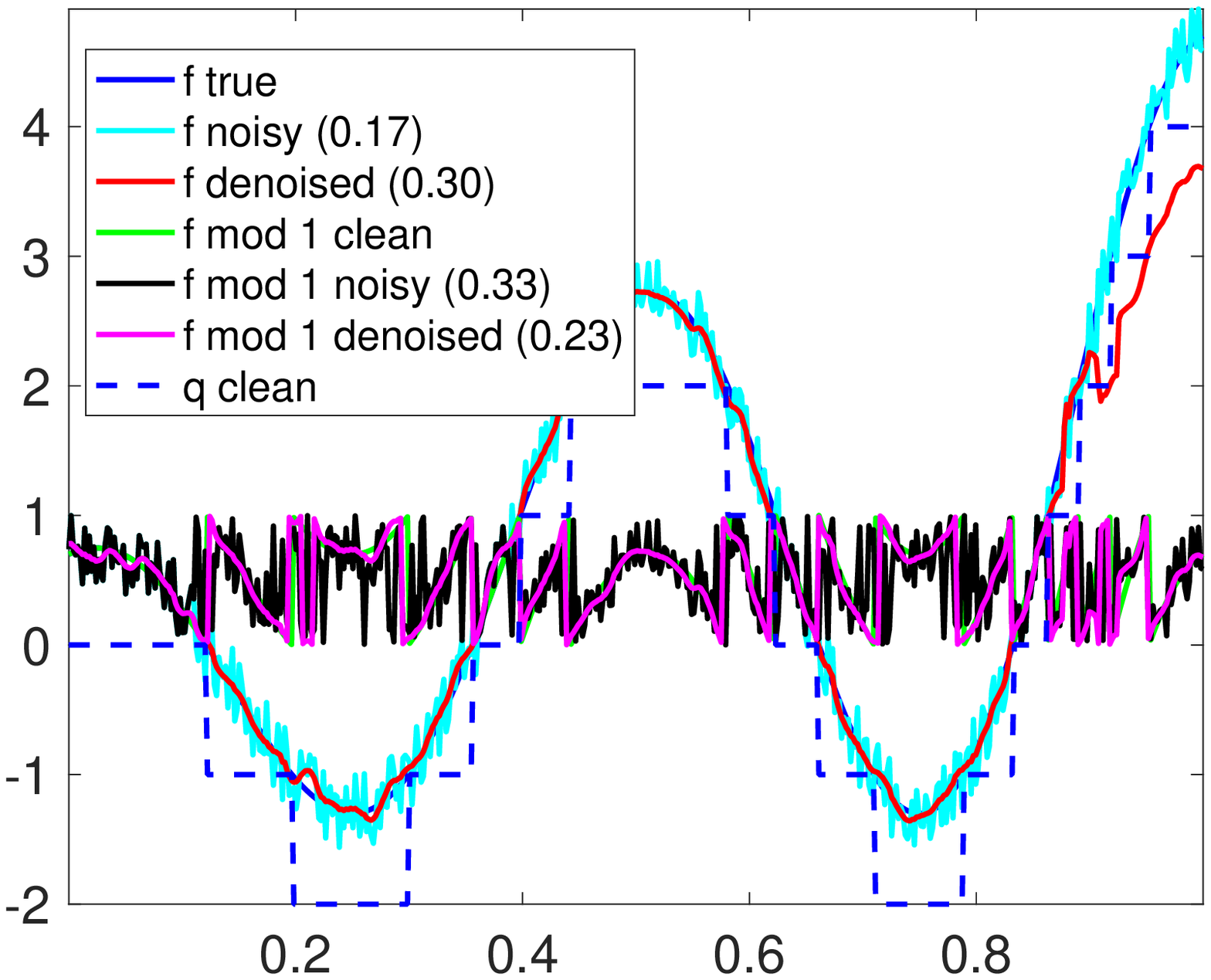} }
%
%
%
\vspace{-2mm}
\captionsetup{width=1\linewidth}
\caption[Short Caption]{Denoised instances under the Uniform noise model, for \textbf{OLS}, \textbf{QCQP} and \textbf{iQCQP},  as we increase the noise level $\gamma$. We keep fixed the parameters $n=500$, $k=2$, $\lambda= 0.1$.
}
\vspace{-3mm}
\label{fig:instances_f1_Bounded_main}
\end{figure}

\begin{figure*}
\centering
\subcaptionbox[]{ $k=2$, $\lambda= 0.1$
}[ 0.24\textwidth ]
{\includegraphics[width=0.24\textwidth] {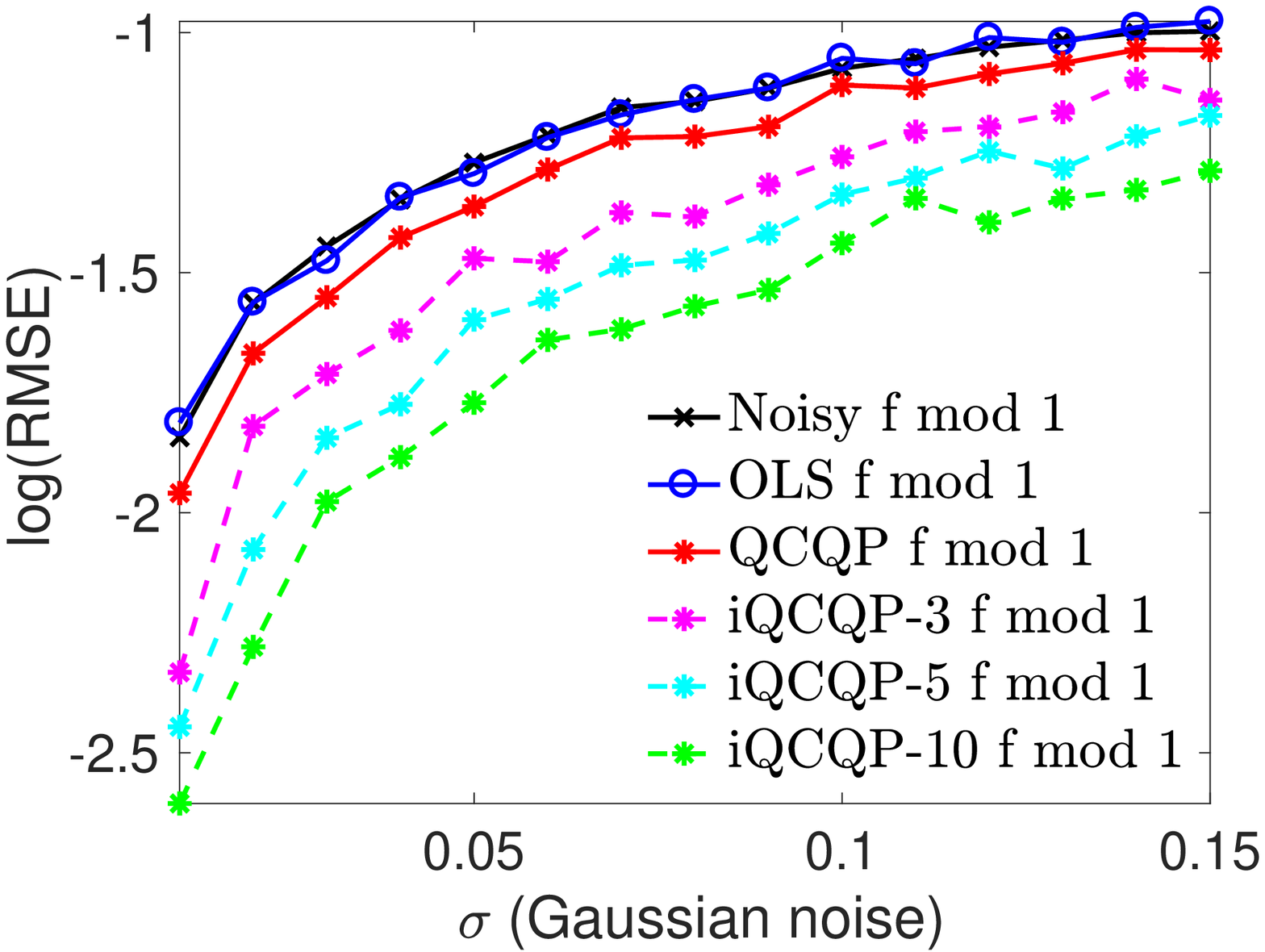} }
%
\subcaptionbox[]{ $k=3$, $\lambda= 0.1$
}[ 0.24\textwidth ]
{\includegraphics[width=0.24\textwidth] {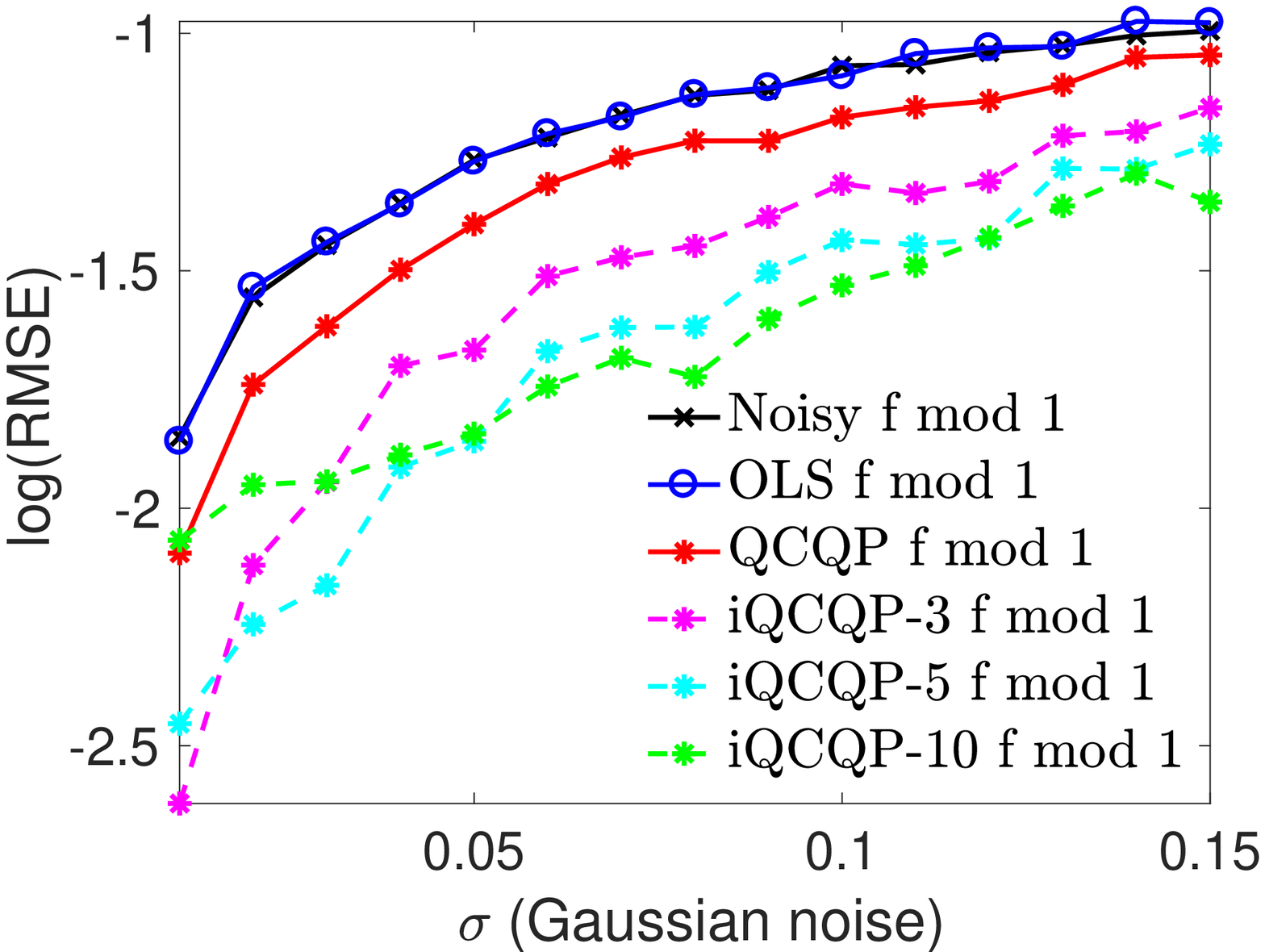} }
%
\subcaptionbox[]{ $k=2$, $\lambda= 0.3$
}[ 0.24\textwidth ]
{\includegraphics[width=0.24\textwidth] {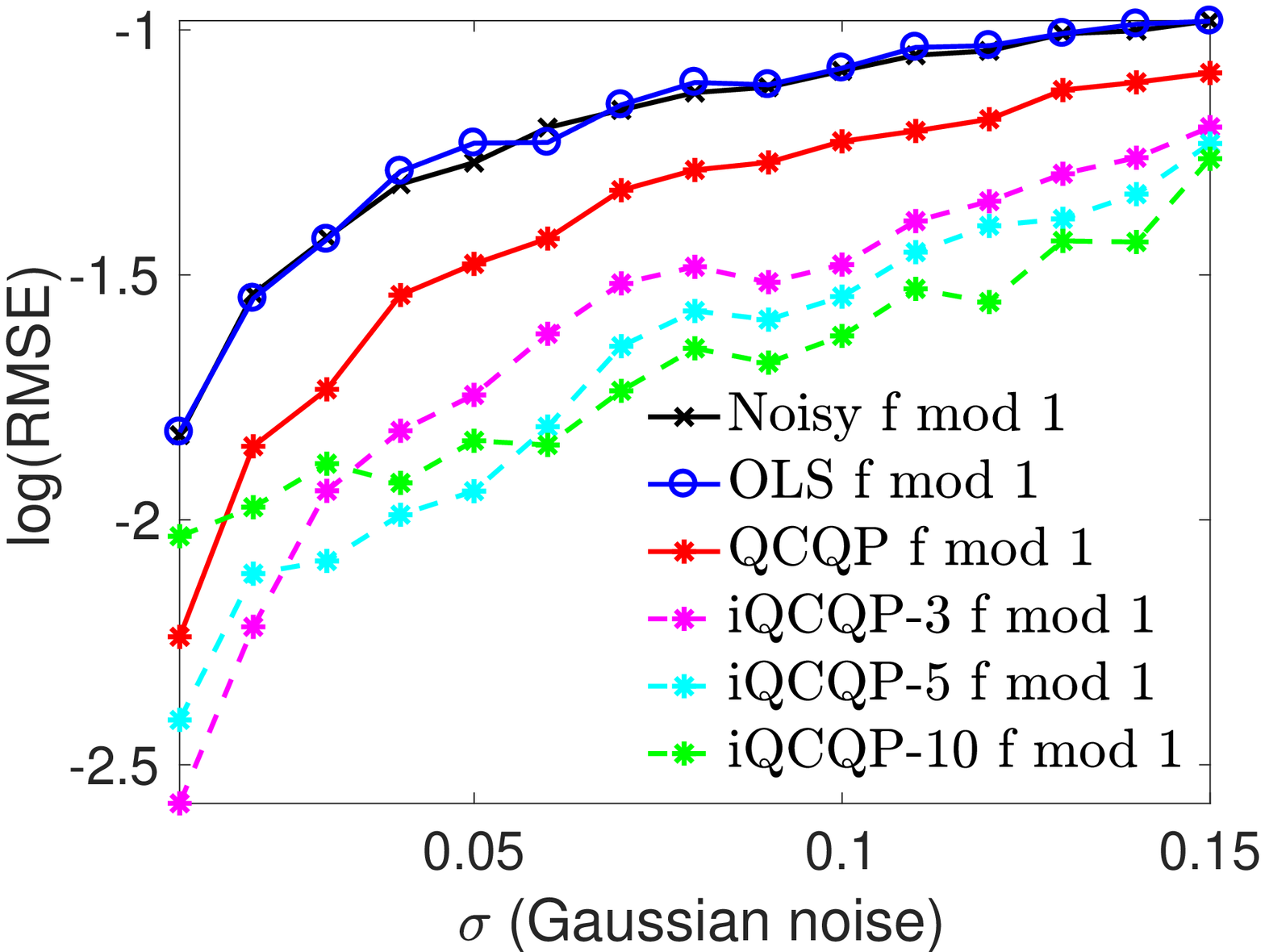} }
%
\subcaptionbox[]{$k=2$, $\lambda= 0.5$
}[ 0.24\textwidth]
{\includegraphics[width=0.24\textwidth] {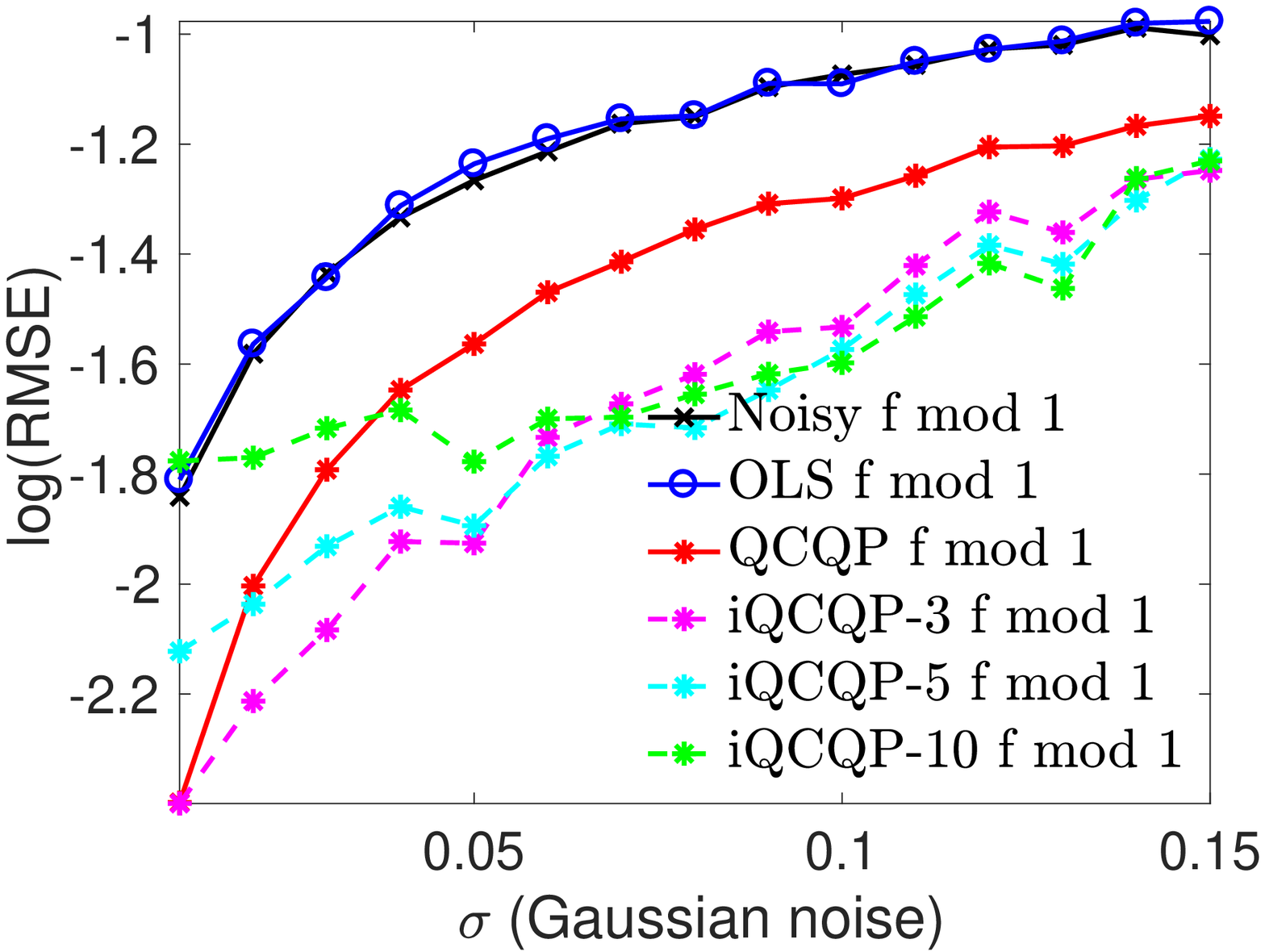} }
\vspace{-3mm}
\captionsetup{width=0.98\linewidth}
\caption[Short Caption]{Recovery errors for the denoised $f$ \hspace{-2mm} mod 1 samples, for the Gaussian noise model (20 trials). 
}
\label{fig:Sims_f1_Gaussian_fmod1}
\end{figure*}

\begin{figure*}
\centering
\subcaptionbox[]{ $k=2$, $\lambda= 0.1$
}[ 0.24\textwidth ]
{\includegraphics[width=0.24\textwidth] {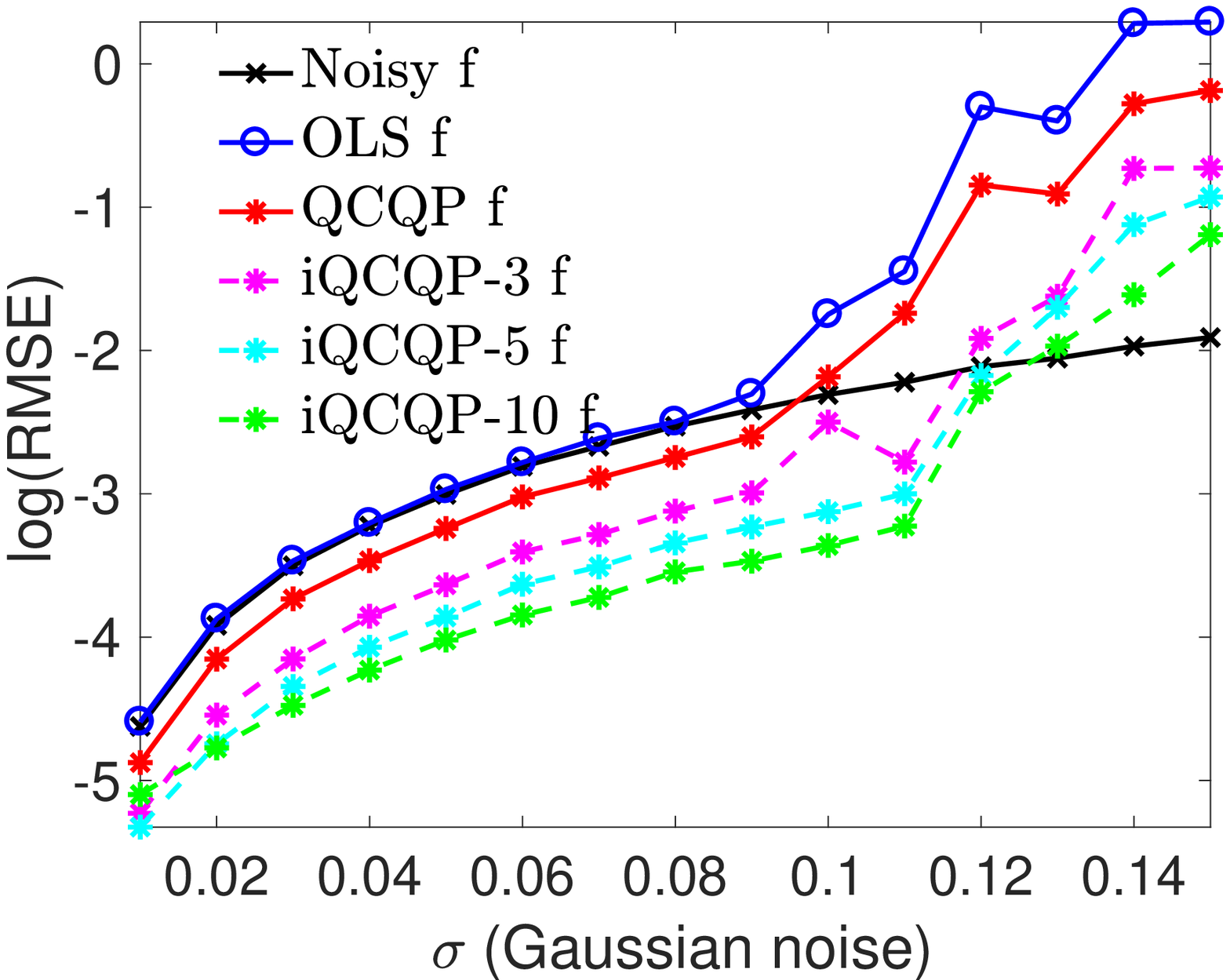} }
%
\subcaptionbox[]{ $k=3$, $\lambda= 0.1$
}[ 0.24\textwidth ]
{\includegraphics[width=0.24\textwidth] {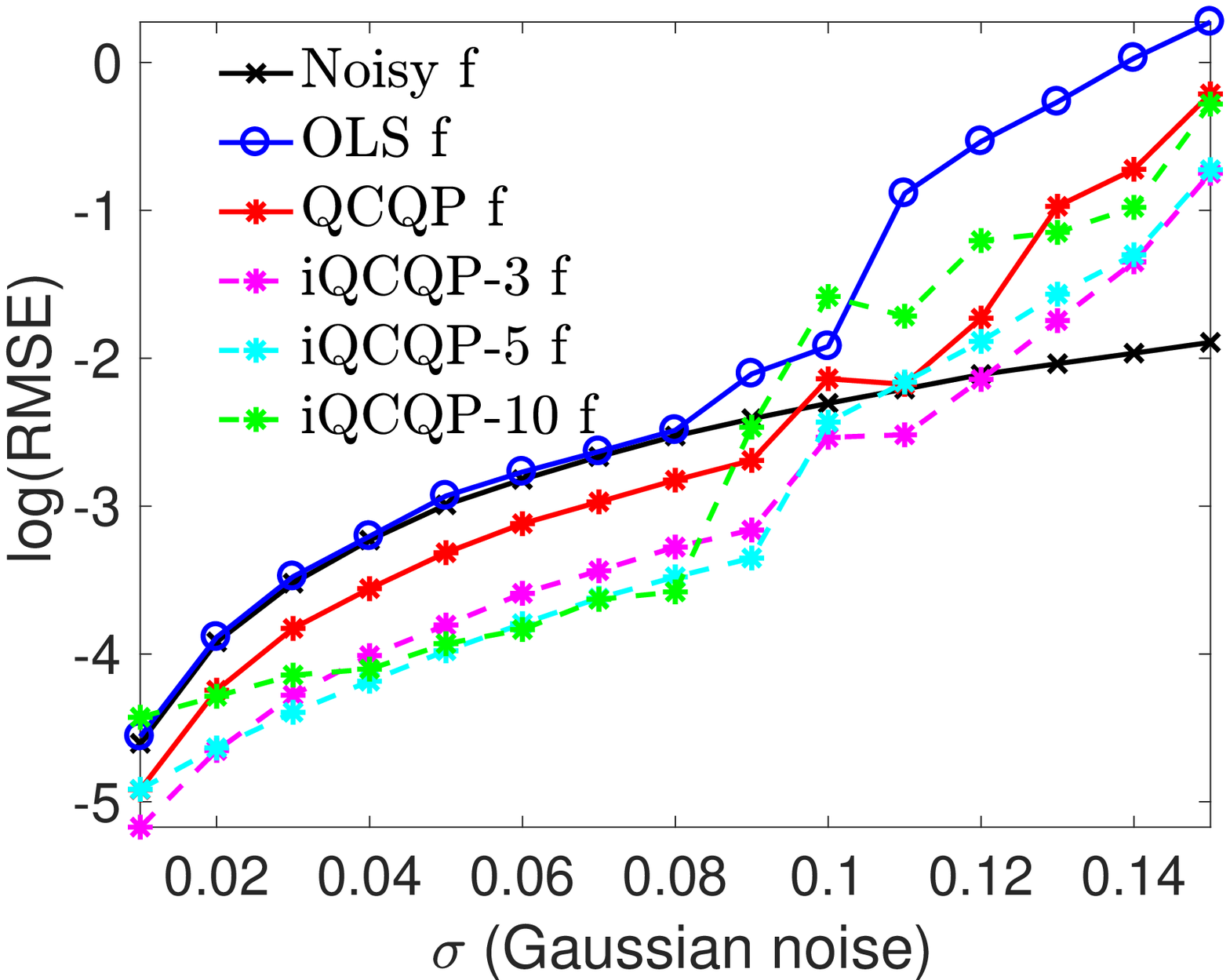} }
\subcaptionbox[]{ $k=2$, $\lambda= 0.3$
}[ 0.24\textwidth ]
{\includegraphics[width=0.24\textwidth] {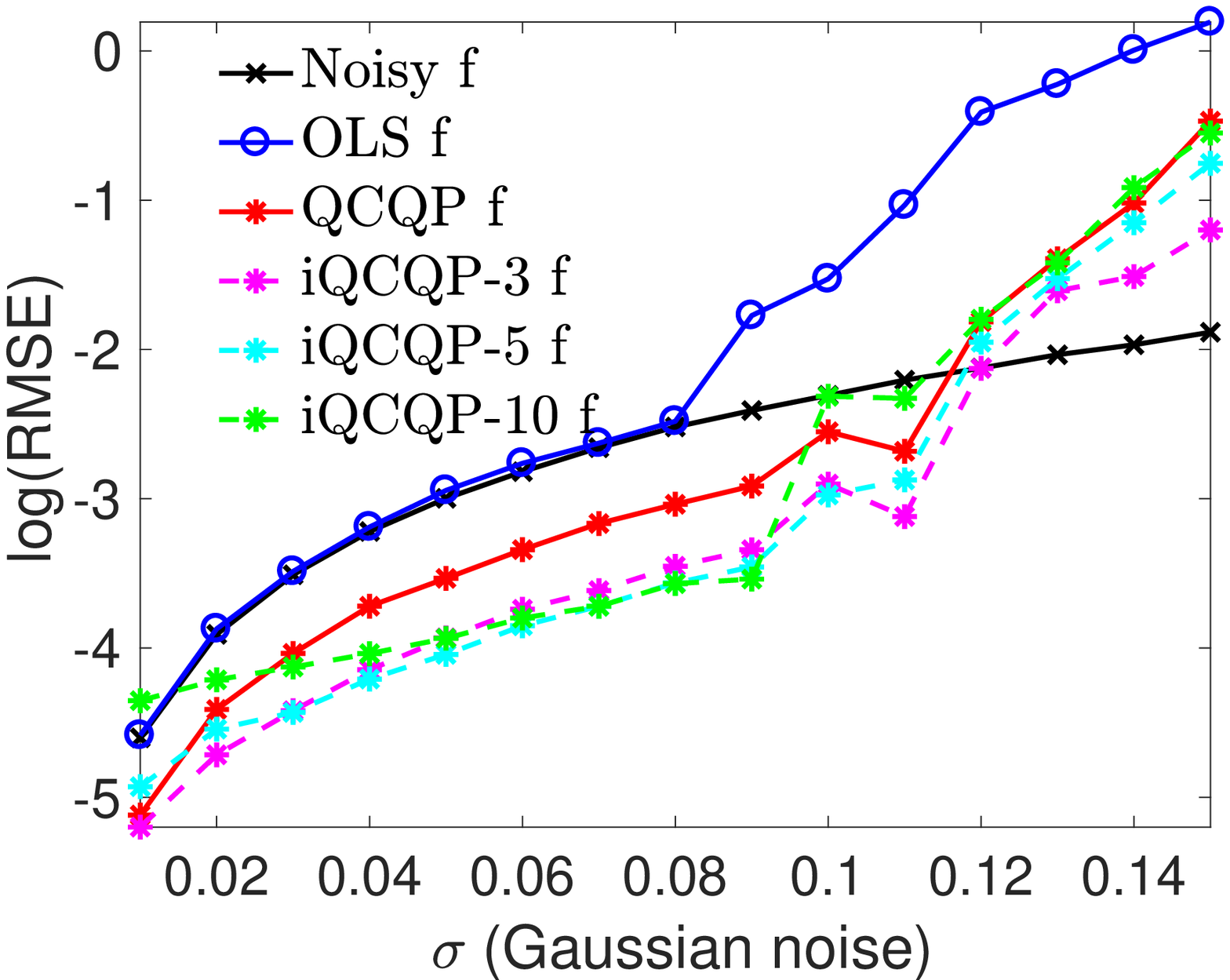} }
\subcaptionbox[]{ $k=2$, $\lambda= 0.5$
}[ 0.24\textwidth ]
{\includegraphics[width=0.24\textwidth] {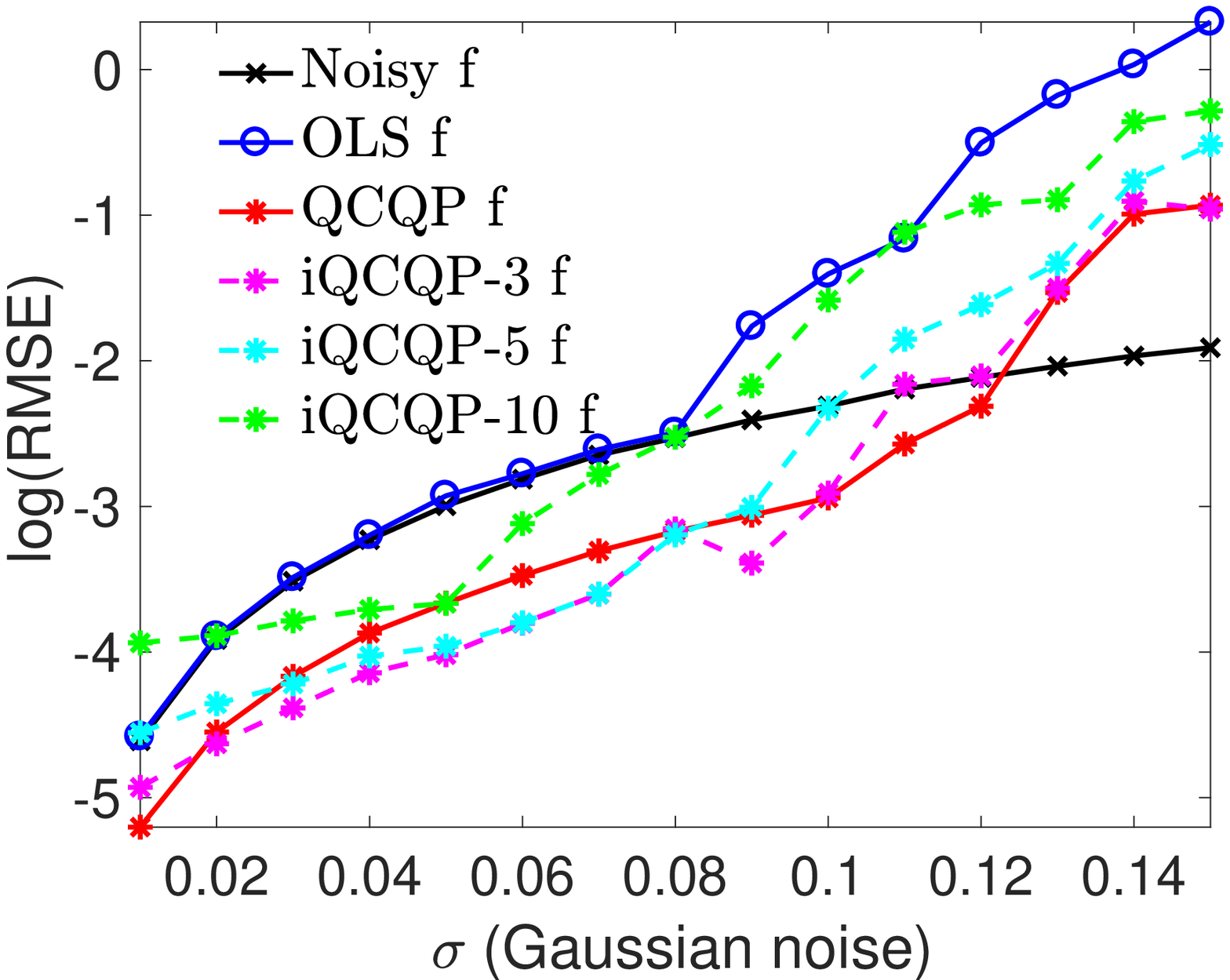} }
\vspace{-3mm}
\captionsetup{width=0.98\linewidth}
\caption[Short Caption]{Recovery errors for the final estimated samples of $f$, under the Gaussian noise model (20 trials). 
}
\label{fig:Sims_f1_Gaussian_f}
\end{figure*}

\begin{figure*}
\centering
\subcaptionbox[]{ $k=2$, $\lambda= 0.3$, $\gamma = 0.01$
}[ 0.24\textwidth ]
{\includegraphics[width=0.24\textwidth] {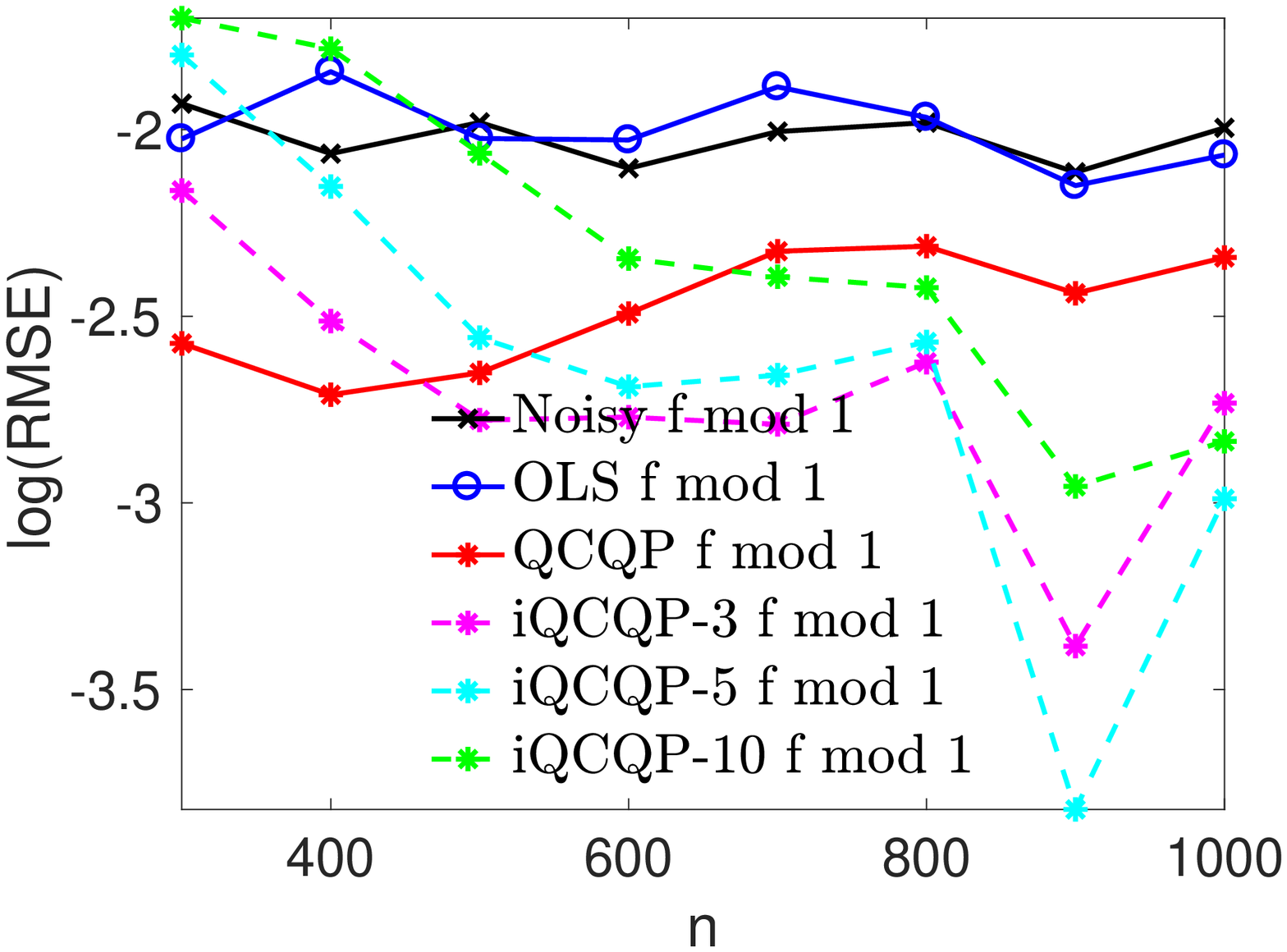} }
%
\subcaptionbox[]{ $k=3$, $\lambda= 0.3$, $\gamma = 0.25$
}[ 0.24\textwidth ]
{\includegraphics[width=0.24\textwidth] {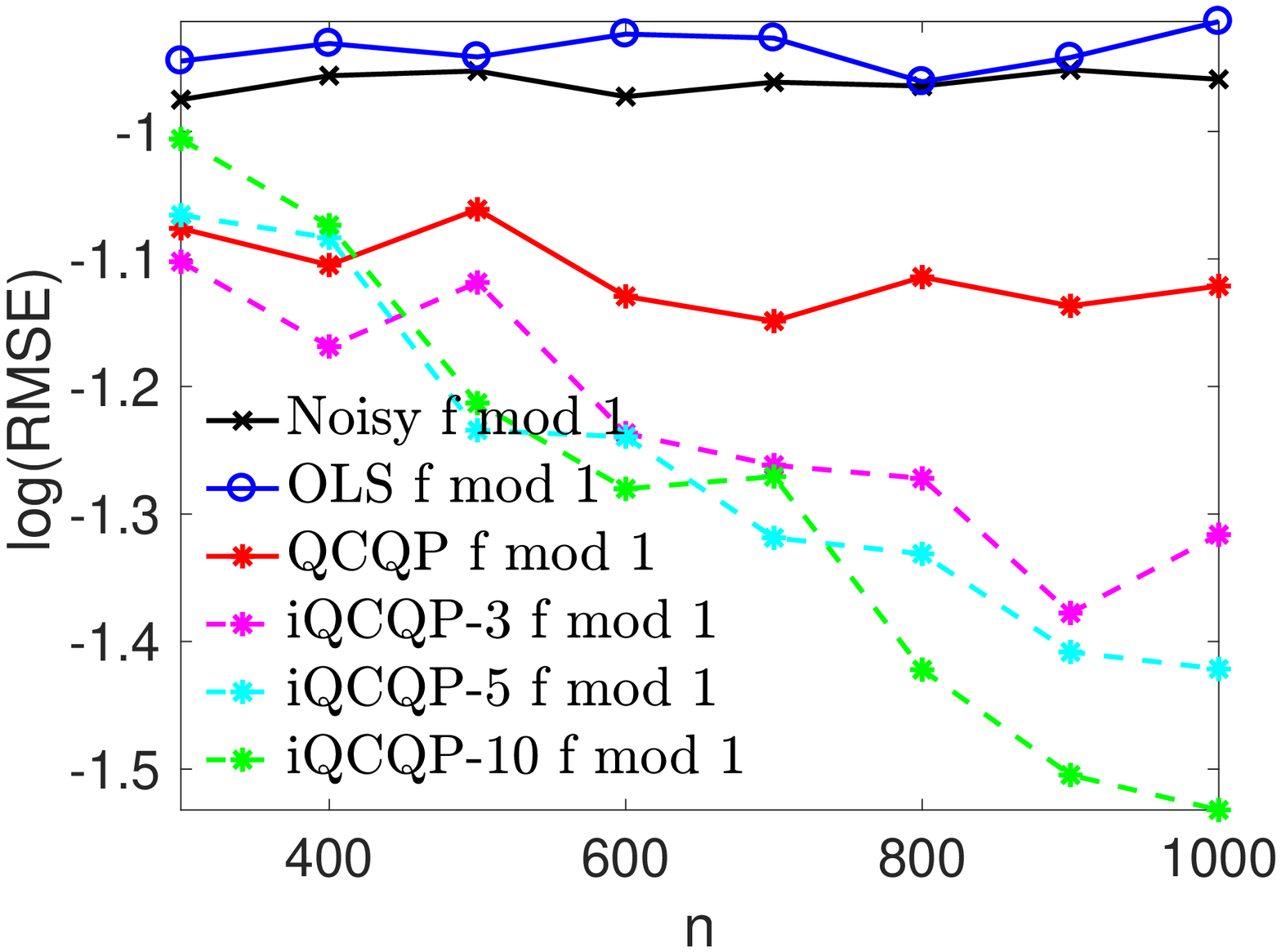} }
%
%
\subcaptionbox[]{ $k=2$, $\lambda= 0.3$, $\gamma = 0.01$
}[ 0.24\textwidth ]
{\includegraphics[width=0.24\textwidth] {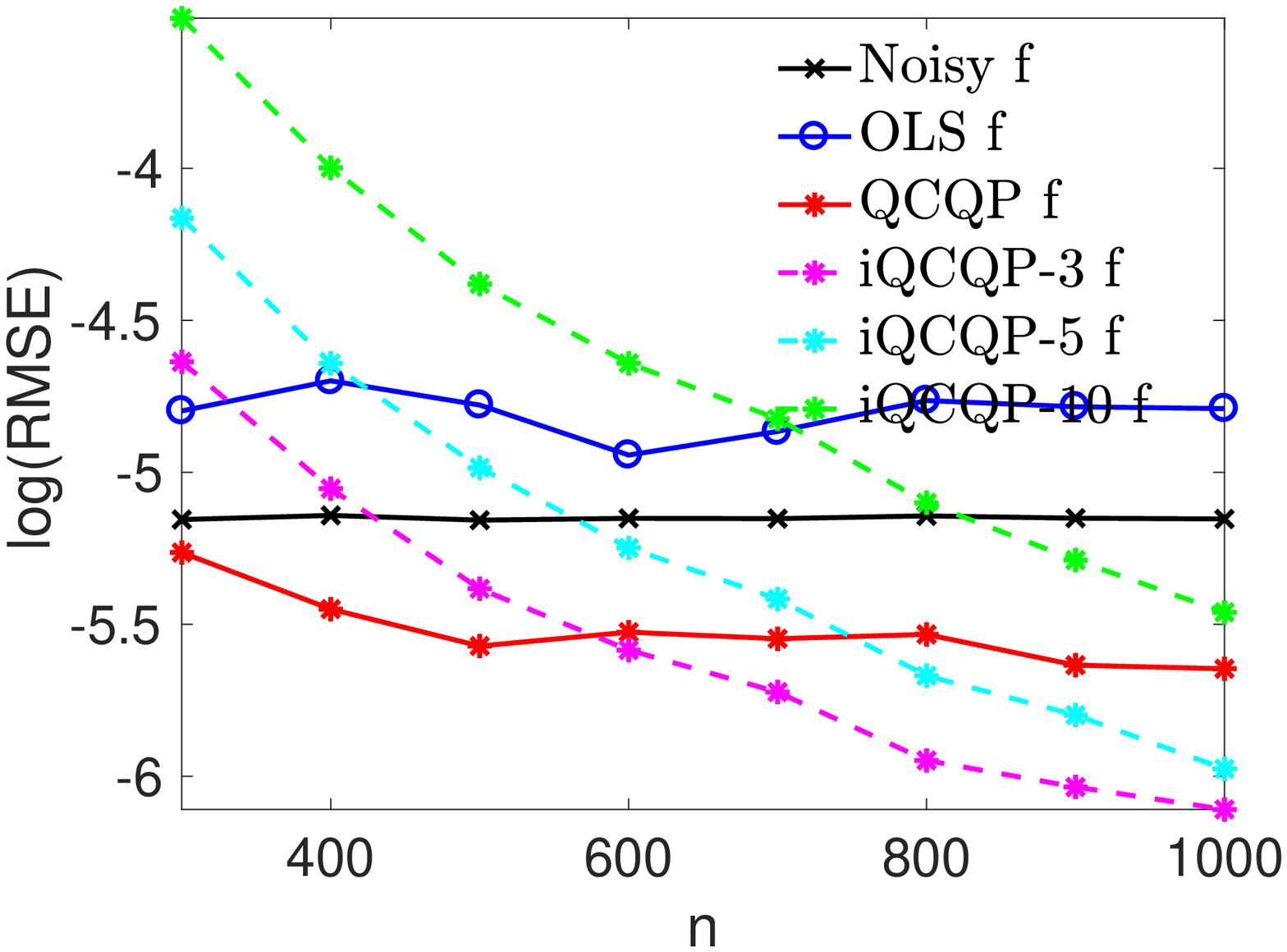} }
%
\subcaptionbox[]{ $k=3$, $\lambda= 0.3$, $\gamma = 0.25$
}[ 0.24\textwidth ]
{\includegraphics[width=0.24\textwidth] {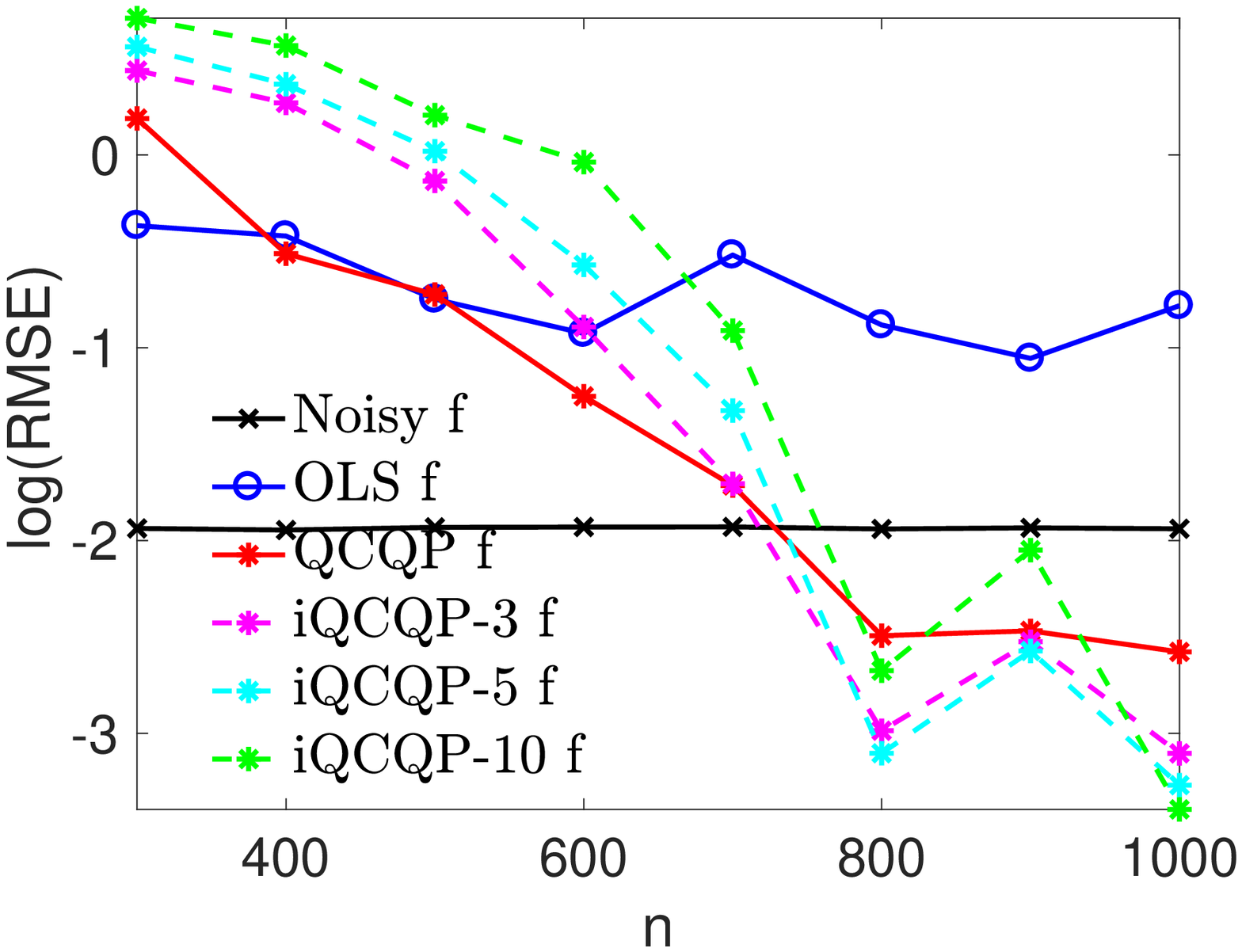} }
\vspace{-3mm}
\captionsetup{width=0.98\linewidth}
\caption[Short Caption]{Recovery errors for \textbf{OLS}, \textbf{QCQP}, and  \textbf{iQCQP} 
as a function of $n$ (number of samples), for both the $f$ mod 1 samples (leftmost two plots) and the final $f$ estimates (rightmost two plots) under the Uniform noise model, for different values of $k$, $\lambda$ and $\gamma$. Results are averaged over 20 runs.
}
\label{fig:Sims_f1_Bounded_ScanID2_ffmod1}
\end{figure*}

%
%
%
Figure \ref{fig:instances_f1_Bounded_main} shows several denoising instances as we increase the noise level 
in the Uniform noise model ($\gamma \in \{0.27, 0.30\}$). Notice that \textbf{OLS} starts failing at $\gamma = 0.27$, 
while \textbf{QCQP} still estimates the samples of $f$ well. Interestingly, \textbf{iQCQP} performs quite well, 
even for $\gamma = 0.30$ (where \textbf{QCQP} starts failing) and produces highly smooth, and accurate estimates.
It would be interesting to investigate the properties of \textbf{iQCQP} in future work.
Analogous results for the Gaussian model are shown in the appendix.  
Figures \ref{fig:Sims_f1_Bounded_fmod1}, \ref{fig:Sims_f1_Bounded_f} plot RMSE (on a log scale) for denoised 
$f$ mod 1 and $f$ samples versus the noise level, for the Uniform noise model. They illustrate the importance of 
the choice of the regularization parameters $\lambda, k$. If $\lambda$ is too small (eg., $\lambda = 0.03$), 
then \textbf{QCQP} has negligible improvement in performance, and sometimes also has worse RMSE than the raw noisy samples. 
However, for a larger $\lambda$ ($\lambda \in \set{0.3,0.5}$), \textbf{QCQP} has a strictly smaller error than $\textbf{OLS}$ and the raw noisy 
samples. Interestingly, \textbf{iQCQP} typically performs very well, even for $\lambda = 0.03$.  
%
%
%
%
Figures \ref{fig:Sims_f1_Gaussian_fmod1}, \ref{fig:Sims_f1_Gaussian_f} show similar plots for the Gaussian noise model.
Figure \ref{fig:Sims_f1_Bounded_ScanID2_ffmod1} plots the RMSE (on a log scale) for both the denoised $f$ mod 1 samples, and samples of $f$, 
versus $n$ (for Uniform noise model). Observe that for large enough $n$, \textbf{QCQP} shows strictly smaller RMSE than both the initial input noisy data, and \textbf{OLS}. We also remark that \textbf{iQCQP} typically has superior performance to \textbf{QCQP} except for small values of $n$.
%
%
%
We defer to the appendix  a comparison with the recent algorithm of Bhandari et al. \cite{bhandari17}.

\vspace{-1mm}
\vspace{-3mm}

%
%
%

%
%
%
%
 %
 %
 %
 %
\begin{figure*}
\centering
\subcaptionbox[]{ $k=2$, $\lambda= 0.03$
}[ 0.24\textwidth ]
{\includegraphics[width=0.24\textwidth] {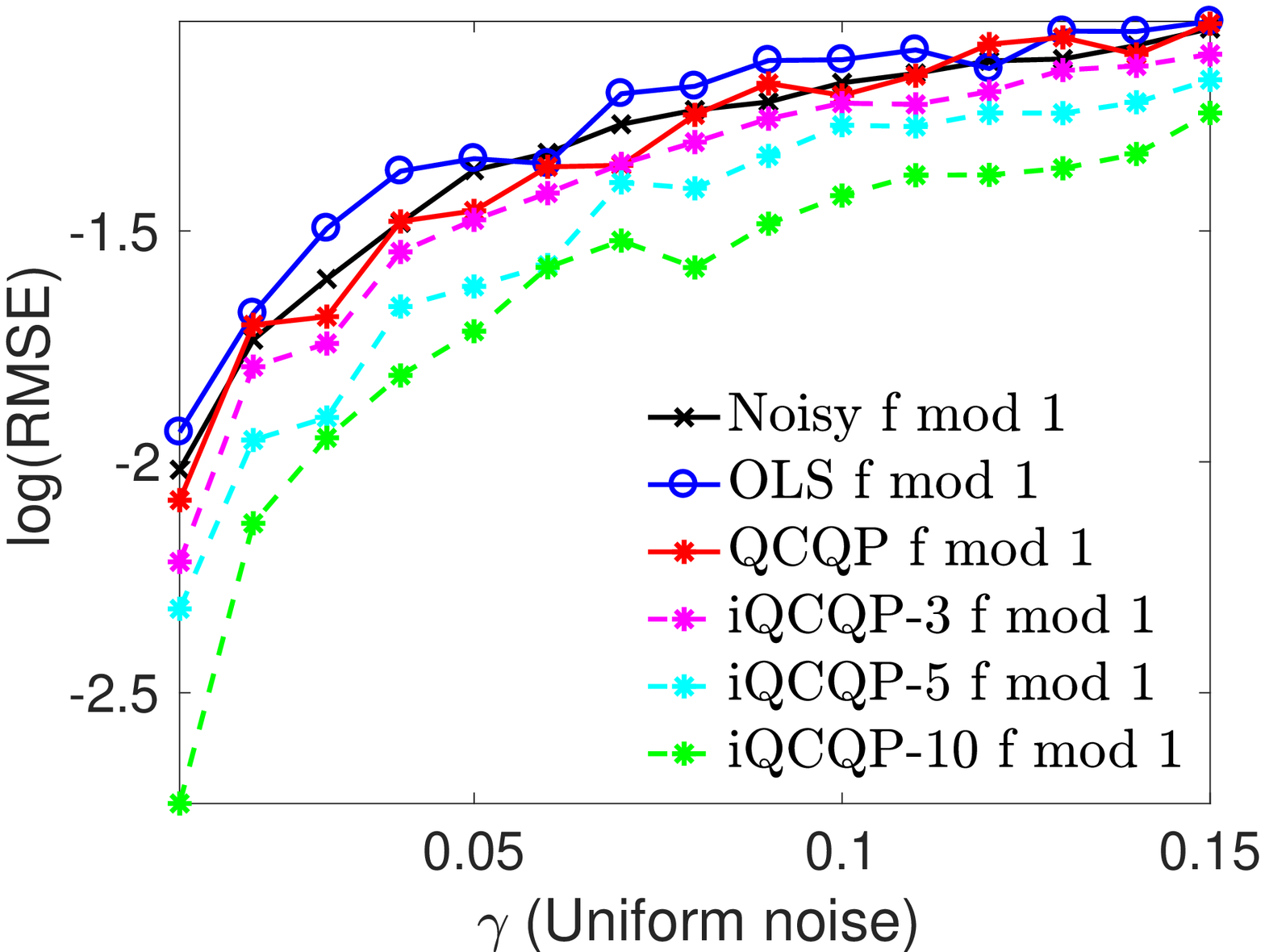} }
%
\subcaptionbox[]{ $k=3$, $\lambda= 0.03$
}[ 0.24\textwidth ]
{\includegraphics[width=0.24\textwidth] {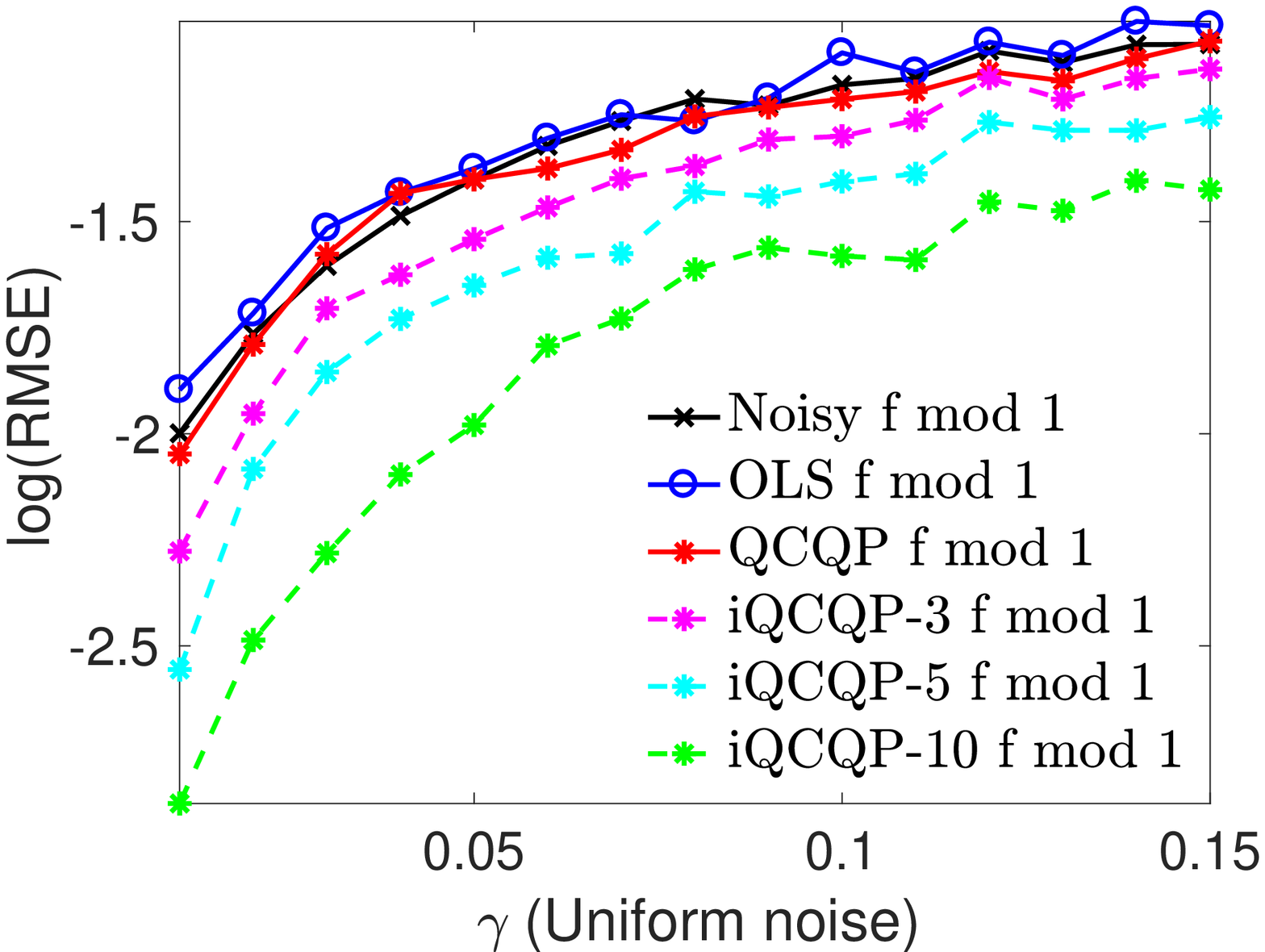} }
\subcaptionbox[]{ $k=2$, $\lambda= 0.3$
}[ 0.24\textwidth ]
{\includegraphics[width=0.24\textwidth] {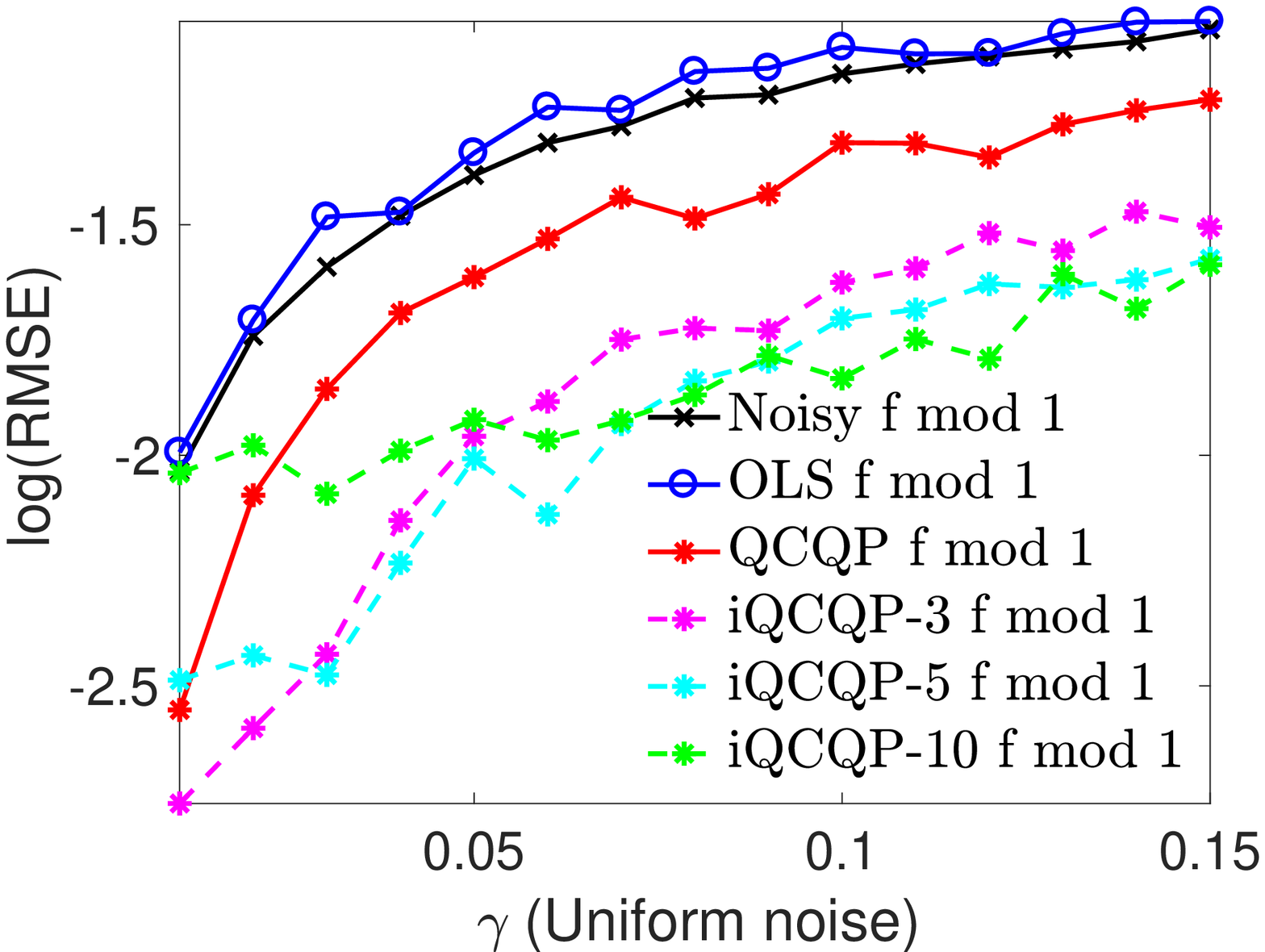} }
\subcaptionbox[]{ $k=2$, $\lambda= 0.5$
}[ 0.24\textwidth ]
{\includegraphics[width=0.24\textwidth] {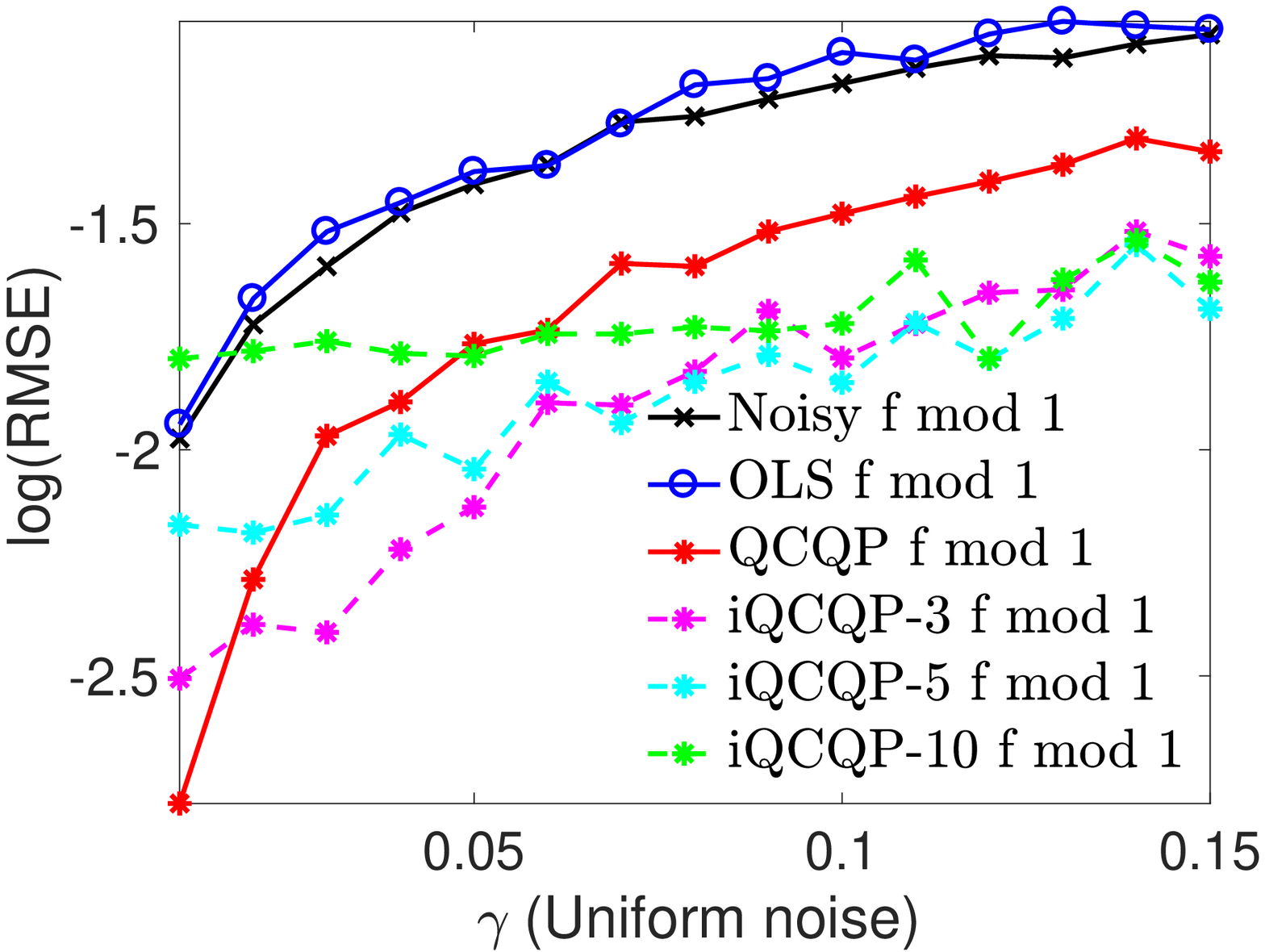} }
%
\vspace{-3mm}
\captionsetup{width=1\linewidth}
\caption[Short Caption]{Numerical experiments for \textbf{OLS}, \textbf{QCQP}, and  \textbf{iQCQP} (with 3,5, and 10 iterations) showing the recovery RMSE error (on a log scale) when denoising the $f$ mod 1 samples, under the Uniform noise model. Results are averaged over 20 trials.
}
\label{fig:Sims_f1_Bounded_fmod1}
\end{figure*}
\begin{figure*}
\centering
\subcaptionbox[]{ $k=2$, $\lambda= 0.03$
}[ 0.24\textwidth ]
{\includegraphics[width=0.24\textwidth] {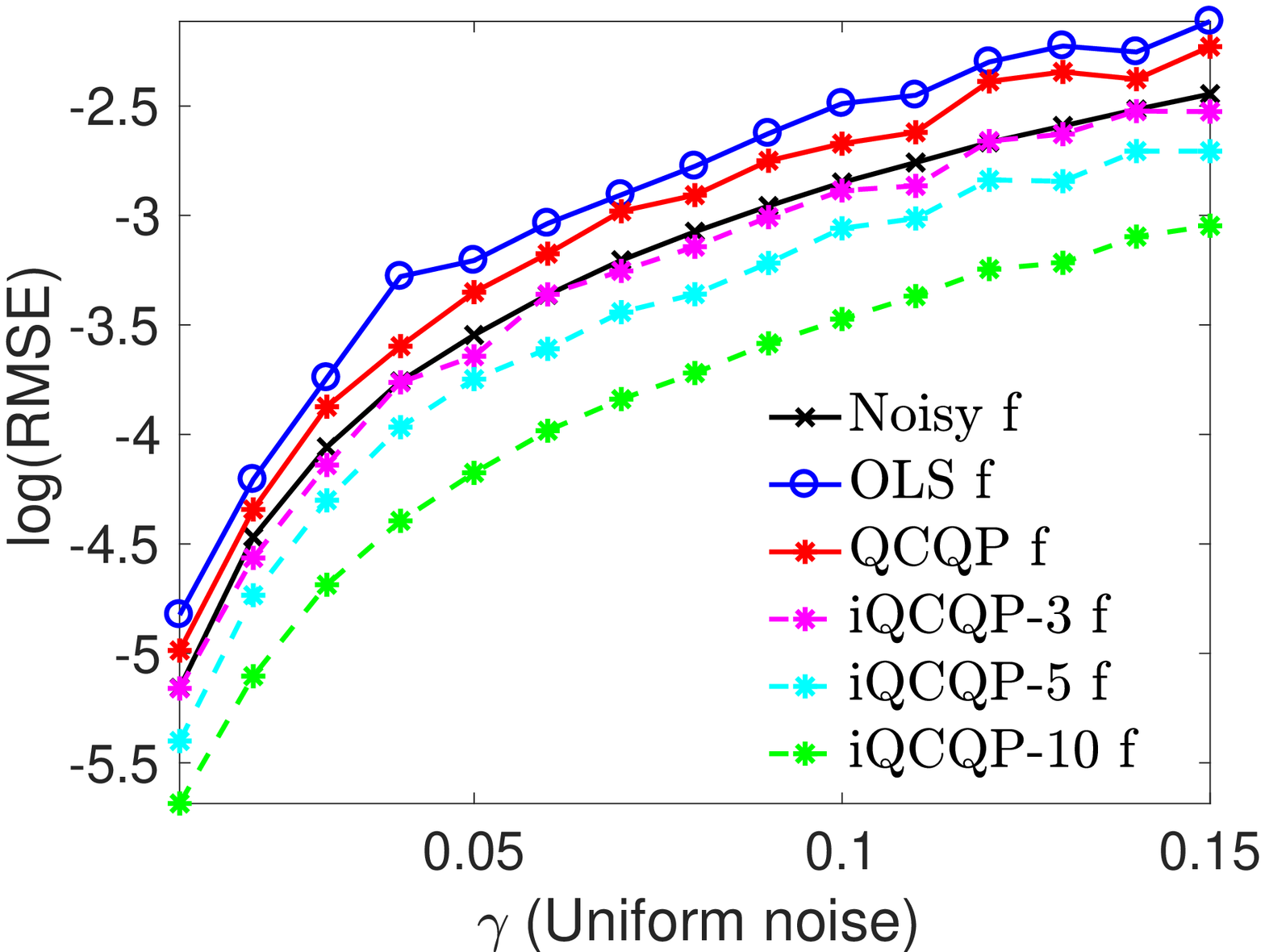} }
%
\subcaptionbox[]{ $k=3$, $\lambda= 0.03$
}[ 0.24\textwidth ]
{\includegraphics[width=0.24\textwidth] {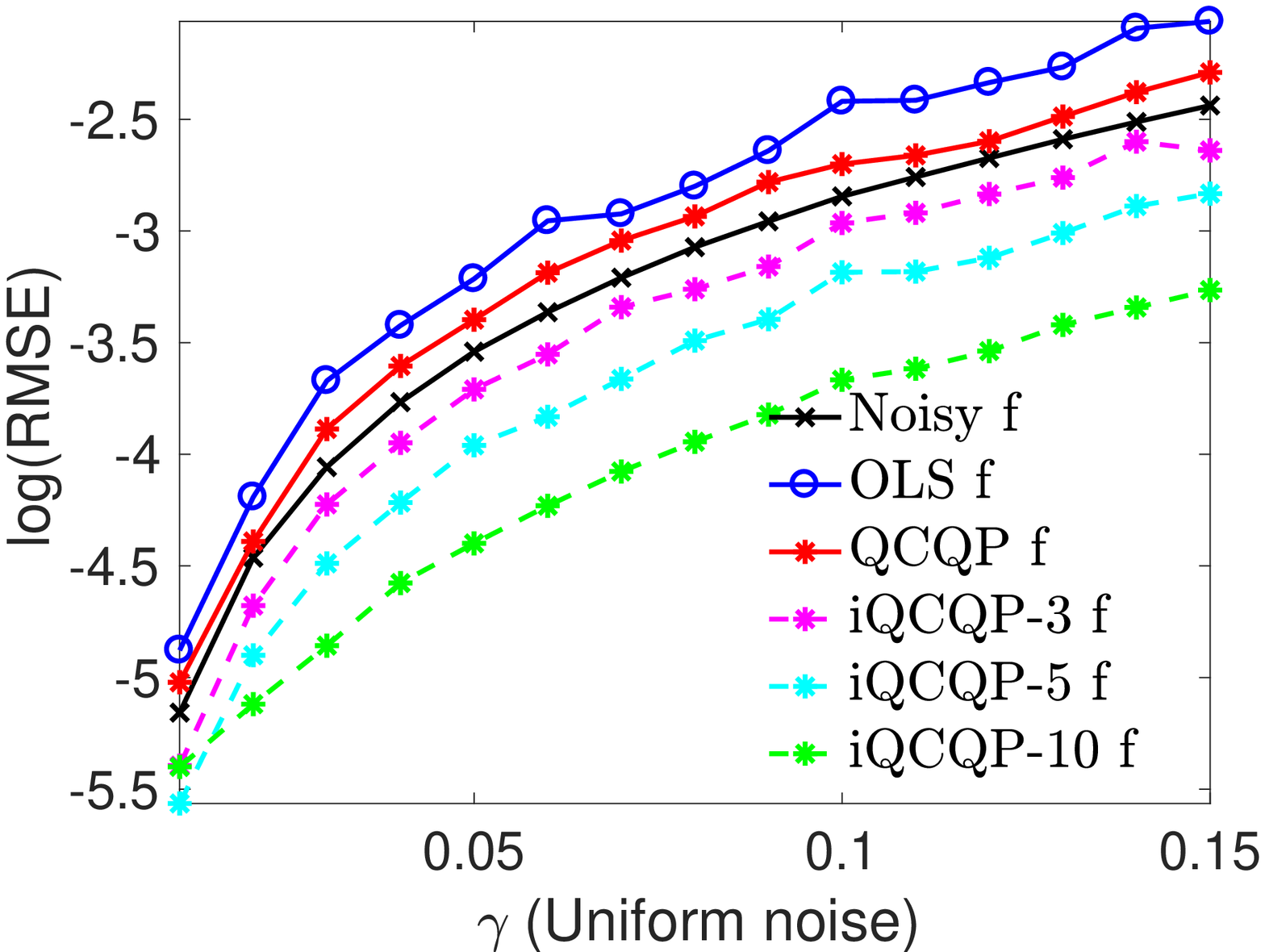} }
\hspace{0.01\textwidth} 
%
%
\subcaptionbox[]{ $k=2$, $\lambda= 0.3$
}[ 0.24\textwidth ]
{\includegraphics[width=0.24\textwidth] {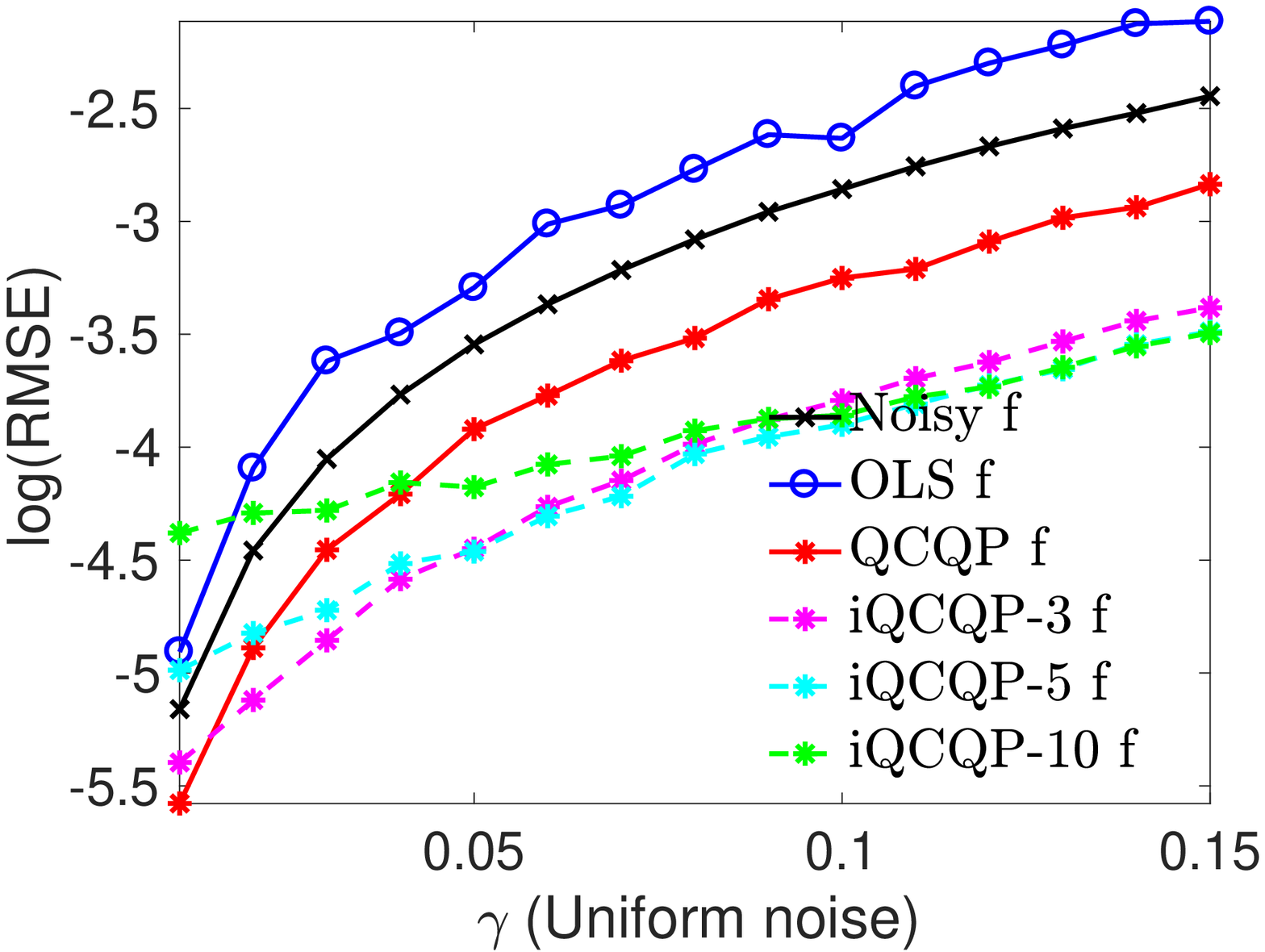} }
%
\subcaptionbox[]{ $k=2$, $\lambda= 0.5$
}[ 0.24\textwidth ]
{\includegraphics[width=0.24\textwidth] {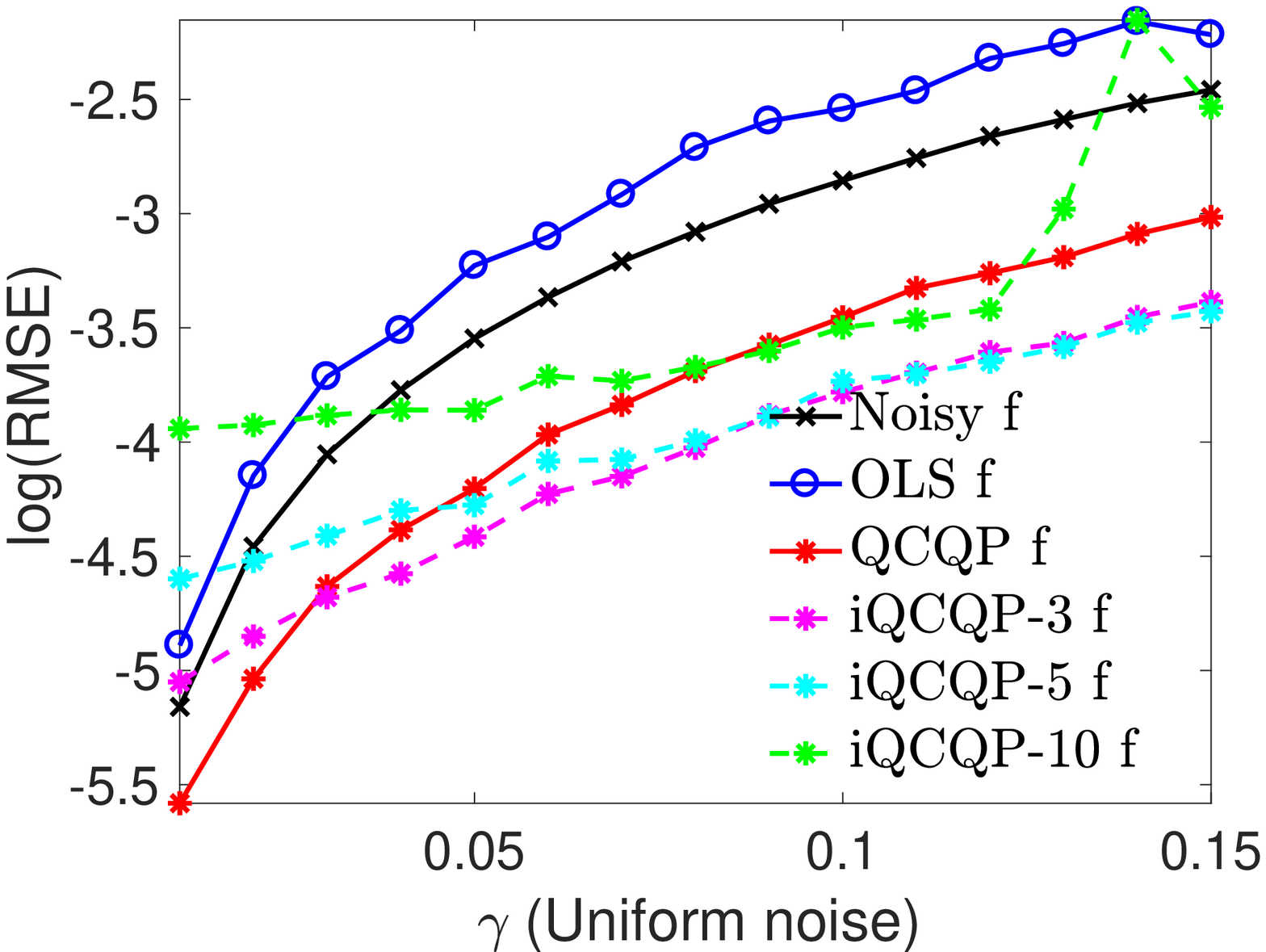} }
%
\vspace{-3mm}
\captionsetup{width=0.98\linewidth}
\caption[Short Caption]{Recovery errors for the final estimated samples of $f$, under the Uniform noise model (20 trials).
}
\label{fig:Sims_f1_Bounded_f}
\end{figure*}
%
%
%
%


%
%
%


 \vspace{-2mm}
\vspace{-4mm}
\section{Concluding remarks} \label{sec:conclusion}
\vspace{-3mm}
%
There are several possible directions for future work. One would be to better understand the unwrapping stage of our approach, and potentially to explore a patch-based divide-and-conquer method that solves the problem locally and then integrates the local solutions (patches) into a globally consistent framework, in the spirit of existing methods from the group synchronization literature \cite{sync,syncRank}. 
One could also consider ``single-stage methods''  that directly output denoised estimates to the original real-valued samples. 
Other interesting approaches to analyze would be based on: (a) dynamic programming (discussed in Section \ref{sec:trust_reg_relax}), 
(b) Semi-definite programming (SDP) relaxation of \eqref{eq:orig_denoise_hard_1} and, (c) tools from optimization on manifolds \cite{manopt} (as $\calC_n$ is 
a manifold \cite{absil08}).
In ongoing work \cite{Cucu18}, we extend the results in this paper to Gaussian and Bernoulli-uniform 
noise models, and to the multivariate setting. By using the popular Manopt toolbox \cite{manopt}, we are able to solve instances of the two-dimensional phase
unwrapping problem with a million sample points in under $20$ seconds, on a personal laptop.

%
%
%
%
%
%
%
%



\vspace{-3mm}
\paragraph{Acknowledgements} 
We are grateful to Afonso Bandeira, Ayush Bhandari, Nicolas Boumal, Yuji Nakatsukasa and Joel Tropp  for insightful discussions. This work was supported by EPSRC grant EP/N510129/1 at the Alan Turing Institute.  


\bibliographystyle{plain}
\newpage
\bibliography{funcmod}

\newpage
\onecolumn
\appendix 
\setcounter{section}{0}
\setcounter{remark}{0}

\begin{centering}
\large \bf Supplementary Material : On denoising noisy modulo $1$ samples of a function.
\end{centering}

%
%
%
\begin{table}[tpb]
\begin{minipage}[b]{0.99\linewidth}
\begin{center}
\begin{tabular}{|c|c|c|}
\hline
Symbol & Description   \\
\hline
$f$ 		& unknown real-valued function    \\
$r$ 	    & clean $ f \mod 1$    \\
$q$     &  clean reminder $q = f-r $  \\
$y$ 		&  noisy $ f \mod 1$   \\
\hline
$ \vechtil $     &    clean signal in angular domain  \\
$ \vecz $ 	 	& noisy signal in angular domain    \\
$ \vecg $     &  free complex-valued variable  \\
\hline
$ \vechtilbar $     &  real-valued version of $\vechtil$   \\
$ \veczbar $    &  real-valued version of $\vecz$  \\
$\vecgbar $     &  real-valued version of  $\vecg $   \\
\hline
$L$ & $n \times n$ Laplacian matrix of graph $G$  \\
$\Hbar$ & $2n \times 2n $ block diagonal version of L \\
\hline
\end{tabular}
\end{center}
\end{minipage}
\vspace{-2mm}
\caption{Summary of frequently used symbols in the paper.}
\label{tab:methodAbbrev}
\end{table}
%
%

%
%
\section{Rewriting QCQP in real domain} \label{sec:qcqp_compl_to_real}
We first show that $\lambda \vecg^{*} L \vecg = \vecgbar^T \Hbar \vecgbar$. Indeed, 
\begin{eqnarray}
\vecgbar^T \Hbar \vecgbar
&=& (\real(\vecg)^T \imag(\vecg)^T) 
\begin{pmatrix}
  \lambda L \quad & 0 \\ 0 \quad & \lambda L  
 \end{pmatrix}  
\begin{pmatrix}
  \real(\vecg) \\ \imag(\vecg) 
 \end{pmatrix} \\
&=& (\real(\vecg)^T \imag(\vecg)^T) 
\begin{pmatrix}
  \lambda L \real(\vecg) \\ \lambda L \imag(\vecg) 
 \end{pmatrix} \\
&=& \real(\vecg)^T (\lambda L) \real(\vecg) + \imag(\vecg)^T (\lambda L) \imag(\vecg) \\
&=& \lambda (\real(\vecg) - \iota \imag(\vecg))^T L (\real(\vecg) + \iota \imag(\vecg)) \\
&=& \lambda \vecg^{*} L \vecg.
\end{eqnarray}
Next, we can verify that 
\begin{eqnarray}
\real(\vecg^{*}\vecz) 
&=& \real((\real(\vecg) - \iota \imag(\vecg))^T (\real(\vecz) - \iota \imag(\vecz))) \\ 
&=& \real(\vecg)^T \real(\vecz) + \imag(\vecg)^T \imag(\vecz) \\
&=& \vecgbar^T \veczbar.
\end{eqnarray}
Lastly, it is trivially seen that $\norm{\vecgbar}_2^2 = \norm{\vecg}_2^2 = n$. 
Hence \eqref{eq:qcqp_denoise_complex} and \eqref{eq:qcqp_denoise_real} are equivalent. 

\section{Analyzing the solution to our TRS formulation in \eqref{eq:qcqp_denoise_real}}
Let $\set{\lambda_j(\Hbar)}_{j=1}^{2n}$, with $\lambda_1(\Hbar) \leq \lambda_2(\Hbar) \leq \cdots  \lambda_{2n}(\Hbar) $, 
and $\set{\vecq_j}_{j=1}^{2n}$ denote the eigenvalues, respectively eigenvectors,  of  $\Hbar$. 
Note that $\lambda_1(\Hbar) = \lambda_2(\Hbar) = 0$, and $\lambda_3(\Hbar) > 0$ since $G$ is connected. 
Let us denote the null space of $\Hbar$ by $\calN(\Hbar)$, so $\calN(\Hbar) = \text{span} \set{\vecq_1,\vecq_2}$. 
We can now analyze the solution to \eqref{eq:qcqp_denoise_real} with the help of Lemma \ref{lemma:qcqp_denoise_real}, by considering 
the following two cases. 
%

\textbf{Case 1. }  \underline{\textit{$\veczbar \not\perp \calN(\Hbar)$.}} The solution is given by 

\vspace{-5mm}
\begin{equation} 
\hspace{-3mm} \vecgbarest(\mu^{*}) = 2(2\Hbar + \mu^{*}\matI)^{-1} \veczbar = 2\sum_{j=1}^{2n} \frac{\dotprod{\veczbar}{\vecq_j}}{2\lambda_j(\Hbar) + \mu^{*}}\vecq_j, 
\end{equation}
\vspace{-3mm}

for a unique $\mu^{*} \in (0,\infty)$ satisfying $\norm{\vecgbarest(\mu^{*})}^2 = n$. Indeed, 
denoting $\phi(\mu) = \norm{\vecgbarest(\mu)}^2 = 4\sum_{j=1}^{2n} \frac{\dotprod{\veczbar}{\vecq_j}^2}{(2\lambda_j(\Hbar) + \mu)^2}$, 
we can see that $\phi(\mu)$ has a pole at $\mu = 0$ and decreases monotonically to $0$ as $\mu \rightarrow \infty$. Hence, there 
exists a unique $\mu^{*} \in (0,\infty)$ such that $\norm{\vecgbarest(\mu^{*})}^2 = n$. The solution $\vecgbarest(\mu^{*})$ will 
be unique by Lemma \ref{lemma:qcqp_denoise_real},  since $2\Hbar + \mu^{*}\matI \succ 0$ holds.

\textbf{Case 2. } \underline{\textit{$\veczbar \perp \calN(\Hbar)$.}} This second scenario requires additional attention. To begin with, note that   

\vspace{-3mm}
\begin{equation}
\phi(0) = 4\sum_{j=1}^{2n} \frac{\dotprod{\veczbar}{\vecq_j}^2}{(2\lambda_j(\Hbar))^2} = \sum_{j=3}^{2n} \frac{\dotprod{\veczbar}{\vecq_j}^2}{\lambda_j(\Hbar)^2}
\end{equation}
\vspace{-3mm}

is now well defined, i.e., $0$ is not a pole of $\phi(\mu)$ anymore. 
If $\phi(0) > n$, then as before, we can again find a unique $\mu^{*} \in (0,\infty)$ satisfying $\phi(\mu^{*}) = n$. 
The solution is given by $\vecgbarest(\mu^{*}) = 2(2\Hbar + \mu^{*}\matI)^{-1} \veczbar$ and is unique since  $2\Hbar + \mu^{*}\matI \succ 0$ (by Lemma \ref{lemma:qcqp_denoise_real}). 

In case $\phi(0) \leq n$, we set $\mu^{*} = 0$ and define our solution to be of the form 
\begin{equation}
\vecgbarest(\theta,\vecv) = (\Hbar)^{\dagger} \veczbar + \theta \vecv; \quad \vecv \in \calN(\Hbar), \norm{\vecv} = 1,
\end{equation}
where $\dagger$ denotes pseudo-inverse and $\theta \in \matR$. In particular, for any given $\vecv \in \calN(\Hbar), \norm{\vecv} = 1$, 
we obtain $\vecgbarest(\theta^{*},\vecv), \vecgbarest(-\theta^{*},\vecv)$ 
as the solutions to \eqref{eq:qcqp_denoise_real}, with $\pm \theta^{*}$ being the solutions to the equation

\vspace{-5mm}
\begin{equation}
\norm{\vecgbarest(\theta,\vecv)}^2 = n \Leftrightarrow \underbrace{\norm{(\Hbar)^{\dagger} \veczbar}^2}_{= \phi(0) \leq n} + \theta^2 = n
\end{equation}
\vspace{-5mm}

Hence the solution is not unique if $\phi(0) < n$. 
%

\section{Trust region subproblem with $\ell_2$ ball/sphere constraint} \label{sec:trust_region_discuss}
Consider the following two optimization problems
\begin{equation}
\begin{split}
\begin{rcases}
\min_{\vecx} \quad \vecb^T \vecx + \frac{1}{2} \vecx^T \matP \vecx \\
\text{s.t} \norm{\vecx} \leq r
\end{rcases} \text{(P1)}
\end{split}
\qquad \vline \qquad
\begin{split}
\begin{rcases}
\min_{\vecx} \quad \vecb^T \vecx + \frac{1}{2} \vecx^T \matP \vecx \\
\text{s.t} \norm{\vecx} = r
\end{rcases} \text{(P2)}
\end{split}
\label{eq:trust_region_opt} 
\end{equation}
with $\matP \in \matR^{n \times n}$ being a symmetric matrix. 
(P1) is known as the trust region subproblem in the optimization literature and has been 
studied extensively with a rich body of work; (P2) is closely related to (P1) with a non-convex equality constraint. 
There exist algorithms that efficiently find the global solution of (P1) and (P2), to arbitrary accuracy. 
In this section, we provide a discussion on the characterization of the solution of these two problems. 

To begin with, it is useful to note for (P1) that
\begin{itemize}
\item If the solution lies in the interior of the feasible domain, then it implies 
$\matP \succeq 0$. This follows from the second order necessary condition for a local minimizer. 


\item In the other direction, if $\matP \not\succeq 0$ then the solution will always lie on the boundary.
\end{itemize}
Surprisingly, we can characterize the solution of (P1), as shown in the following
\begin{lemma}[\cite{Sorensen82}] \label{lemma:trs_ball}
$\vecx^{*}$ is a solution to (P1) iff $\norm{\vecx^{*}} \leq r$ and $\exists \mu^{*} \geq 0$ such that 
(a) $\mu^{*}(\norm{\vecx^{*}} - r) = 0$, (b) $(\matP + \mu^{*}\matI)\vecx^{*} = -\vecb$ 
and (c) $\matP + \mu^{*}\matI \succeq 0$. Moreover, if $\matP + \mu^{*}\matI \succ 0$, then the solution is unique.
\end{lemma}
Note that if the solution lies in the interior, and if $\matP$ is p.s.d and singular, then 
there will also be a pair of solutions on the boundary of the ball. This is easily verified. 
The solution to (P2) is characterized by the following 
\begin{lemma}[\cite{Hager01,Sorensen82}] \label{lemma:trs_sphere}
$\vecx^{*}$ is a solution to (P2) iff $\norm{\vecx^{*}} = r$ and $\exists \mu^{*}$ such that
(a) $\matP + \mu^{*}\matI \succeq 0$ and (b) $(\matP + \mu^{*}\matI) \vecx^{*} = -\vecb$. 
Moreover, if $\matP + \mu^{*}\matI \succ 0$, then the solution is unique.
\end{lemma}
The solution to (P1), (P2) is closely linked to solving the non linear equation $\norm{(\matP + \mu\matI)^{-1} \vecb} = r$. Let
\begin{equation}
\vecx(\mu) = -(\matP + \mu\matI)^{-1} \vecb = -\sum_{j=1}^{n} \frac{\dotprod{\vecb}{\vecq_j}}{\mu + \mu_j} \vecq_j, 
\end{equation}
where $\mu_1 \leq \mu_2 \leq \dots \leq \mu_n$ and $\set{\vecq_j}$ are the eigenvalues and eigenvectors of 
$\matP$ respectively. Let us define $\phi(\mu)$ as  
\begin{equation}
\phi(\mu) := \norm{\vecx(\mu)}^2 = \sum_{j=1}^{n} \frac{\dotprod{\vecb}{\vecq_j}^2}{(\mu + \mu_j)^2}.
\end{equation}
Denoting $\calS = \set{\vecq \in \matR^n : \matP\vecq = \mu_1 \vecq}$, there are two cases to consider.
\begin{enumerate}
%
%
\item \underline{\textit{$\dotprod{\vecb}{\vecq} \neq 0$ for some $\vecq \in \calS$}}

This is the easy case. $\phi(\mu)$ has a pole at $-\mu_1$ and is monotonically decreasing 
in ($-\mu_1,\infty$) with $\lim_{\mu \rightarrow \infty} \phi(\mu) = 0$ and 
$\lim_{\mu \rightarrow -\mu_1} \phi(\mu) = \infty$. Hence there is a unique 
$\mu^{*} \in (-\mu_1,\infty)$ such that $\phi(\mu) = r^2$, 
and $\vecx(\mu^{*}) = -\sum_{j=1}^{n} \frac{\dotprod{\vecb}{\vecq_j}}{\mu^{*} + \mu_j} \vecq_j$ 
will be the unique solution to (P2). Some remarks are in order. 

\begin{itemize}
\item If $\matP$ was p.s.d and singular, then there is no solution to $\matP\vecx = -\vecb$, 
since $\vecb \not\in$ colspan($\matP$). Also, since $\mu_1 = 0$, we would 
have $\mu^{*} \in (0,\infty)$. Hence the corresponding solution $\vecx(\mu^{*})$ would be the same for 
(P1), (P2) and would be on the boundary. Moreover, the solution will be unique due to Lemma \ref{lemma:trs_sphere}
since $\matP + \mu^{*}\matI \succ 0$. 

\item If $\matP$ was p.d and $\phi(0) < r^2$, then 
this would mean that the global solution to the unconstrained problem is a feasible point for (P1). 
In other words, $\vecx(0) = -\matP^{-1}\vecb$ would be the unique solution to (P1) with $\mu^{*} = 0$. 
Moreover, $\vecx(\mu^{*})$ with $\mu^{*} \in (-\mu_1,\infty)$ satisfying $\phi(\mu^{*}) = r^2$,  
would be the unique solution to (P2) (with $\mu^{*} < 0$); the uniqueness follows from Lemma \ref{lemma:trs_sphere} since 
$\matP + \mu^{*}\matI \succ 0$.
\end{itemize}
\item \underline{\textit{$\dotprod{\vecb}{\vecq} = 0$ for all $\vecq \in \calS$}}

This is referred to as the ``hard'' case in the literature - $\phi(\mu)$ does not  have a pole at $-\mu_1$, so $\phi(-\mu_1)$ is well defined.
There are two possibilities.
\begin{enumerate}
\item If $\phi(-\mu_1) \geq r^2$ then the solution is straightforward -- simply 
find the unique $\mu^{*} \in [-\mu_1,\infty)$ such that $\phi(\mu^{*}) = r^2$. 
This is possible since $\phi(\mu)$ is monotonically decreasing in $[-\mu_1,\infty)$. 
Hence,  
$$\vecx(\mu^{*}) = -\sum_{j: \mu_j \neq \mu_1} \frac{\dotprod{\vecb}{\vecq_j}}{\mu^{*} + \mu_j} \vecq_j$$
is the unique solution to (P2). If $\matP$ was p.s.d and singular, then $\mu_1 = 0$, and 
so $\mu^{*} \geq 0$. Hence, $\vecx(\mu^{*})$ would be the solution to both (P1), (P2) and 
would be on the boundary.

\item If $\phi(-\mu_1) < r^2$ then slightly more work is needed. For $\theta \in \matR$ and 
any $\vecz \in \calS$ with $\norm{\vecz} = 1$, define 
$$\vecx(\theta) := \underbrace{-\sum_{j: \mu_j \neq \mu_1} \frac{\dotprod{\vecb}{\vecq_j}}{\mu_j - \mu_1} \vecq_j}_{\vecx(-\mu_1)} + \theta \vecz.$$
Then, 
\begin{align}
\norm{\vecx(\theta)}^2 
&= \sum_{j: \mu_j \neq \mu_1} \frac{\dotprod{\vecb}{\vecq_j}^2}{(\mu_j - \mu_1)^2} + \theta^2 \\
&= \phi(-\mu_1) + \theta^2.
\end{align}
Solving $\norm{\vecx(\theta)}^2 = r^2$ for $\theta$, we see that for any solution $\theta^{*}$, we will also have 
$-\theta^{*}$ as a solution. Hence, $\vecx(\theta^{*}), \vecx(-\theta^{*})$ will be solutions to (P2) with $\mu^{*} = -\mu_1$. 
If $\matP$ was p.s.d and singular, then $\mu^{*} = 0$, and $\vecx(\pm \theta^{*}) = \vecx(0) \pm \theta^{*}\vecz$ would 
be solutions to both (P1) and (P2). Note that $\vecx(-\mu_1)$ is a solution to (P1). In fact, any 
point in the interior of the form $\vecx(-\mu_1) + \theta\vecz$ is a solution to (P1).
\end{enumerate}
\end{enumerate}

\section{Proof of Lemma \ref{lem:wrap_dist_fin_bd}}
\begin{proof}
To begin with, note that $\abs{\gest_i - h_i} \leq \epsilon$  implies $\abs{\gest_i} \in [1-\epsilon,1+\epsilon]$. This means that  
$\abs{\gest_i} > 0$ holds if $\epsilon < 1$. Consequently, we obtain
\begin{align}
\Bigl| \frac{\gest_i}{\abs{\gest_i}} - h_i\Bigr| 
&= \Bigl| \frac{\gest_i}{\abs{\gest_i}} - \frac{h_i}{\abs{\gest_i}} + \frac{h_i}{\abs{\gest_i}} - h_i\Bigr| \\
&\leq \frac{\abs{\gest_i - h_i}}{\abs{\gest_i}} + \abs{h_i}\left(\frac{\abs{\abs{\gest_i} - 1}}{\abs{\gest_i}}\right) \\
&\leq \frac{2\epsilon}{\abs{\gest_i}} \leq \frac{2\epsilon}{1-\epsilon}. \label{eq:denoise_mod1_rec_temp1}
\end{align}  
We will now show that provided $0 < \epsilon < 1/2$ holds, then \eqref{eq:denoise_mod1_rec_temp1} implies the bound \eqref{eq:wrap_dist_fin_bd}.
Indeed, from the definition of $h_i$, and of $\gest_i/\abs{\gest_i}$ (from \eqref{eq:extr_mod1_vals}), we have 
\begin{align}
\Bigl| \frac{\gest_i}{\abs{\gest_i}} - h_i\Bigr| 
&= \abs{\exp(\iota 2\pi (\widehat{f_i} \bmod 1)) - \exp(\iota 2\pi (f_i \bmod 1))} \\
&= \abs{1 - \exp(\iota 2\pi (f_i \bmod 1 - \widehat{f_i} \bmod 1))} \\
&= 2\abs{\sin [\pi \underbrace{(f_i \bmod 1 - \widehat{f_i} \bmod 1)}_{\in (-1,1)}]} \\
&= 2\sin [\pi \abs{(f_i \bmod 1 - \widehat{f_i} \bmod 1)}] \\ 
&= 2\sin [\pi (1- \abs{(f_i \bmod 1 - \widehat{f_i} \bmod 1)})]. \label{eq:denoise_mod1_rec_temp2}
\end{align}
Then, \eqref{eq:wrap_dist_fin_bd} follows from \eqref{eq:denoise_mod1_rec_temp2}, \eqref{eq:denoise_mod1_rec_temp1} 
and by noting that $0 < \epsilon/(1-\epsilon) < 1$ for $0 < \epsilon < 1/2$.
\end{proof}
%

\section{Proof of Lemma \ref{lem:lowbd_feas_based}} \label{sec:lem_lowbd_feas_based}
\begin{proof}
To begin with, note that
\begin{equation} \label{eq:lem1_temp1}
\norm{\veczbar - \vechtilbar}_2 \leq \delta\sqrt{n} \Leftrightarrow \dotprod{\veczbar}{\vechtilbar} \geq n - \frac{\delta^2 n}{2}.
\end{equation}
Since $\vechtilbar \in \matR^{2n}$ is feasible for \eqref{eq:qcqp_denoise_real}, we get 
\begin{align}
&\vechtilbar^{T} \Hbar \vechtilbar  - 2 \dotprod{\vechtilbar}{\veczbar} \geq \vecgbarest^{T} \Hbar \vecgbarest  - 2 \dotprod{\vecgbarest}{\veczbar} \\
&\Leftrightarrow \dotprod{\vecgbarest}{\veczbar} \geq \dotprod{\vechtilbar}{\veczbar} - \frac{1}{2} \vechtilbar^{T} \Hbar \vechtilbar
+ \frac{1}{2} \vecgbarest^{T} \Hbar \vecgbarest \\
&\geq n - \frac{\delta^2 n}{2} - \frac{1}{2} \vechtilbar^{T} \Hbar \vechtilbar 
+ \frac{1}{2} \vecgbarest^{T} \Hbar \vecgbarest \quad \text{(from \eqref{eq:lem1_temp1})}.  \label{eq:lem1_temp2}
\end{align}
Moreover, we can upper bound $\dotprod{\vecgbarest}{\veczbar}$ as follows.
\begin{align}
\dotprod{\vecgbarest}{\veczbar} 
&= \dotprod{\vecgbarest}{\veczbar - \vechtilbar} + \dotprod{\vecgbarest}{\vechtilbar} \\
&\leq \norm{\vecgbarest}_2 \norm{{\veczbar - \vechtilbar}}_2 + \dotprod{\vecgbarest}{\vechtilbar} \quad \text{(Cauchy-Schwarz)} \\
&\leq \sqrt{n} (\sqrt{n} \delta) + \dotprod{\vecgbarest}{\vechtilbar} \quad \text{(from \eqref{eq:lem1_temp1})}. \label{eq:lem1_temp3}
\end{align}
Plugging \eqref{eq:lem1_temp3} in \eqref{eq:lem1_temp2} and using $\delta^2 \leq \delta$ for $\delta \in [0,1]$ completes the proof.
\end{proof}

\section{Proof of Lemma \ref{lem:upbd_clean_quad_form}} \label{sec:upbd_clean_quad_form}
\begin{proof}
Denoting $\vechtil = (\htil_1 \dots \htil_n)^T \in \mathbb{C}^n$ to be the complex valued 
representation of $\vechtilbar \in \matR^{2n}$ as per \eqref{eq:real_notat}, clearly
{\small
\begin{align} 
\frac{1}{2n} \vechtilbar^T \Hbar \vechtilbar 
= \frac{\lambda}{2n} \sum_{(i,j) \in E} \abs{\htil_i - \htil_j}^2 
\leq \frac{\lambda}{2n} \abs{E} \max_{(i,j) \in E} \abs{\htil_i - \htil_j}^2.  \label{eq:lem2_temp1}
\end{align}}
Since for each $i \in V$ we have $\text{deg}(i) \leq 2k$, hence $\abs{E} = (1/2) \sum_{i \in V} \text{deg}(i) \leq k n$. 
Next, for any $(i,j) \in E$ note that by H\"older continuity of $f$ we have
\begin{equation} \label{eq:lem2_temp2}
\abs{f_i - f_j} \leq M \abs{x_i - x_j}^{\alpha} \leq M \left(\frac{k}{n-1}\right)^{\alpha} \leq M\left(\frac{2k}{n}\right)^{\alpha}
\end{equation}
if $n \geq 2$ (since then $n-1 \geq n/2$). Finally, we can bound $\abs{\htil_i - \htil_j}$ as follows.
%
%
\begin{align}
\abs{\htil_i - \htil_j} &= \abs{1 - \exp(\iota 2\pi (f_j - f_i))} \\
&= 2\abs{\sin \pi (f_j - f_i)} \\
&\leq 2\pi \abs{f_j - f_i} \quad (\text{since} \ \abs{\sin x} \leq \abs{x}; \ \forall x \in \matR) \\ 
&\leq \frac{2\pi M (2k)^{\alpha}}{n^{\alpha}} \quad (\text{using} \ \eqref{eq:lem2_temp2}).  \label{eq:lem2_temp3}
\end{align}
Plugging \eqref{eq:lem2_temp3} in \eqref{eq:lem2_temp1} with the bound $\abs{E} \leq kn$ yields the stated bound.
\end{proof}

\section{Proof of Lemma \ref{lem:lowbd_sol_quad_form}} \label{sec:proof_lemma_lowbd_sol_quad_form}
\begin{proof}
Let $\set{\lambda_j(\Hbar)}_{j=1}^{2n}$ (with $\lambda_1(\Hbar) \leq \lambda_2(\Hbar) \leq \cdots$) and $\set{\vecq_j}_{j=1}^{2n}$ denote 
the eigenvalues and eigenvectors respectively for $\Hbar$.  
Also, let $0 = \beta_1(L) < \beta_2(L) \leq \beta_3(L) \leq \cdots \leq \beta_n(L)$ denote the eigenvalues of the Laplacian $L$. 
Note that $\beta_2(L) > 0$ since $G$ is connected. By Gershgorin's disk theorem, 
it is easy to see\footnote{Denote $L_{ij}$ to be the $(i,j)^{\text{th}}$ entry of $L$. Then by Gershgorin's disk theorem, we 
know that each eigenvalue lies in $\bigcup_{i=1}^{2n}\set{x:\abs{x-L_{ii}} \leq \sum_{j \neq i} \abs{L_{ij}}}$. Since $L_{ii} \leq 2k$ 
and $\sum_{j \neq i} \abs{L_{ij}} \leq 2k$ holds for each $i$, the claim follows.}
that $\beta_n(L) \leq 4k$ for the graph $G$. Hence,
\begin{equation} \label{eq:lem3_temp0}
0 = \lambda_1(\Hbar) = \lambda_2(\Hbar) < \lambda_3(\Hbar) \leq \cdots \leq \lambda_{2n}(\Hbar) \leq 4\lambda k
\end{equation}
and $\calN(\Hbar) = \text{span}\set{\vecq_1,\vecq_2}$. We now consider the two cases separately below. 
\begin{enumerate}
\item Consider the case where $\veczbar \not\perp \calN(\Hbar)$. 
We know that $\vecgbarest = 2(2\Hbar + \mu^{*}\matI)^{-1}\veczbar$ 
for a unique $\mu^{*} \in (0,\infty)$ (and so $\vecgbarest$ is the unique solution to (P) by 
Lemma \ref{lemma:qcqp_denoise_real} since $2\Hbar + \mu^{*}\matI \succ 0$) satisfying 
\begin{equation}\label{eq:lem3_temp1}
\norm{\vecgbarest}^2 = 4 \sum_{j=1}^{2n} \frac{\dotprod{\veczbar}{\vecq_j}^2}{(2\lambda_j(\Hbar) + \mu^{*})^2} = n.
\end{equation}
Since $\lambda_j(\Hbar) \geq 0$ for all $j$, hence we obtain from \eqref{eq:lem3_temp1} that
\begin{align}
n &\leq 4\sum_{j=1}^{2n} \frac{\dotprod{\veczbar}{\vecq_j}^2}{(\mu^{*})^2} = \frac{4n}{(\mu^{*})^2} \\ 
\Rightarrow \mu^{*} &\leq 2. \label{eq:lem3_temp2}
\end{align}
Note that equality holds in \eqref{eq:lem3_temp2} if $\veczbar \in \calN(\Hbar)$. We can now lower bound 
$\frac{1}{2n} \vecgbarest^T \Hbar \vecgbarest$ as follows.
\begin{align}
\frac{1}{2n} \vecgbarest^T \Hbar \vecgbarest 
&= \frac{2}{n}\veczbar^T (2\Hbar + \mu^{*}\matI)^{-1} \Hbar (2\Hbar + \mu^{*}\matI)^{-1} \veczbar \\
& = \frac{2}{n} \sum_{j=1}^{2n} \frac{\dotprod{\veczbar}{\vecq_j}^2 \lambda_j(\Hbar)}{(2\lambda_j(\Hbar) + \mu^{*})^2} \\
&\geq \frac{2}{n (8 \lambda k + 2)^2} \sum_{j=1}^{2n}\dotprod{\veczbar}{\vecq_j}^2 \lambda_j(\Hbar) \quad (\text{from} \ \eqref{eq:lem3_temp0}, \eqref{eq:lem3_temp2}) \\ 
&= \frac{1}{(4\lambda k + 1)^2} \left(\frac{1}{2n} \veczbar^T \Hbar \veczbar\right).
\end{align}
%
%
\item Let us now consider the case where $\veczbar \perp \calN(\Hbar)$. 
Denote 
\begin{equation}
\phi(\mu) := \norm{\vecgbarest(\mu)}^2 = 4 \sum_{j=1}^{2n} \frac{\dotprod{\veczbar}{\vecq_j}^2}{(2\lambda_j(\Hbar) + \mu)^2} 
= 4 \sum_{j=3}^{2n} \frac{\dotprod{\veczbar}{\vecq_j}^2}{(2\lambda_j(\Hbar) + \mu)^2}. 
\end{equation}
Observe that $\phi$ does not have a pole at $0$ anymore, $\phi(0)$ is well defined. In order to have a unique 
$\vecgbarest$, it is sufficient if $\phi(0) > n$ holds. Indeed, we would then have a unique $\mu^{*} \in (0,\infty)$ 
such that $\norm{\vecgbarest(\mu^{*})}^2 = n$. Consequently, $\vecgbarest(\mu^{*})$ will be the unique solution to (P) by 
Lemma \ref{lemma:qcqp_denoise_real} since $2\Hbar + \mu^{*}\matI \succ 0$. Now let us note that 
\begin{equation} \label{eq:lem3_case2_temp1}
\phi(0) = \sum_{j=3}^{2n} \frac{\dotprod{\veczbar}{\vecq_j}^2}{\lambda_j(\Hbar)^2} \geq  \frac{n}{16\lambda^2 k^2}
\end{equation}
since $\lambda_j(\Hbar) \leq 4\lambda k$ for all $j$ (recall \eqref{eq:lem3_temp0}). Therefore clearly, the choice 
$\lambda < \frac{1}{4k}$ implies $\phi(0) > n$, and consequently that the solution $\vecgbarest$ is unique. 
Assuming $\lambda < \frac{1}{4k}$ holds, we can derive an upper bound on $\mu^{*}$ as follows.
\begin{align}
n = 4 \sum_{j=3}^{2n} \frac{\dotprod{\veczbar}{\vecq_j}^2}{(2\lambda_j(\Hbar) + \mu^{*})^2} 
\leq 4 \sum_{j=3}^{2n} \frac{\dotprod{\veczbar}{\vecq_j}^2}{(2\lambda_3(\Hbar) + \mu^{*})^2}  
&= \frac{4 n}{(2\lambda_3(\Hbar) + \mu^{*})^2} \\ 
\Rightarrow \mu^{*} &\leq 2 - 2\lambda_3(\Hbar). \label{eq:lem3_case2_temp2}
\end{align}
Hence $\mu^{*} \in (0,2 - 2\lambda_3(\Hbar))$ when $\lambda < \frac{1}{4k}$. We can 
now lower bound $\frac{1}{2n} \vecgbarest^T \Hbar \vecgbarest$ in the same manner as before.
\begin{align}
\frac{1}{2n} \vecgbarest^T \Hbar \vecgbarest 
&= \frac{2}{n}\veczbar^T (2\Hbar + \mu^{*}\matI)^{-1} \Hbar (2\Hbar + \mu^{*}\matI)^{-1} \veczbar \\
& = \frac{2}{n} \sum_{j=1}^{2n} \frac{\dotprod{\veczbar}{\vecq_j}^2 \lambda_j(\Hbar)}{(2\lambda_j(\Hbar) + \mu^{*})^2} \\
&\geq \frac{2}{n (8 \lambda k + 2 - 2\lambda_3(\Hbar))^2} \sum_{j=1}^{2n}\dotprod{\veczbar}{\vecq_j}^2 \lambda_j(\Hbar) \quad 
(\text{from} \eqref{eq:lem3_temp0}, \eqref{eq:lem3_case2_temp2}) \\
&= \frac{1}{(1 + 4\lambda k - \lambda_3(\Hbar))^2} \left(\frac{1}{2n} \veczbar^T \Hbar \veczbar\right). \label{eq:lem3_case2_temp3}
\end{align}
It remains to lower bound $\lambda_3(\Hbar) = \lambda \beta_2(L)$. We do this by using the following result
by Fiedler \cite{Fiedler1973} (adjusted to our notation) for lower bounding the second smallest eigenvalue of the Laplacian of a simple graph.
\begin{theorem}[\cite{Fiedler1973}] \label{thm:fiedler_ss_eigval}
Let $G$ be a simple graph of order $n$ other than a complete graph, 
with vertex connectivity $\kappa(G)$ and edge connectivity $\kappa^{\prime}(G)$. Then, 
\begin{equation}
2\kappa^{\prime}(G) (1 - \cos(\pi/n)) \leq \beta_2(L) \leq \kappa(G) \leq \kappa^{\prime}(G).
\end{equation}
\end{theorem}
The graph $G$ in our setting has $\kappa(G) = k$; indeed, there does not exist a vertex cut of size $k-1$ or less, 
but there does exist a vertex cut of size $k$. 
This in turn means that $\kappa^{\prime}(G) \geq k$, and so from Theorem \ref{thm:fiedler_ss_eigval} we 
get 
\begin{equation} 
\beta_2(L) \geq 2k (1 - \cos(\pi/n)) = 4k \sin^2\left(\frac{\pi}{2n}\right).
\end{equation} 
Hence $\lambda_3(\Hbar) \geq 4 \lambda k \sin^2\left(\frac{\pi}{2n}\right)$. Plugging this in to \eqref{eq:lem3_case2_temp2} completes the proof.
\end{enumerate}
\end{proof}

\section{Appendix Numerical experiments} \label{sec:app_num_exps}

\begin{figure*}
\subcaptionbox[Short Subcaption]{ Clean $ f \mod 1 $ 
}[ 0.32\textwidth ]
{\includegraphics[width=0.32\textwidth] {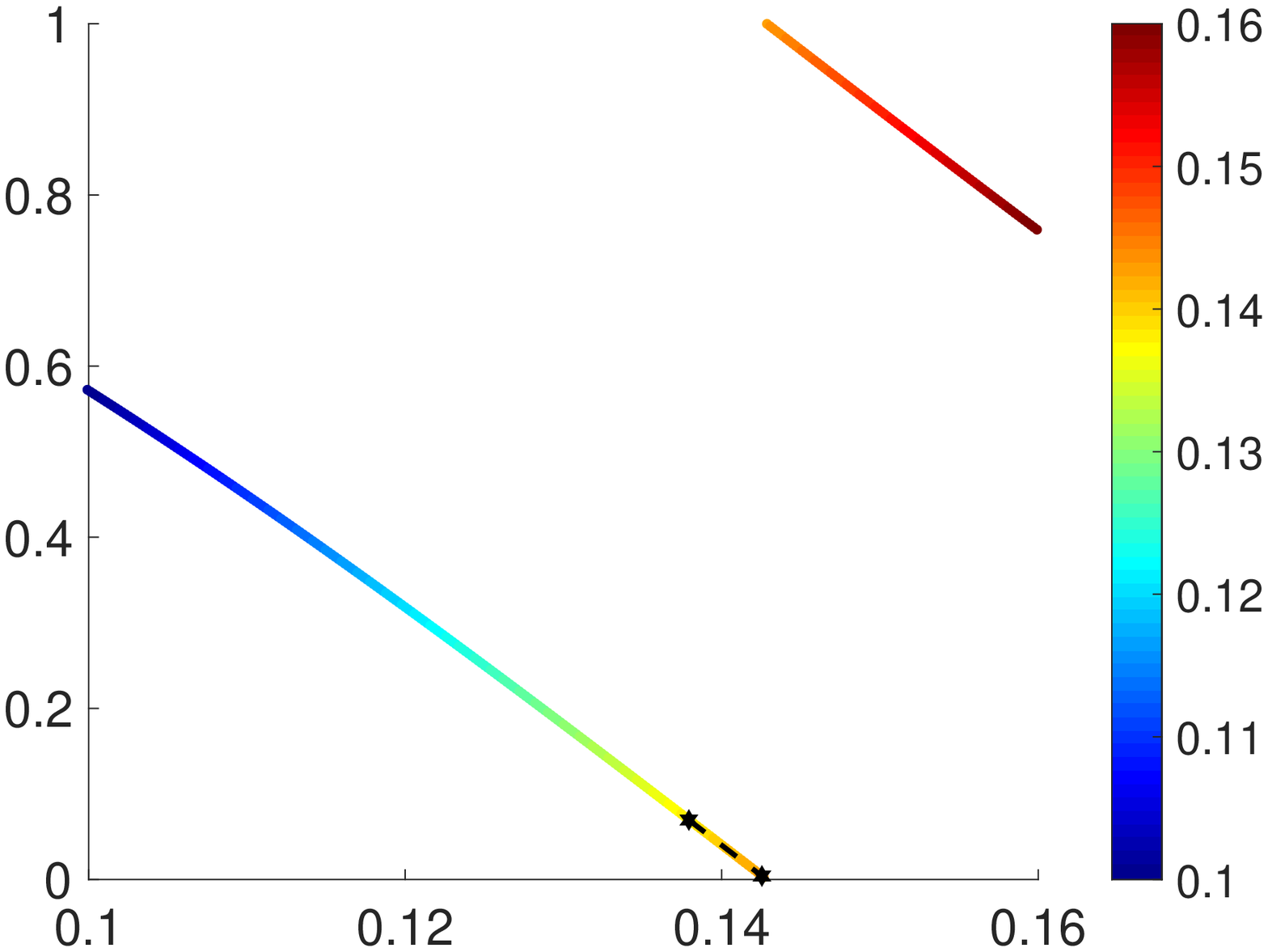} }
\hspace{0.02\textwidth} 
\subcaptionbox[Short Subcaption]{Noisy values  $ f \mod 1 $ 
}[ 0.32\textwidth ]
{\includegraphics[width=0.32\textwidth] {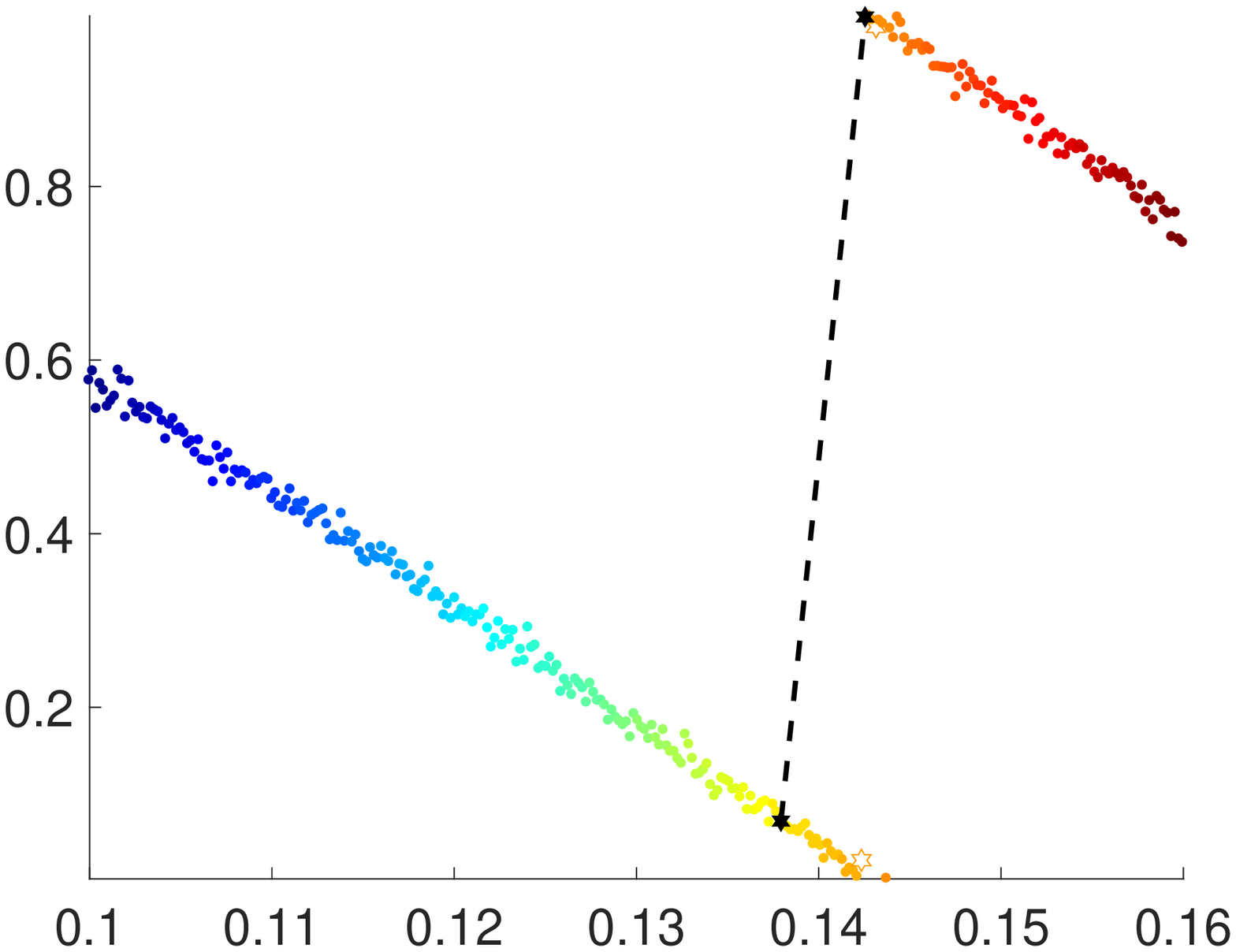} }
\subcaptionbox[Short Subcaption]{ Noisy angular embedding.
\label{subfig:toyExampleFarNear_Noisy_c}
}[ 0.28\textwidth ]
{\includegraphics[width=0.28\textwidth] {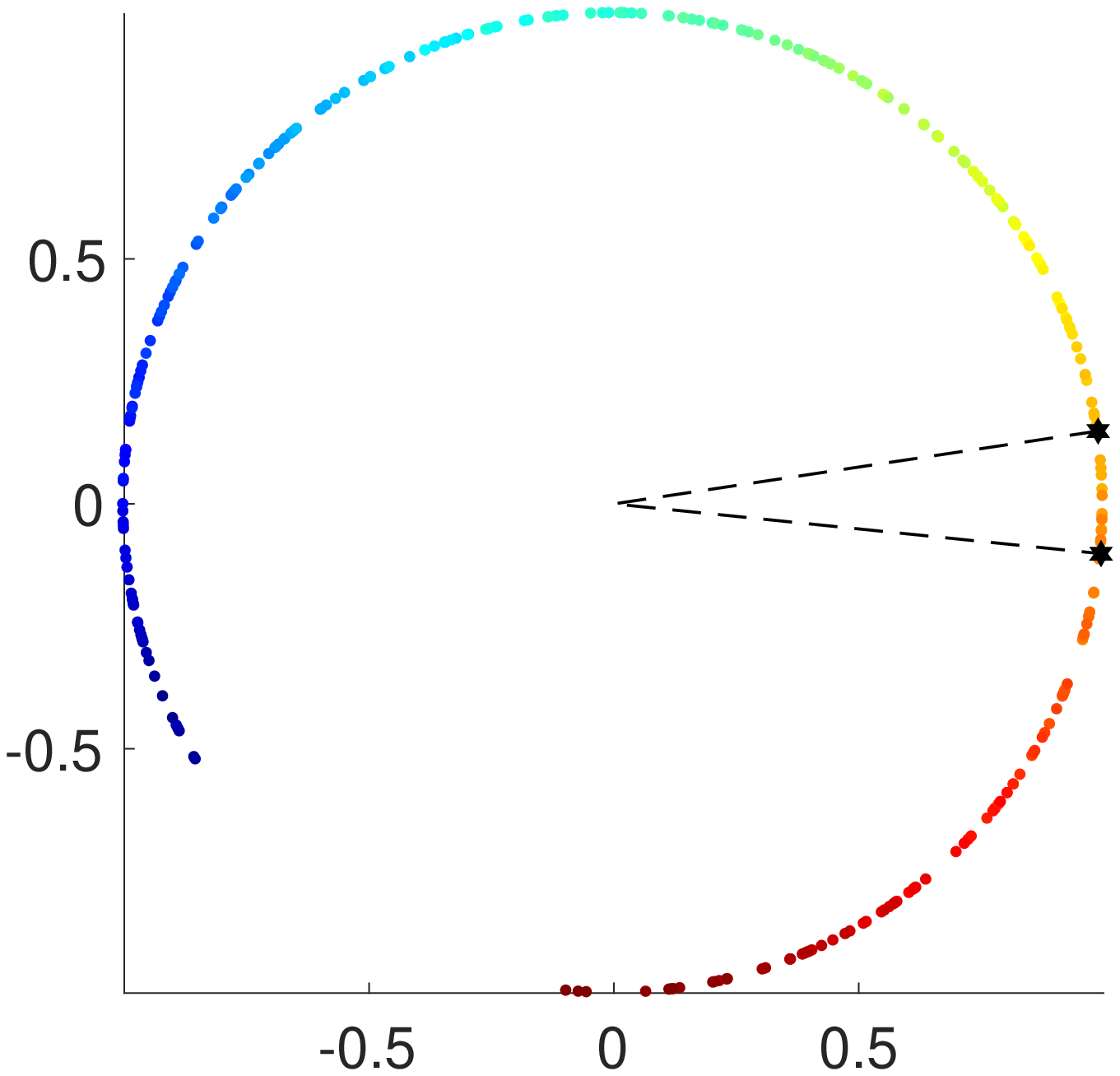} }
\hspace{0.02\textwidth} 
\vspace{-1mm} 
\captionsetup{width=0.95\linewidth}
\caption[Short Caption]{Motivation for the angular embedding approach. Noise perturbations may take nearby points (mod 1 samples) far away in the real domain, yet the points will remain close in the angular domain.}
\label{fig:toyExampleFarNear_Noisy}
\end{figure*}

Figure  \ref{fig:toyExampleFarNear_Noisy} is the analogue of Figure  \ref{fig:toyExampleFarNear}, but for a noisy instance of the problem, making the point that the angular representation facilitates the denoising proces. For points $x_i,x_j$ sufficiently close, the corresponding  samples  $f_i, f_j$ will also be close in the real domain, by H\"older continuity of $f$. When measurements get perturbed by noise, the distance in the real domain between the noisy mod 1 samples can greatly increase and become close to 1 (in this example, the rightmost black point gets perturbed by noise, hits the floor and ''resets" itself). However, in the angular embedding space, the two points still remain close to each other, as depicted in Figure \ref{subfig:toyExampleFarNear_Noisy_c}.

\begin{figure*}
\centering
\subcaptionbox[]{  $\sigma=0.1$
}[ 0.20\textwidth ]
{\includegraphics[width=0.20\textwidth] {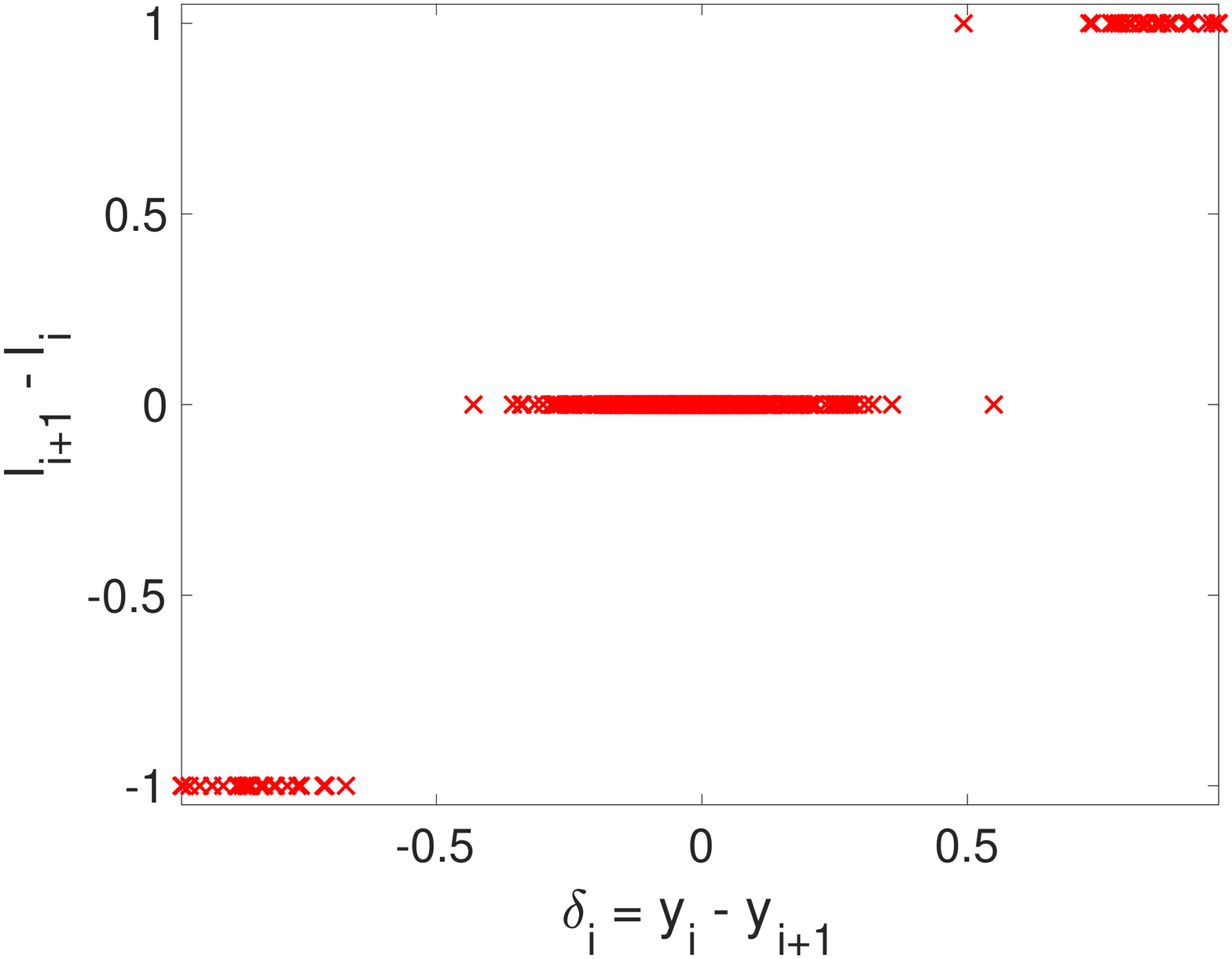} }
\hspace{0.01\textwidth} 
\subcaptionbox[]{  $\sigma=0.1$
}[ 0.74\textwidth ]
{\includegraphics[width=0.74\textwidth] {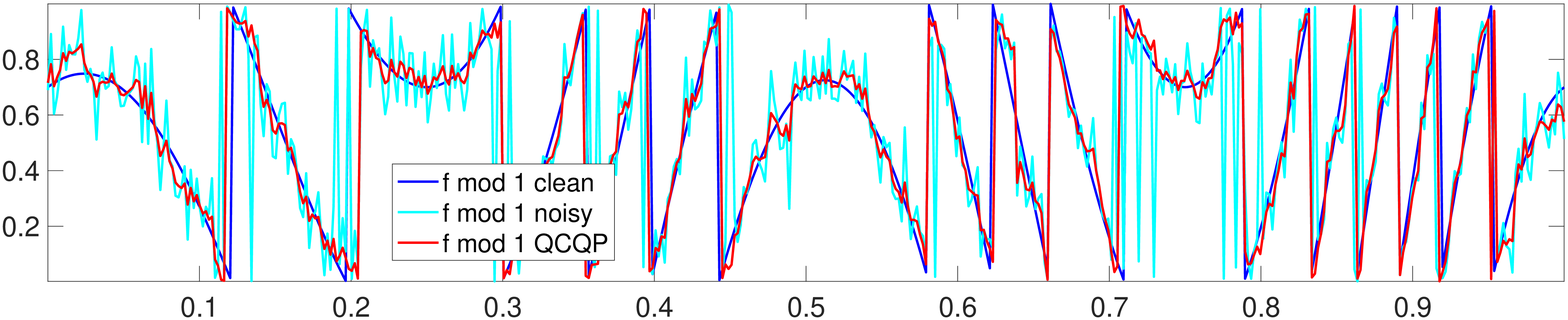} }
\hspace{0.01\textwidth} 
\subcaptionbox[]{  $\sigma=0.15$
}[ 0.20\textwidth ]
{\includegraphics[width=0.20\textwidth] {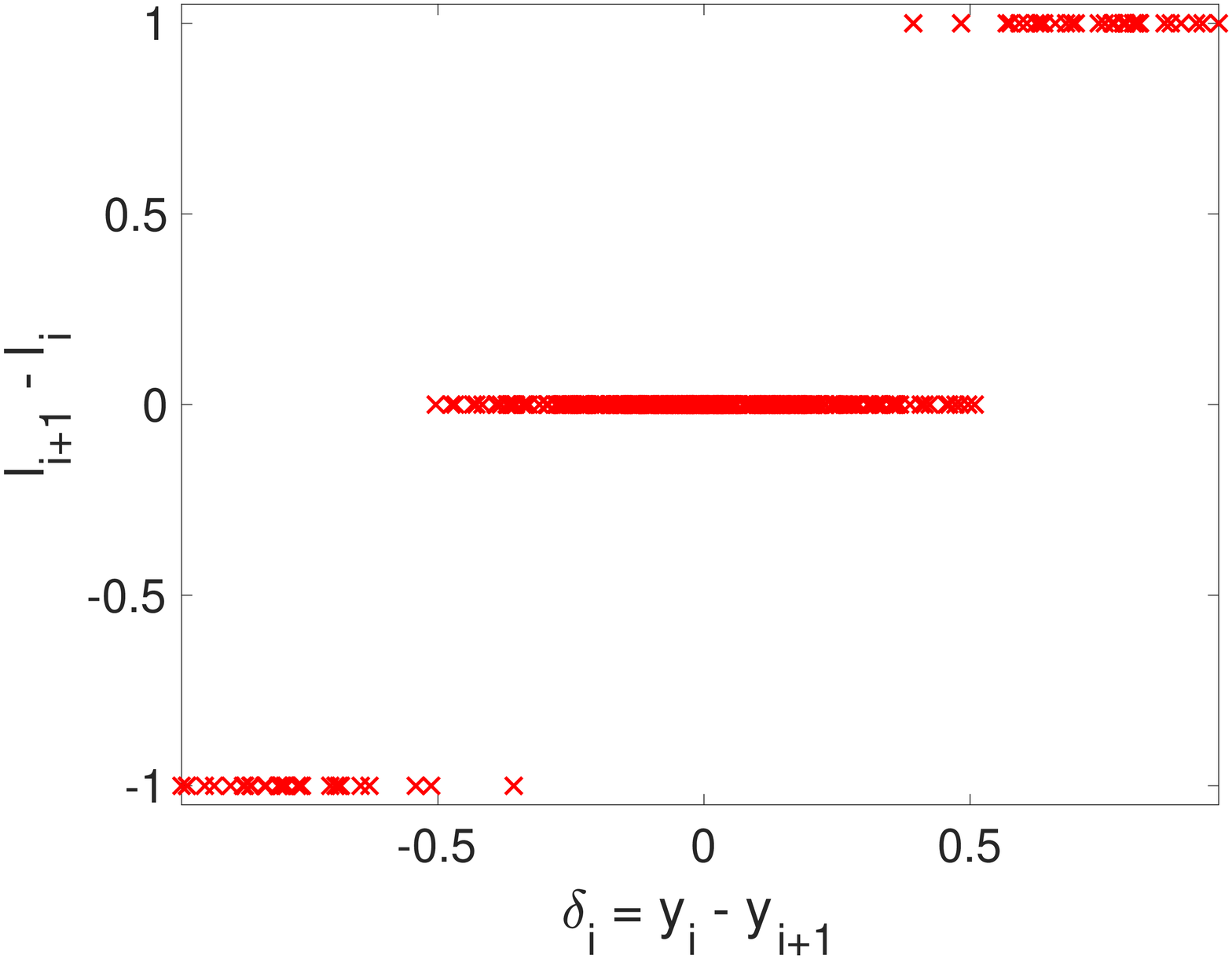} }
\hspace{0.01\textwidth} 
\subcaptionbox[]{  $\sigma=0.15$
}[ 0.74\textwidth ]
{\includegraphics[width=0.74\textwidth] {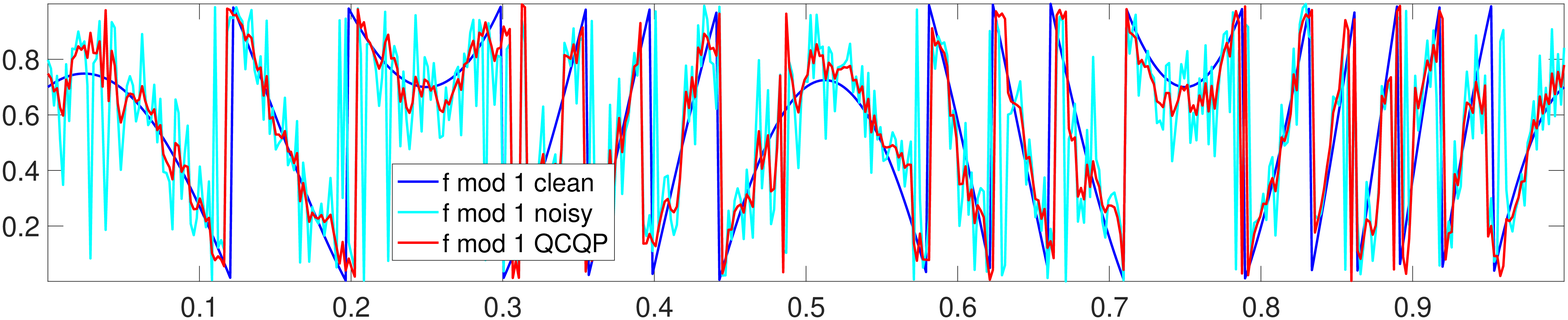} }
\hspace{0.01\textwidth} 
\subcaptionbox[]{  $\sigma=0.2$
}[ 0.20\textwidth ]
{\includegraphics[width=0.20\textwidth] {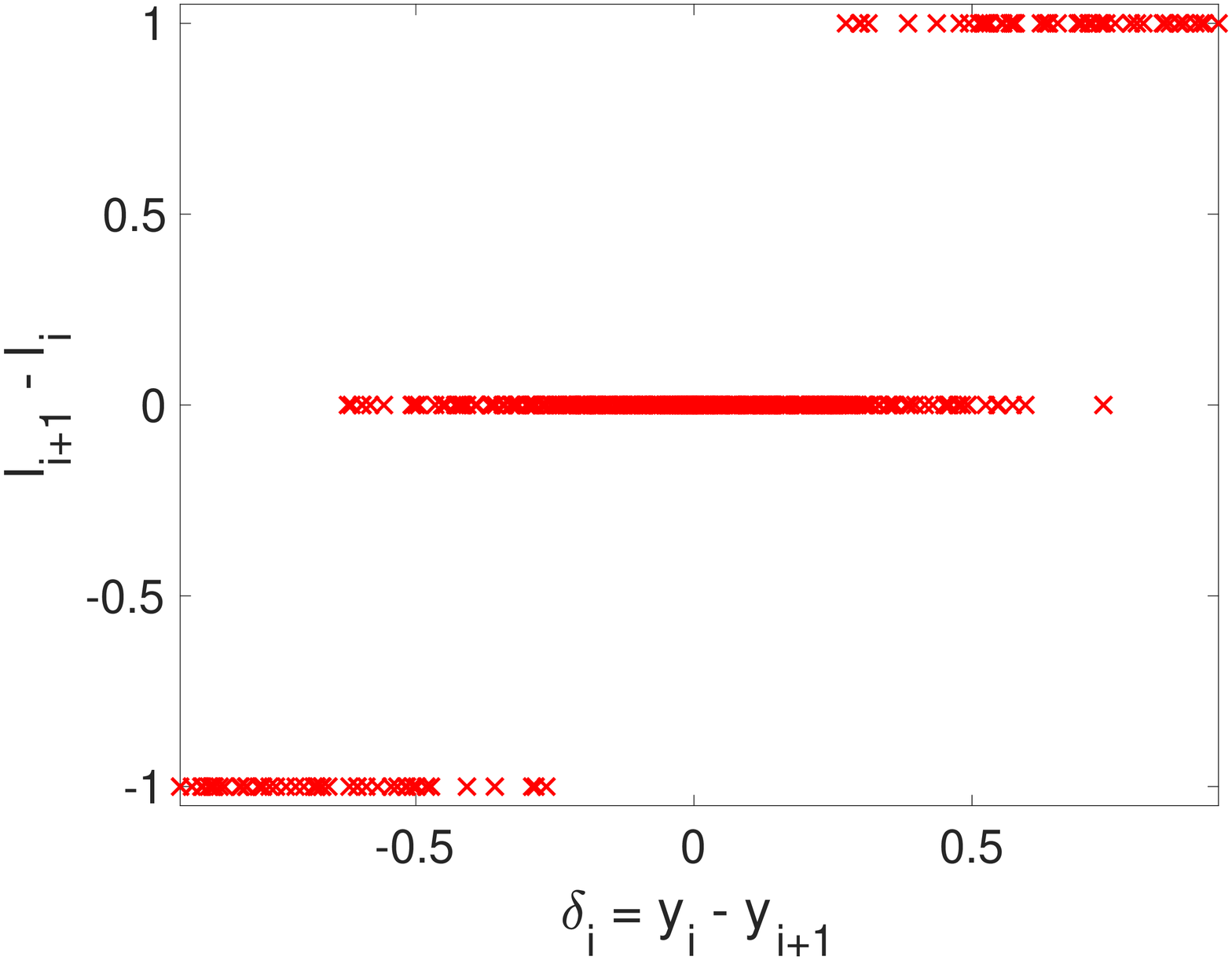} }
\hspace{0.01\textwidth} 
\subcaptionbox[]{  $\sigma=0.2$
}[ 0.74\textwidth ]
{\includegraphics[width=0.74\textwidth] {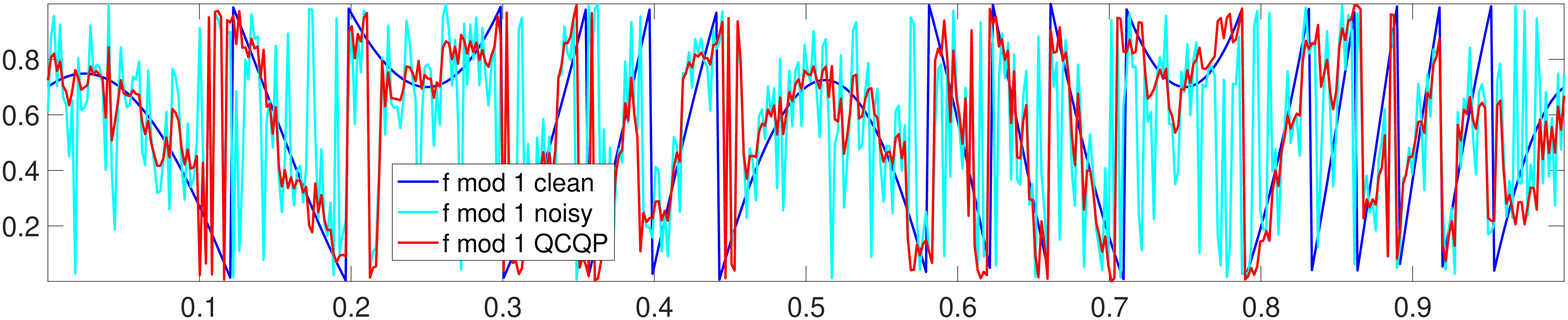} }
%
\vspace{-2mm}
\captionsetup{width=0.95\linewidth}
\caption[Short Caption]{ Noisy instances of the Gaussian noise model ($n=500$). Left: scatter plot of change in $y$ (the observed noisy f mod 1 values) versus change in $l$ (the noisy quotient). Right: plot of the clean f mod 1 values (blue), the noisy f mod 1 values (cyan) and the denoised (via QCQP) f mod 1 values. 
}
\label{fig:instances_f1_Gaussian_delta_cors}
\end{figure*}

Figure \ref{fig:instances_f1_Gaussian_delta_cors} is the analogue of Figure \ref{fig:instances_f1_Bounded_delta_cors}, but for the Gaussian noise model. Its left plots provide intuition for the interplay between the change in $y$ (the observed noisy f mod 1 values) versus change in $l$ (the noisy quotient). Its right plots show the clean, noisy and denoised (via \textbf{QCQP}) mod 1 sample.

\begin{figure*}
\centering
\subcaptionbox[]{  $\gamma=0.15$, \textbf{OLS}
}[ 0.32\textwidth ]
{\includegraphics[width=0.32\textwidth] {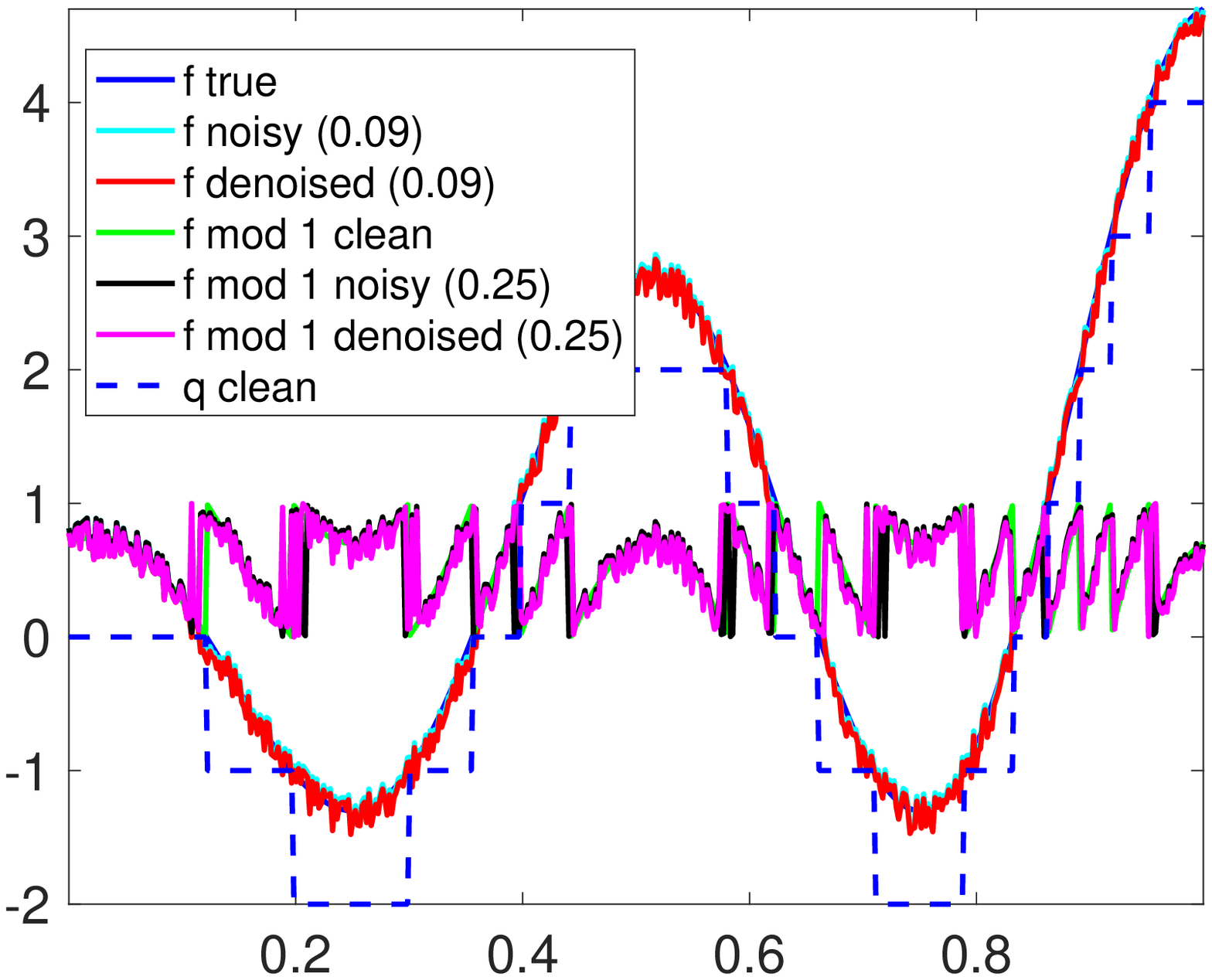} }
%
\subcaptionbox[]{  $\gamma=0.15$, \textbf{QCQP}
}[ 0.32\textwidth ]
{\includegraphics[width=0.32\textwidth] {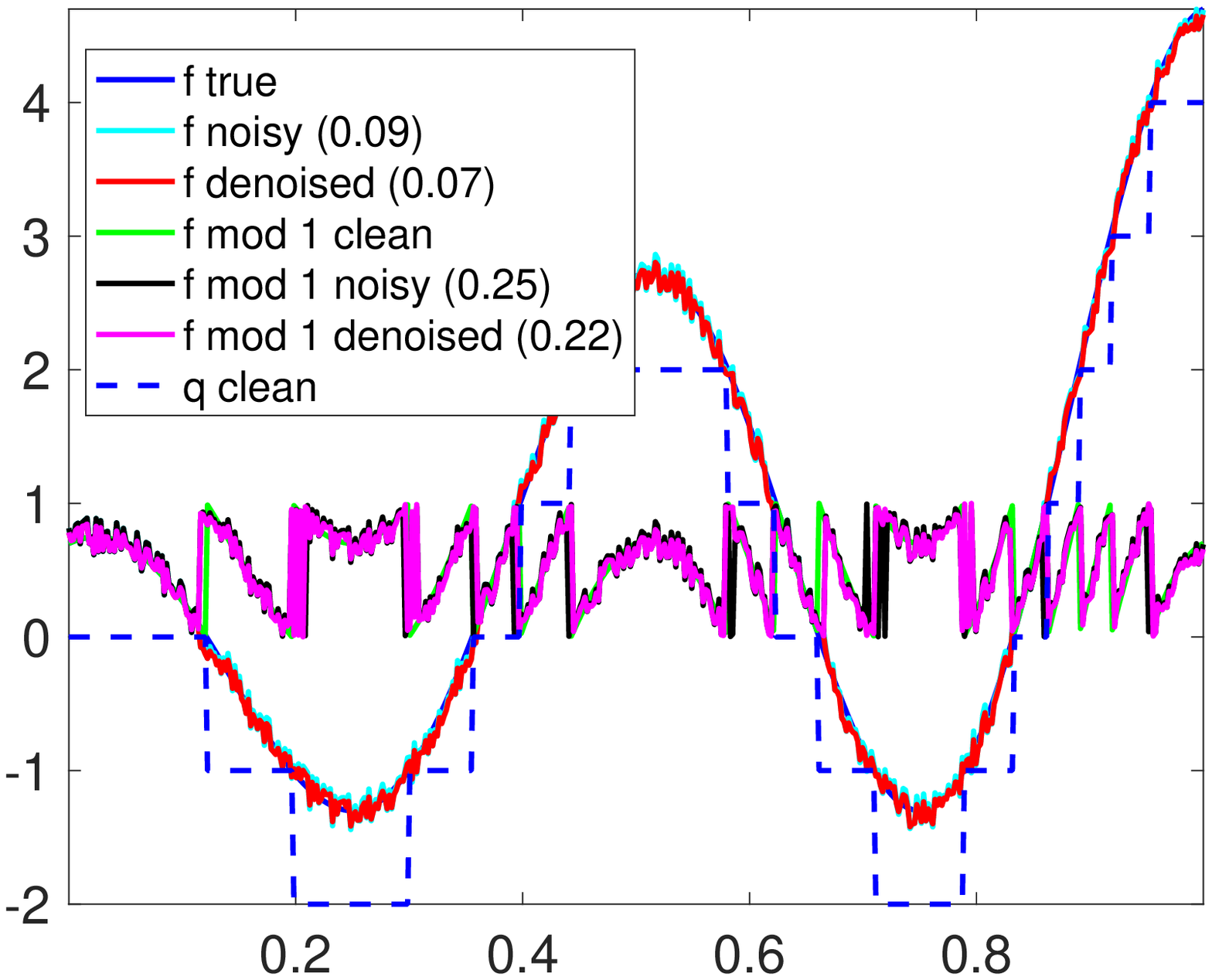} }
%
\subcaptionbox[]{  $\gamma=0.15$, \textbf{iQCQP}  (10  iters.)
}[ 0.32\textwidth ]
{\includegraphics[width=0.32\textwidth] {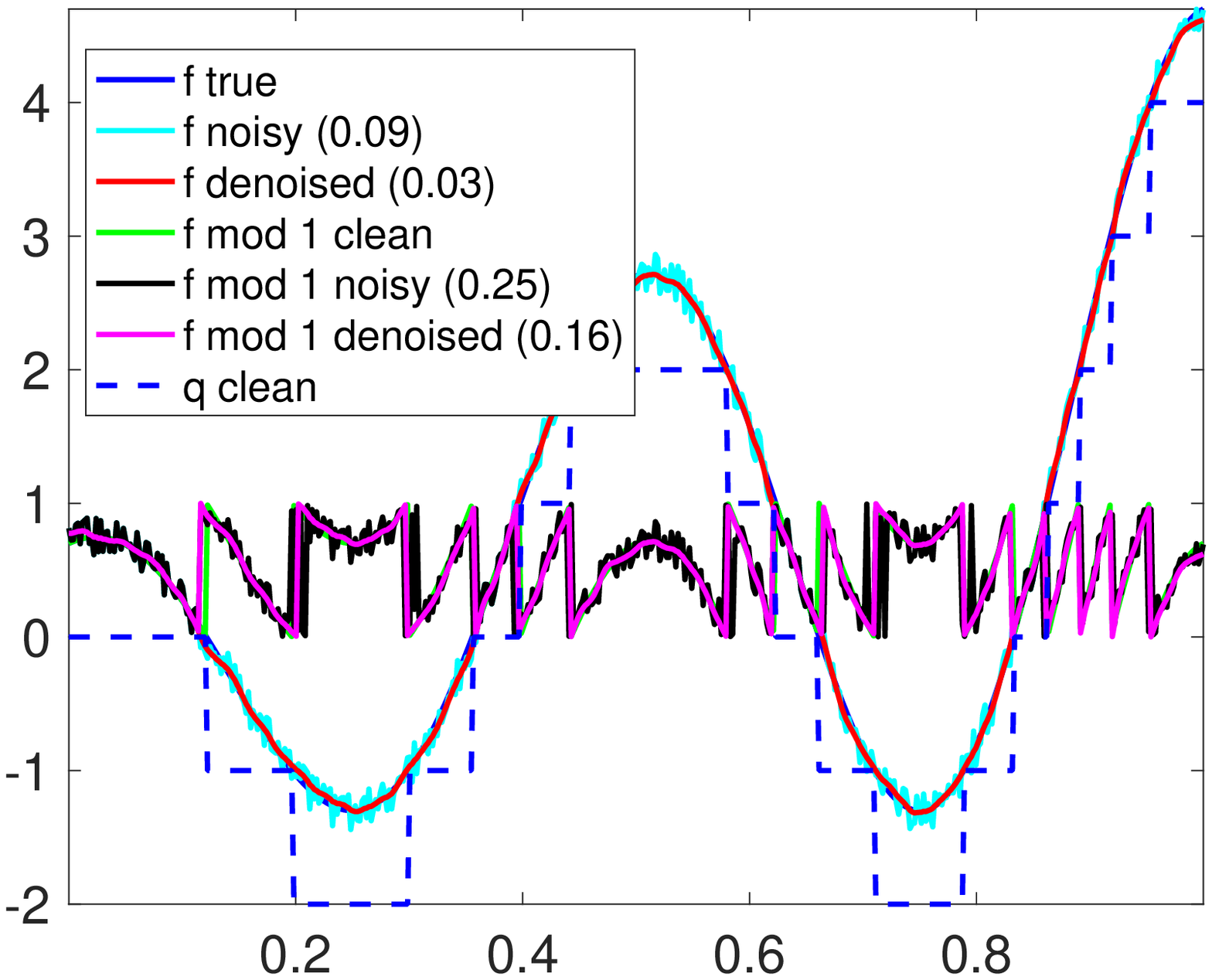} }
%
%

\subcaptionbox[]{  $\gamma=0.27$, \textbf{OLS}
}[ 0.32\textwidth ]
{\includegraphics[width=0.32\textwidth] {figs/instances/f1_Bounded_sigma_0p27_n_500_k_2_lambda_0p1_instance_OLS.eps} }
%
\subcaptionbox[]{  $\gamma=0.27$, \textbf{QCQP}
}[ 0.32\textwidth ]
{\includegraphics[width=0.32\textwidth] {figs/instances/f1_Bounded_sigma_0p27_n_500_k_2_lambda_0p1_instance_QCQP.eps} }
%
\subcaptionbox[]{  $\gamma=0.27$, \textbf{iQCQP} (10 iters.)
}[ 0.32\textwidth ]
{\includegraphics[width=0.32\textwidth] {figs/instances/f1_Bounded_sigma_0p27_n_500_k_2_lambda_0p1_instance_iQCQP.eps} }
%
%

\subcaptionbox[]{  $\gamma=0.30$, \textbf{OLS}
}[ 0.32\textwidth ]
{\includegraphics[width=0.32\textwidth] {figs/instances/f1_Bounded_sigma_0p3_n_500_k_2_lambda_0p1_instance_OLS.eps} }
%
\subcaptionbox[]{  $\gamma=0.30$, \textbf{QCQP}
}[ 0.32\textwidth ]
{\includegraphics[width=0.32\textwidth] {figs/instances/f1_Bounded_sigma_0p3_n_500_k_2_lambda_0p1_instance_QCQP.eps} }
%
\subcaptionbox[]{  $\gamma=0.30$, \textbf{iQCQP}  (10 iters.)
}[ 0.32\textwidth ]
{\includegraphics[width=0.32\textwidth] {figs/instances/f1_Bounded_sigma_0p3_n_500_k_2_lambda_0p1_instance_iQCQP.eps} }
%
%
%
%
%
\vspace{-2mm}
\captionsetup{width=0.95\linewidth}
\caption[Short Caption]{Denoised instances under the Uniform noise model, for \textbf{OLS}, \textbf{QCQP} and \textbf{iQCQP},  as we increase the noise level $\gamma$. We keep fixed the parameters $n=500$, $k=2$, $\lambda= 0.1$.
}
\label{fig:instances_f1_Bounded}
\end{figure*}

\begin{figure*}
\centering
\subcaptionbox[]{  $\sigma=0.05$, \textbf{OLS}
}[ 0.32\textwidth ]
{\includegraphics[width=0.32\textwidth] {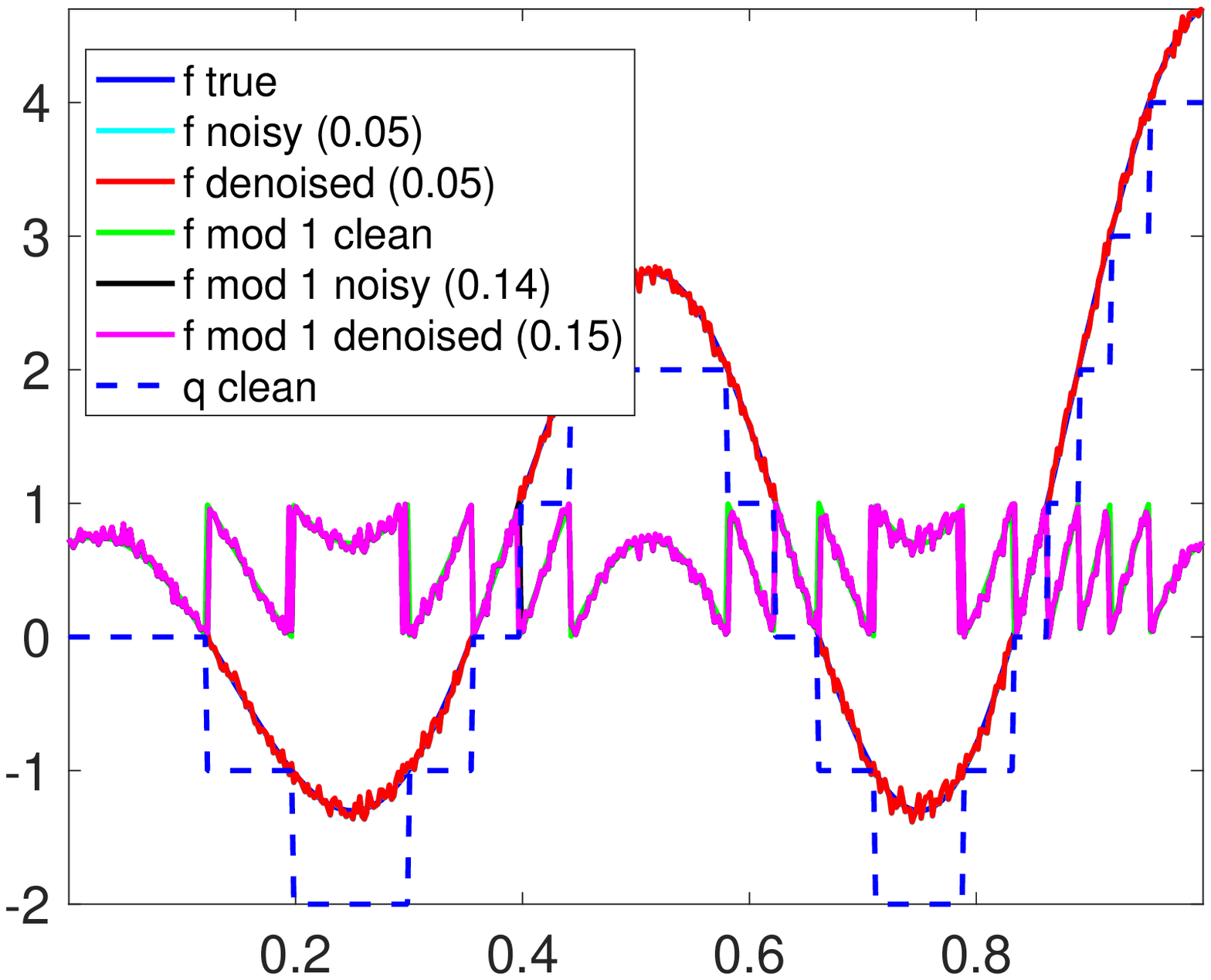} }
%
\subcaptionbox[]{  $\sigma=0.05$, \textbf{QCQP}
}[ 0.32\textwidth ]
{\includegraphics[width=0.32\textwidth] {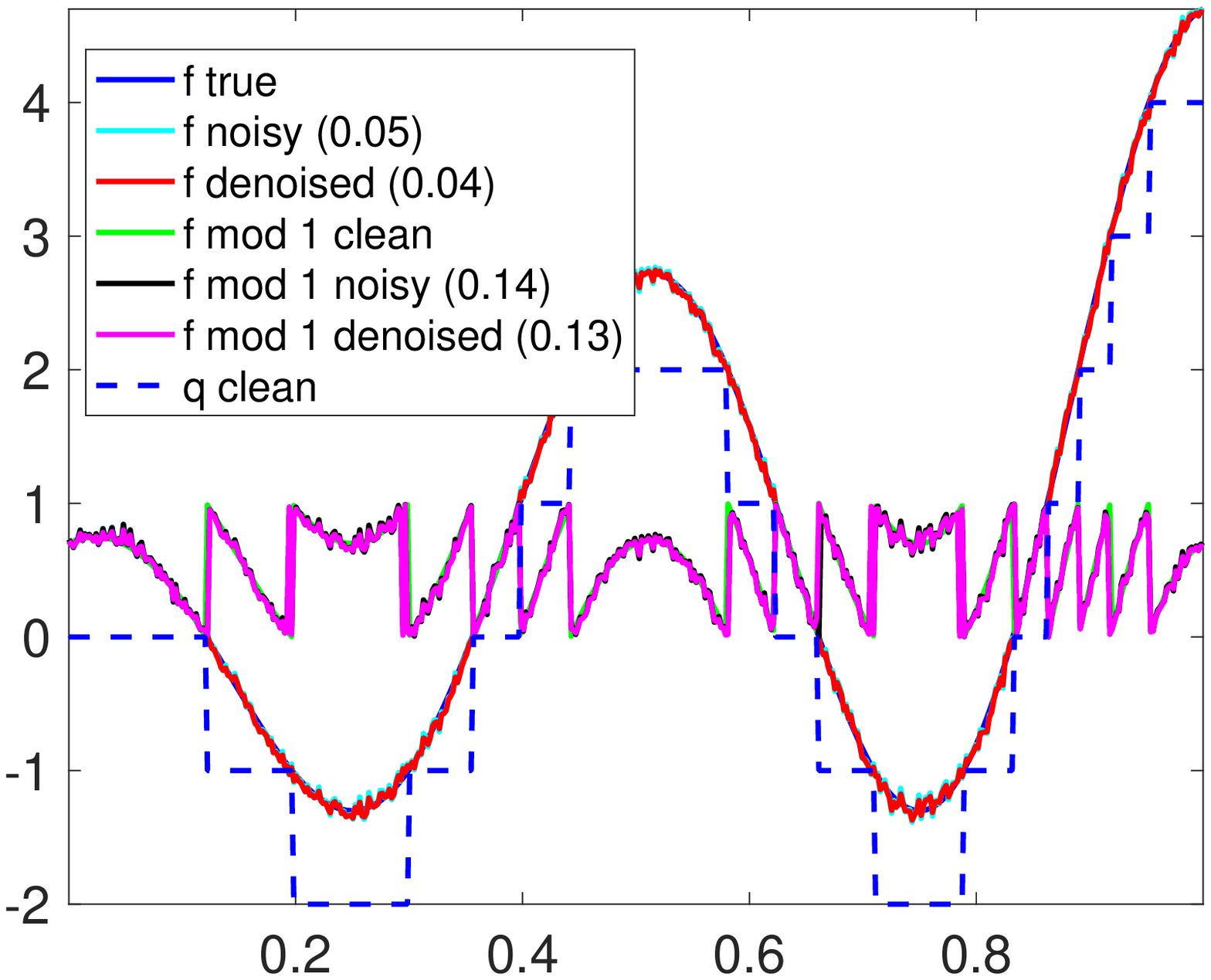} }
%
\subcaptionbox[]{  $\sigma=0.05$, \textbf{iQCQP}  (10  iters.)
}[ 0.32\textwidth ]
{\includegraphics[width=0.32\textwidth] {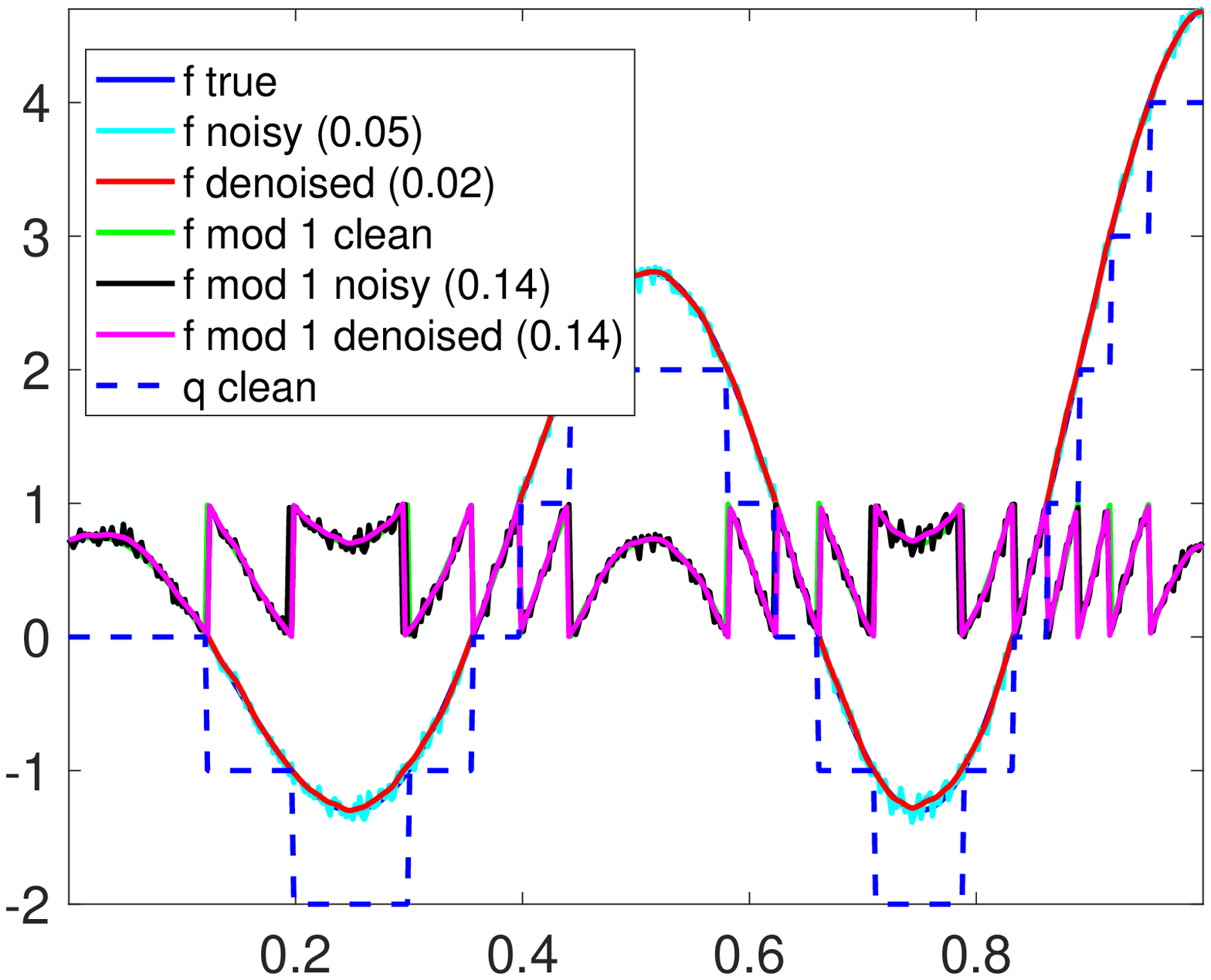} }
%
%
%

\subcaptionbox[]{  $\sigma=0.13$, \textbf{OLS}
}[ 0.32\textwidth ]
{\includegraphics[width=0.32\textwidth] {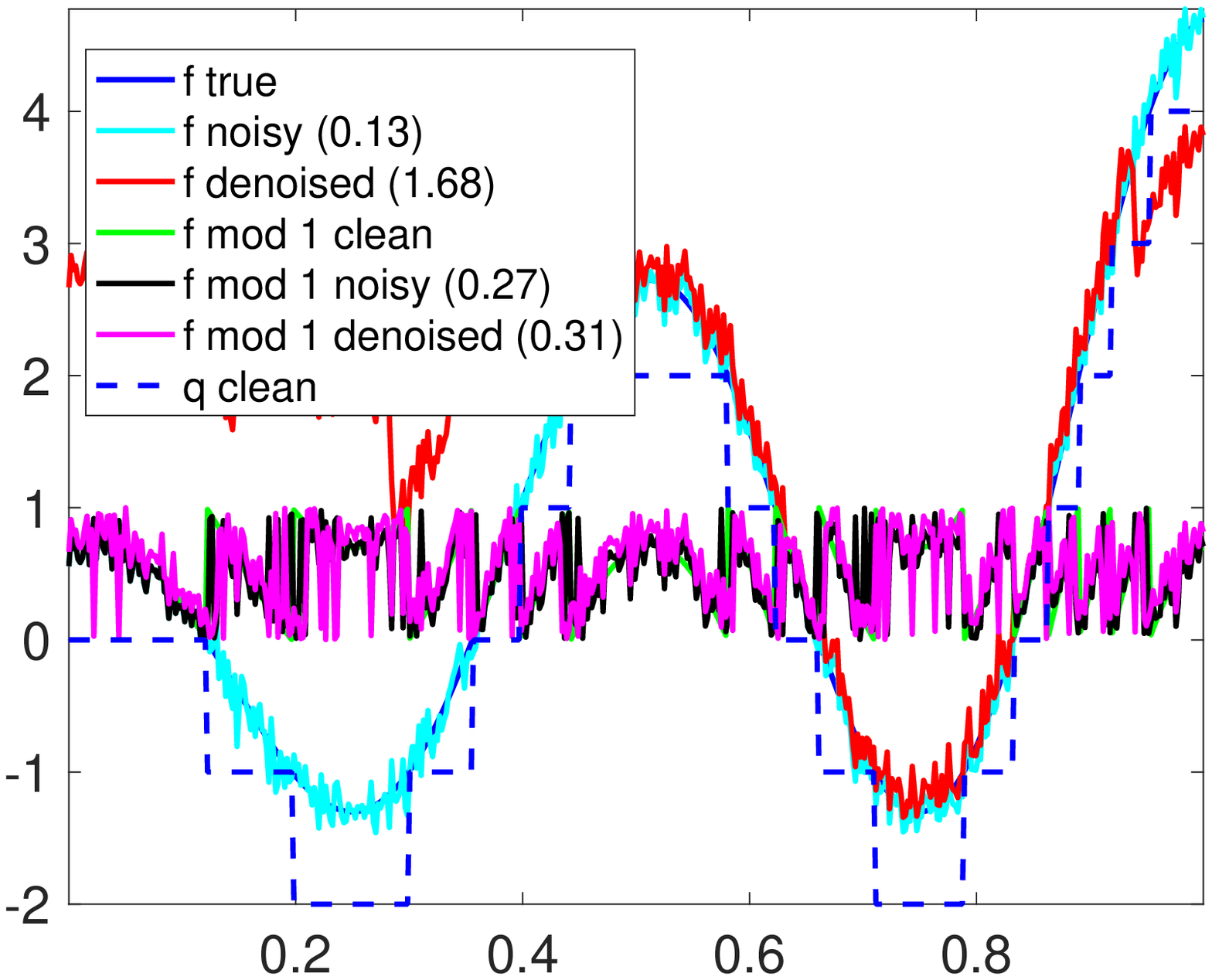} }
%
\subcaptionbox[]{  $\sigma=0.13$, \textbf{QCQP}
}[ 0.32\textwidth ]
{\includegraphics[width=0.32\textwidth] {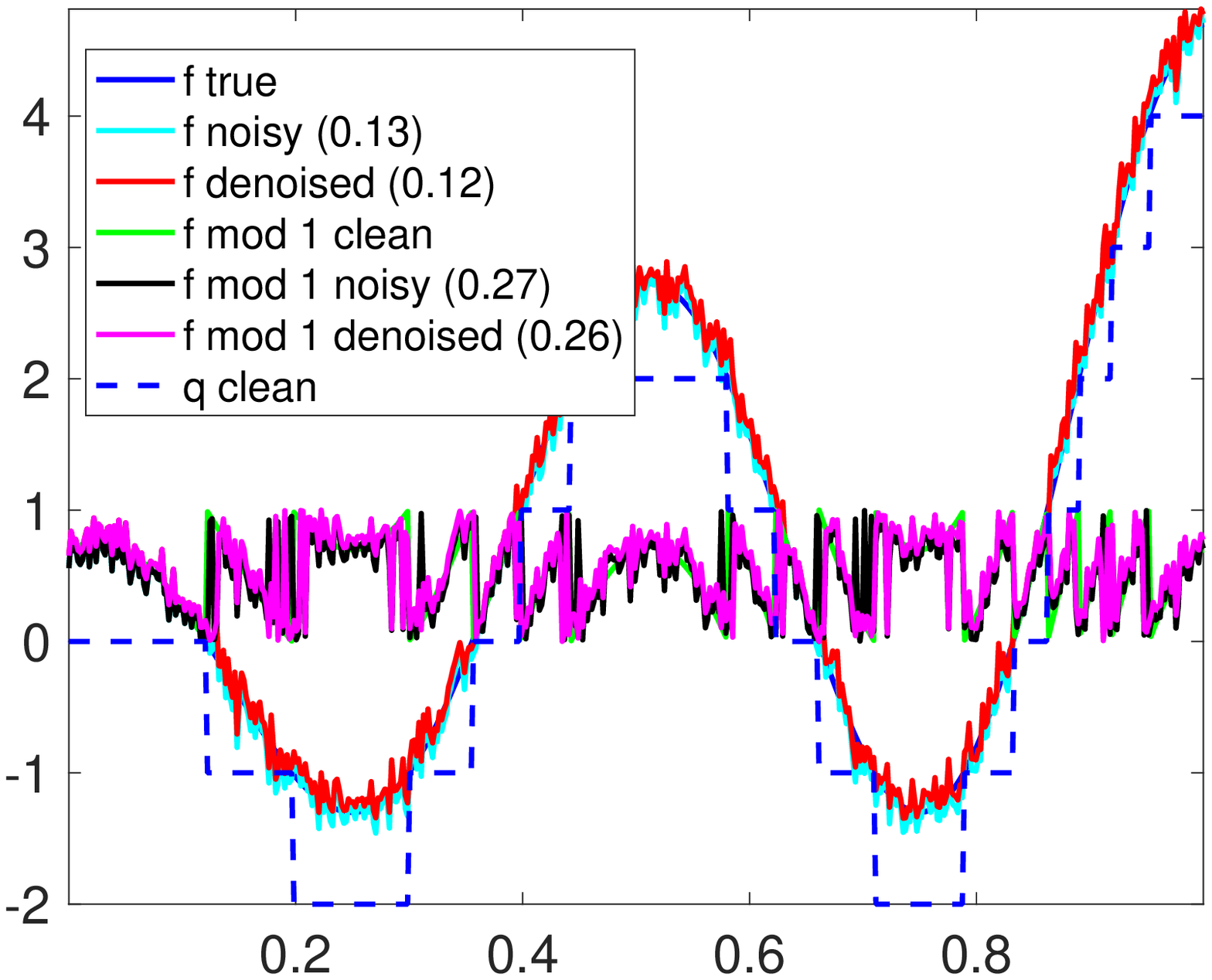} }
%
\subcaptionbox[]{  $\sigma=0.13$, \textbf{iQCQP} (10 iters.)
}[ 0.32\textwidth ]
{\includegraphics[width=0.32\textwidth] {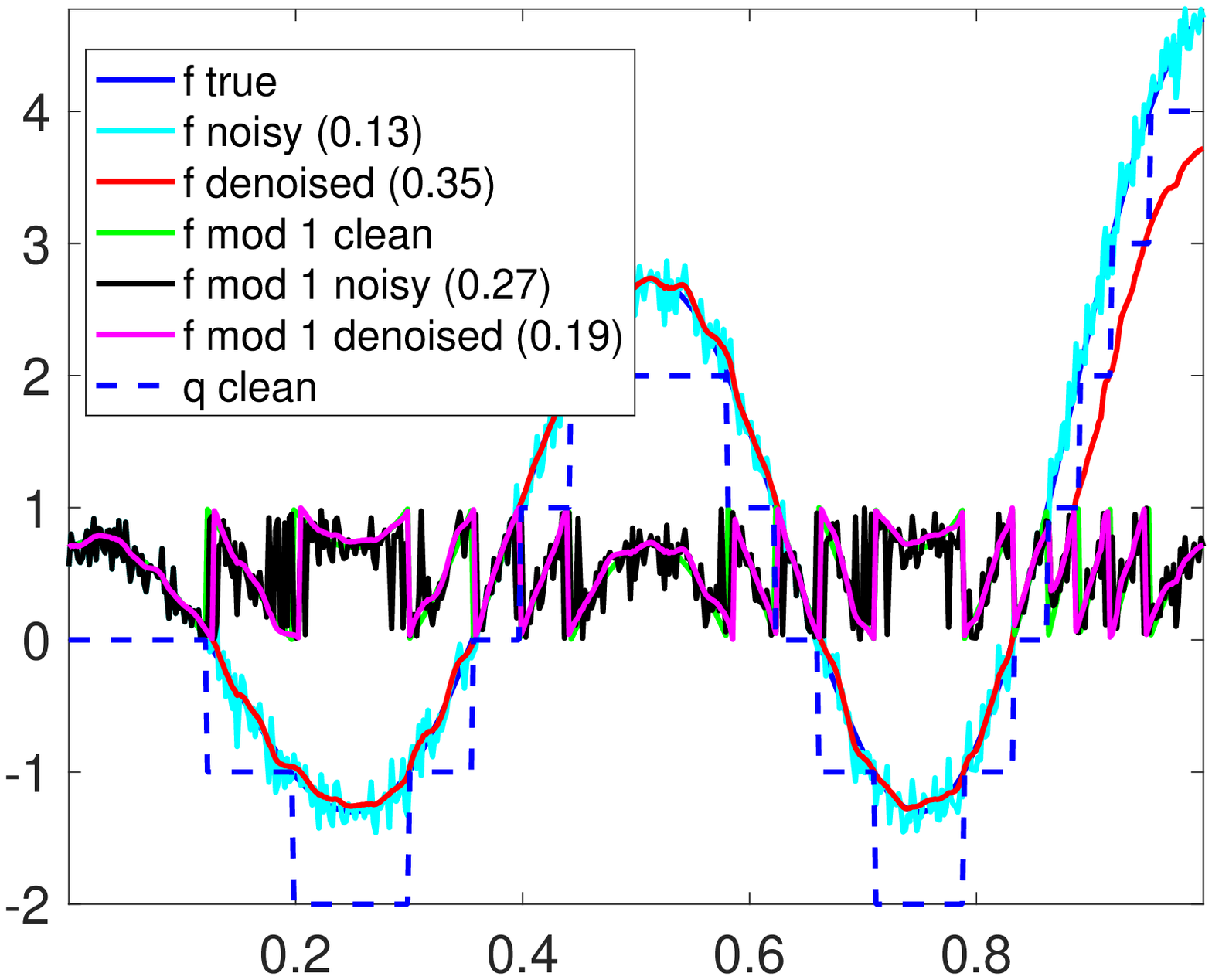} }
%
%
%
%
%

\subcaptionbox[]{  $\sigma=0.17$, \textbf{OLS}
}[ 0.32\textwidth ]
{\includegraphics[width=0.32\textwidth] {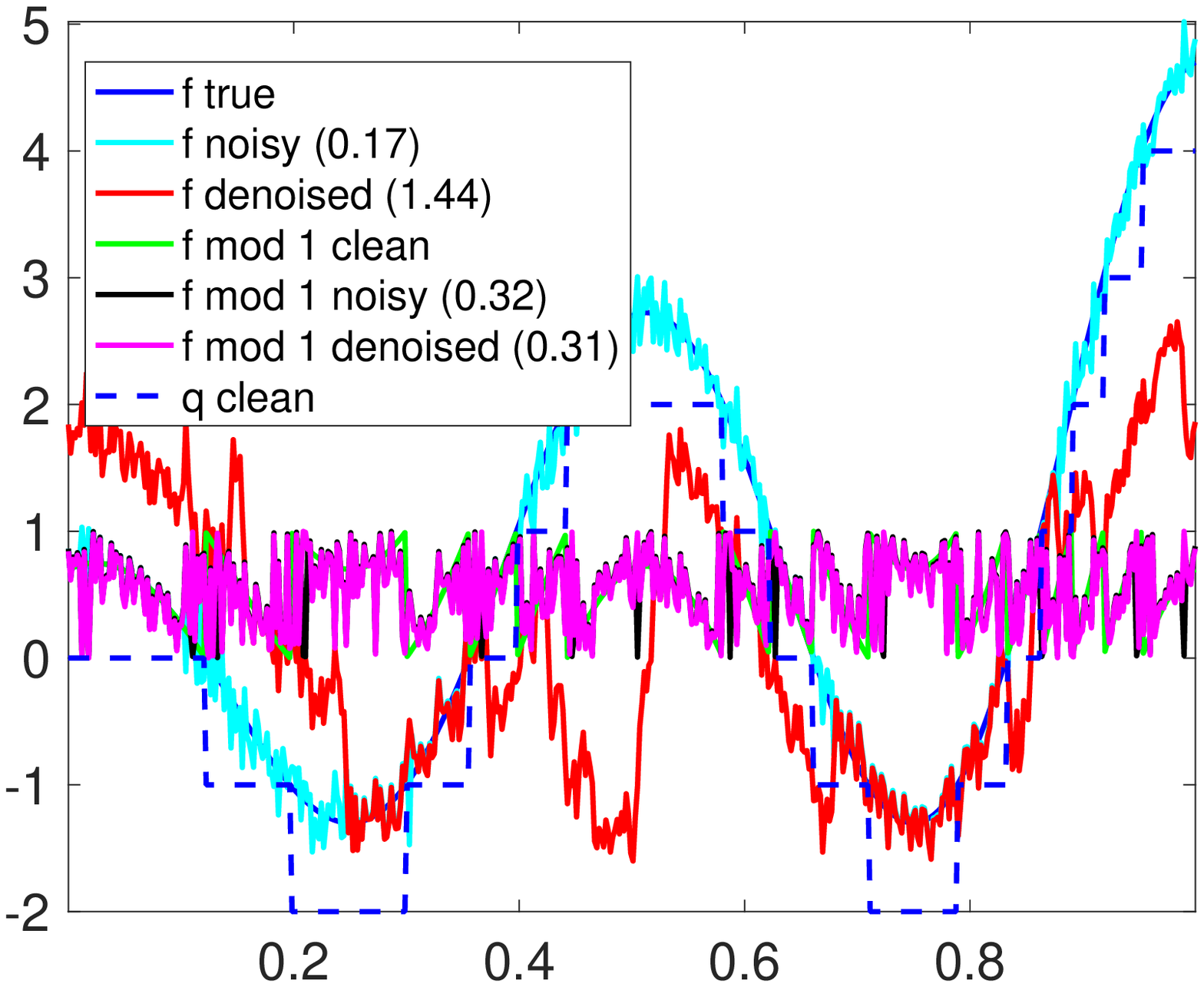} }
%
\subcaptionbox[]{  $\sigma=0.17$, \textbf{QCQP}
}[ 0.32\textwidth ]
{\includegraphics[width=0.32\textwidth] {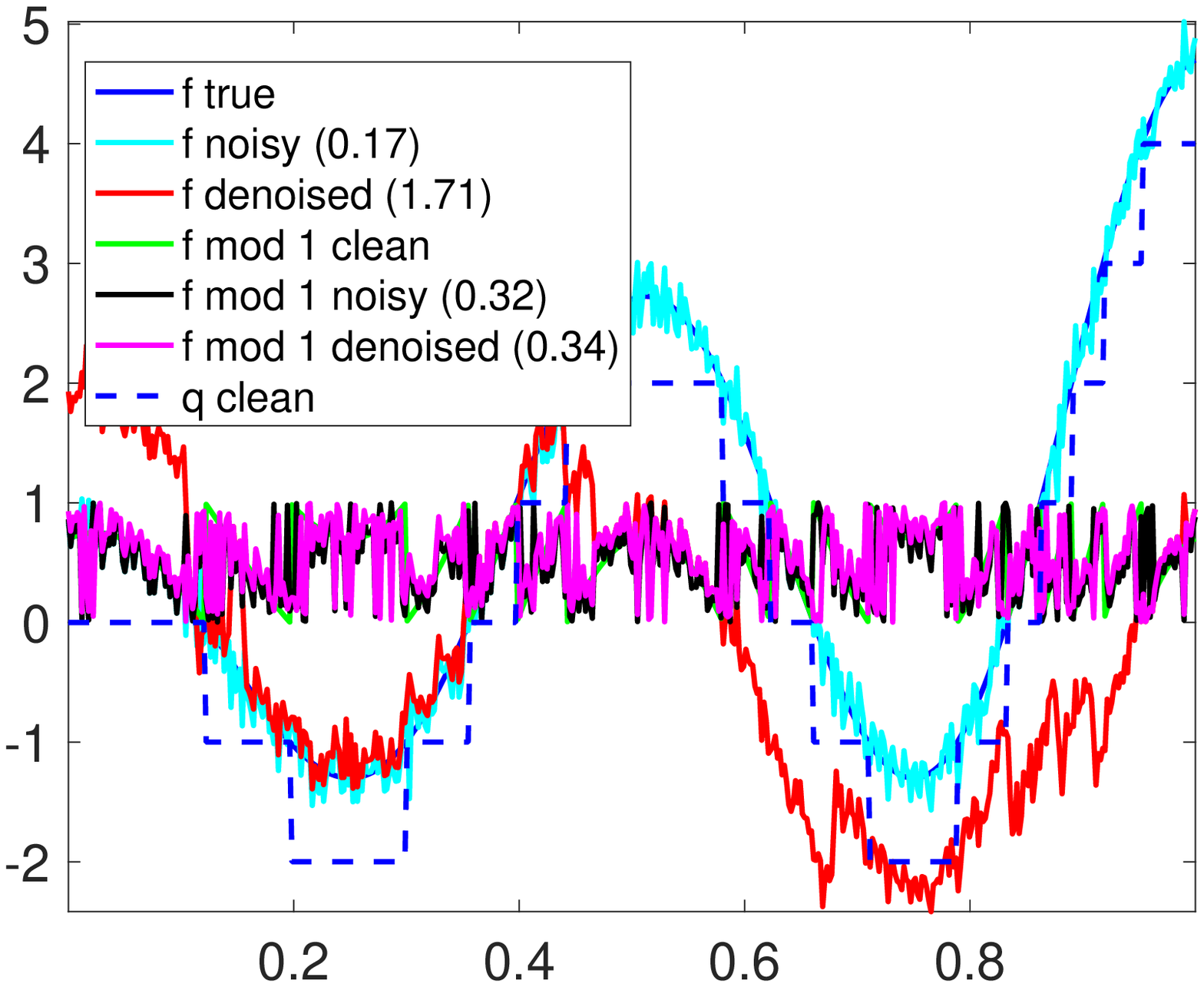} }
%
\subcaptionbox[]{  $\sigma=0.17$, \textbf{iQCQP}(10 iters.)
}[ 0.32\textwidth ]
{\includegraphics[width=0.32\textwidth] {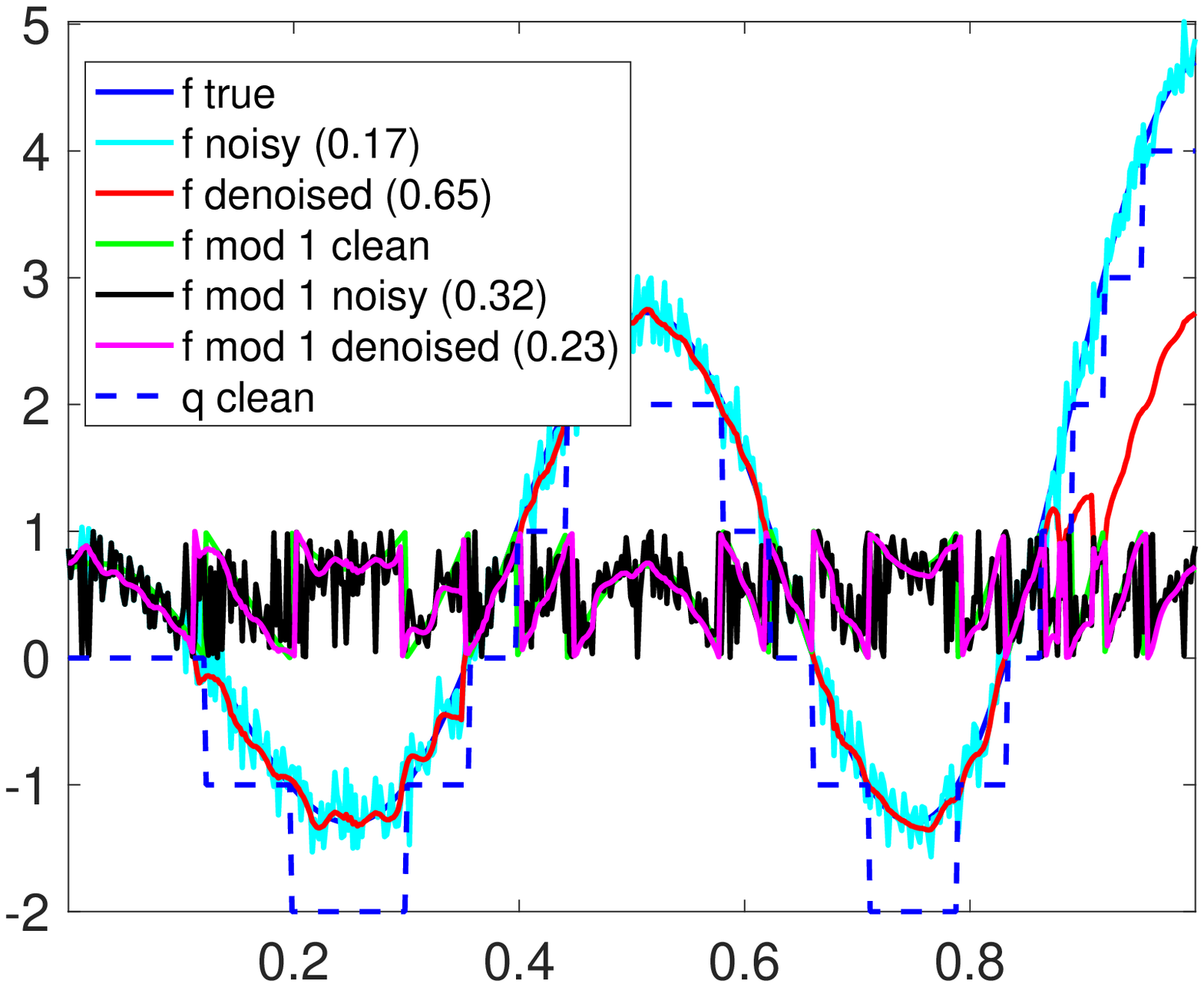} }
%
%
%
\vspace{-2mm}
\captionsetup{width=0.95\linewidth}
\caption[Short Caption]{Denoised instances under the Gaussian noise model, for \textbf{OLS}, \textbf{QCQP} and \textbf{iQCQP},  as we increase the noise level $\sigma$. We keep fixed the parameters $n=500$, $k=2$, $\lambda= 0.1$.
}
\label{fig:instances_f1_Gaussian}
\end{figure*}

Finally, Figure \ref{fig:instances_f1_Bounded}, respective Figure \ref{fig:instances_f1_Gaussian}, present instances of the recovery process under the Uniform noise model, respectively the Gaussian noise model, highlighting the noise level at which each method shows a significant decrease in performance. Our proposed iterated version  \textbf{iQCQP} shows surprising performance even at very high levels of noise, $\gamma=0.30$ in the Uniform noise model, and $\sigma=0.17$ in the Gaussian noise model. 

\section{Comparison with Bhandari et al. \cite{bhandari17}}

\begin{figure*}
\centering
\subcaptionbox[]{  $\gamma=0.13$, \textbf{BKR}
}[ 0.24\textwidth ]
{\includegraphics[width=0.24\textwidth] {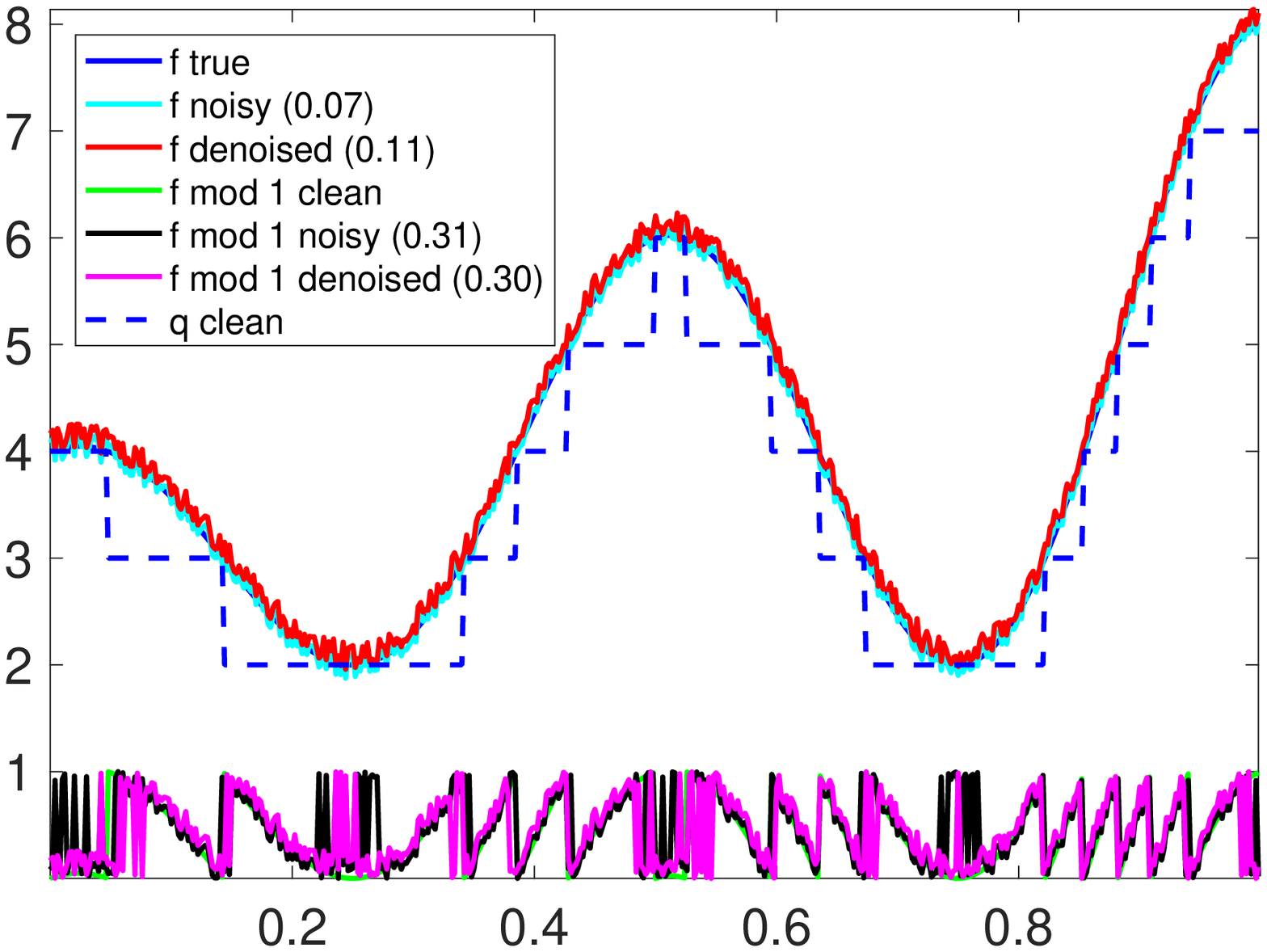} }
%
\subcaptionbox[]{  $\gamma=0.13$, \textbf{OLS}
}[ 0.24\textwidth ]
{\includegraphics[width=0.24\textwidth] {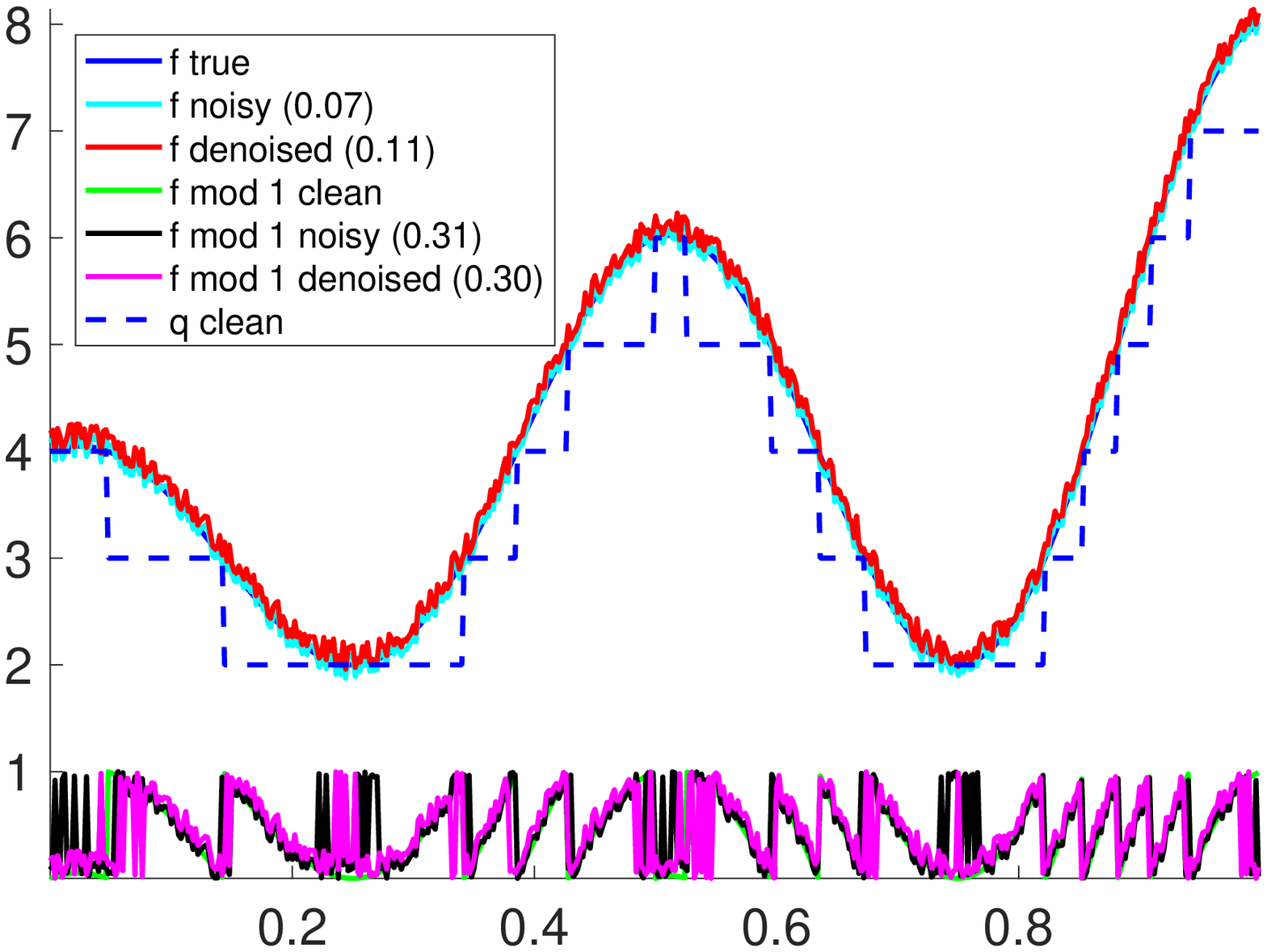} }
%
\subcaptionbox[]{  $\gamma=0.13$, \textbf{QCQP}
}[ 0.24\textwidth ]
{\includegraphics[width=0.24\textwidth] {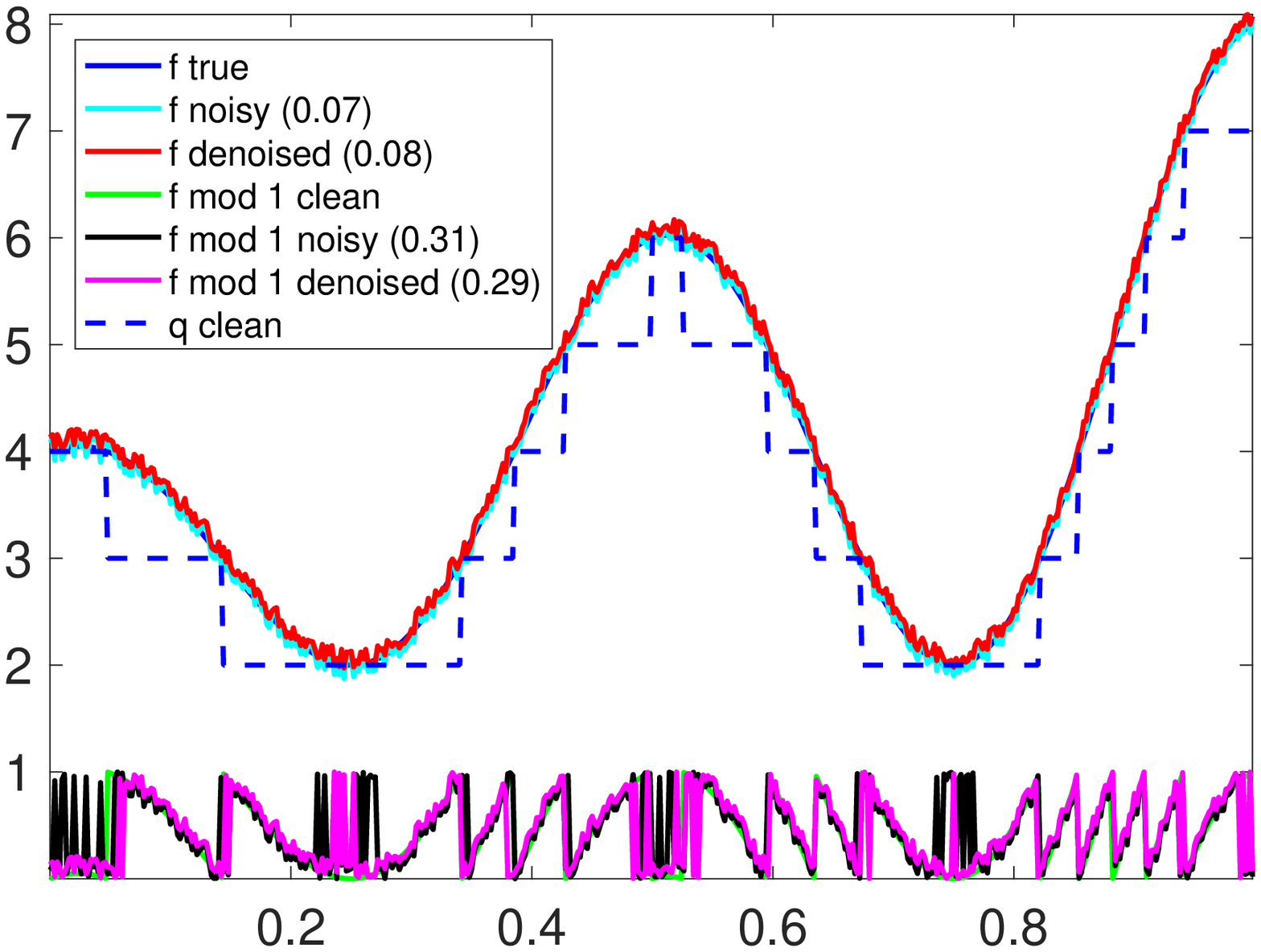} }
%
\subcaptionbox[]{  $\gamma=0.13$, \textbf{iQCQP}
}[ 0.24\textwidth ]
{\includegraphics[width=0.24\textwidth] {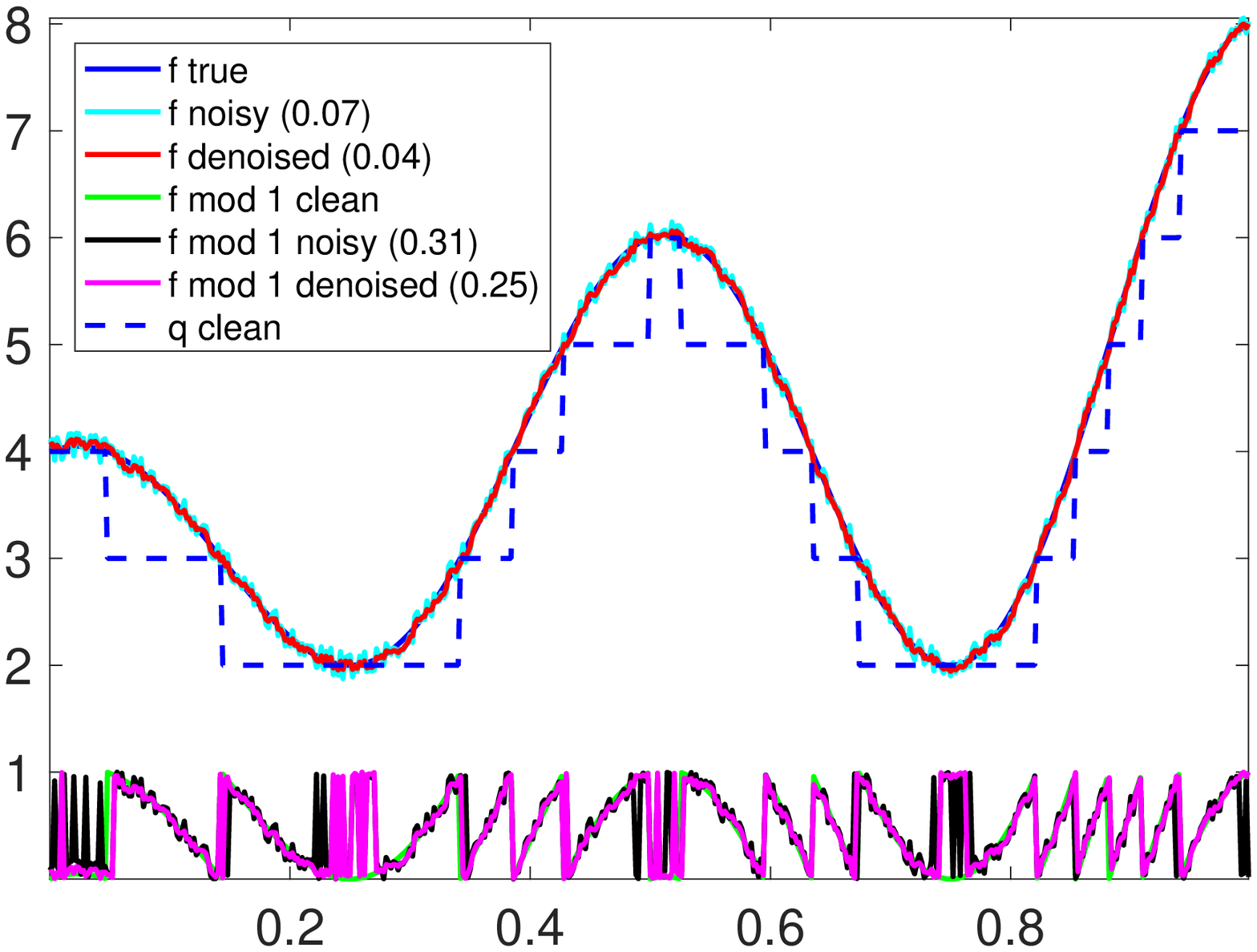} }
%
%
%
\subcaptionbox[]{  $\gamma=0.14$, \textbf{BKR}
}[ 0.24\textwidth ]
{\includegraphics[width=0.24\textwidth] {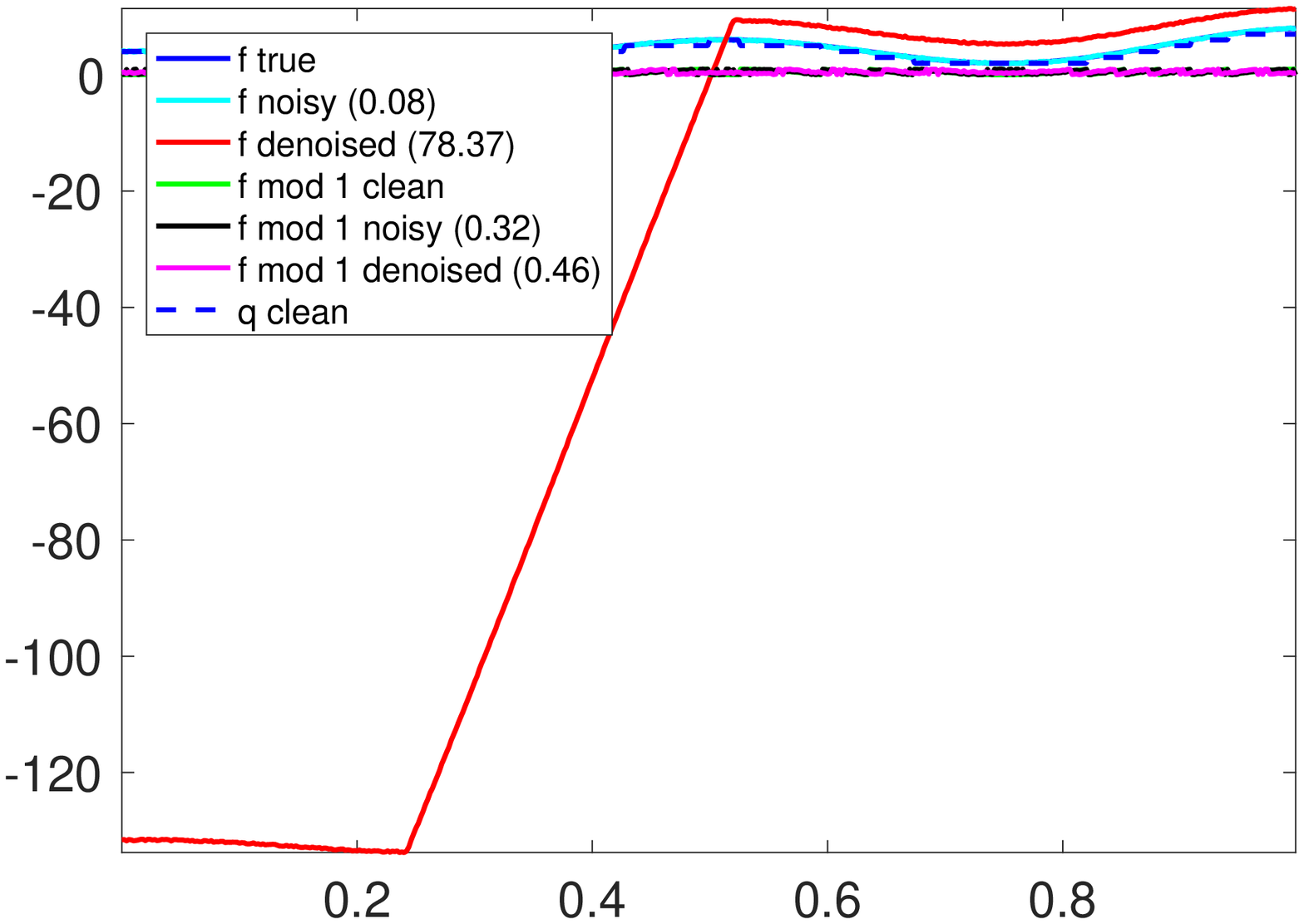} }
%
\subcaptionbox[]{  $\gamma=0.14$, \textbf{OLS}
}[ 0.24\textwidth ]
{\includegraphics[width=0.24\textwidth] {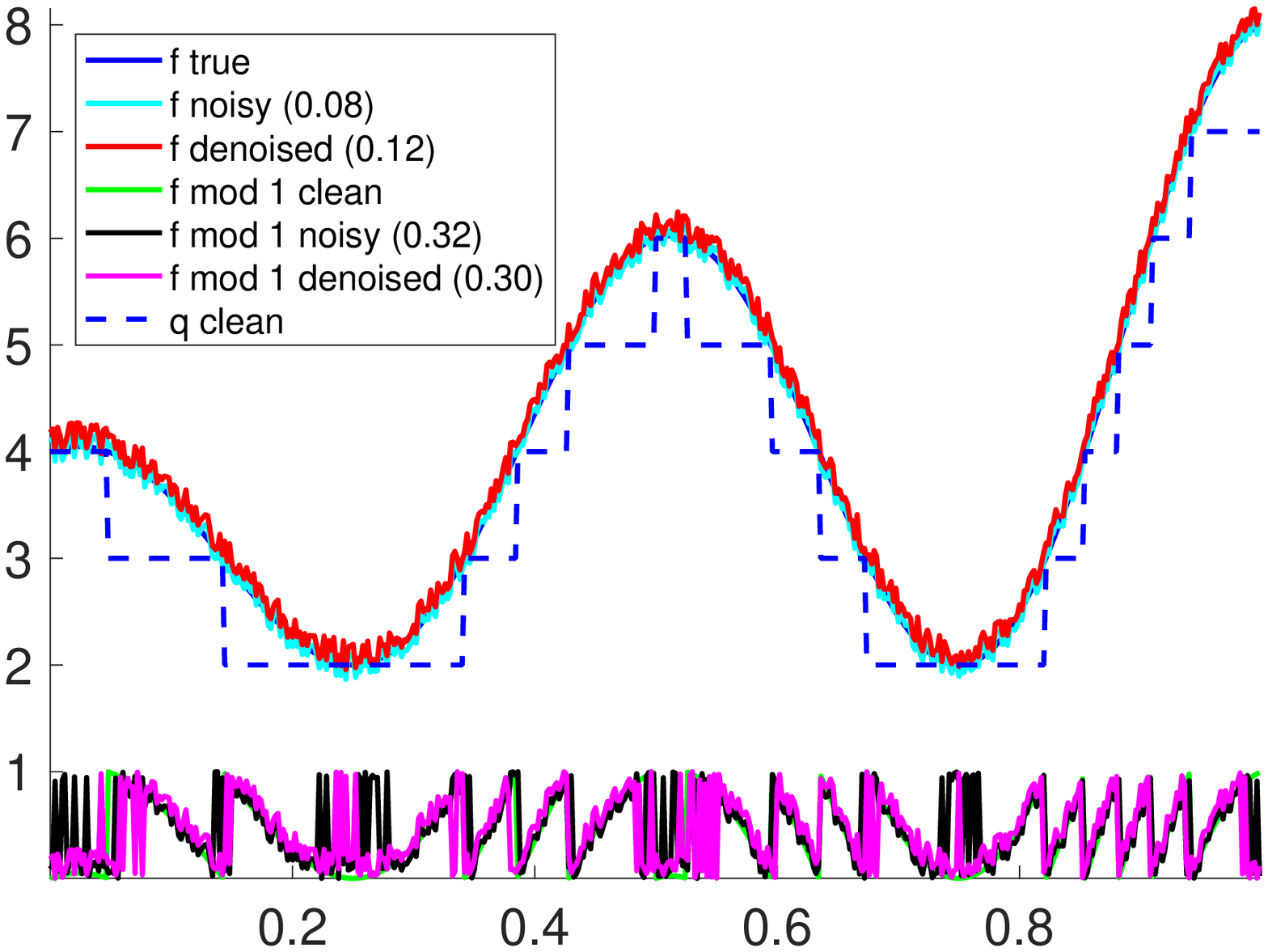} }
%
\subcaptionbox[]{  $\gamma=0.14$, \textbf{QCQP}
}[ 0.24\textwidth ]
{\includegraphics[width=0.24\textwidth] {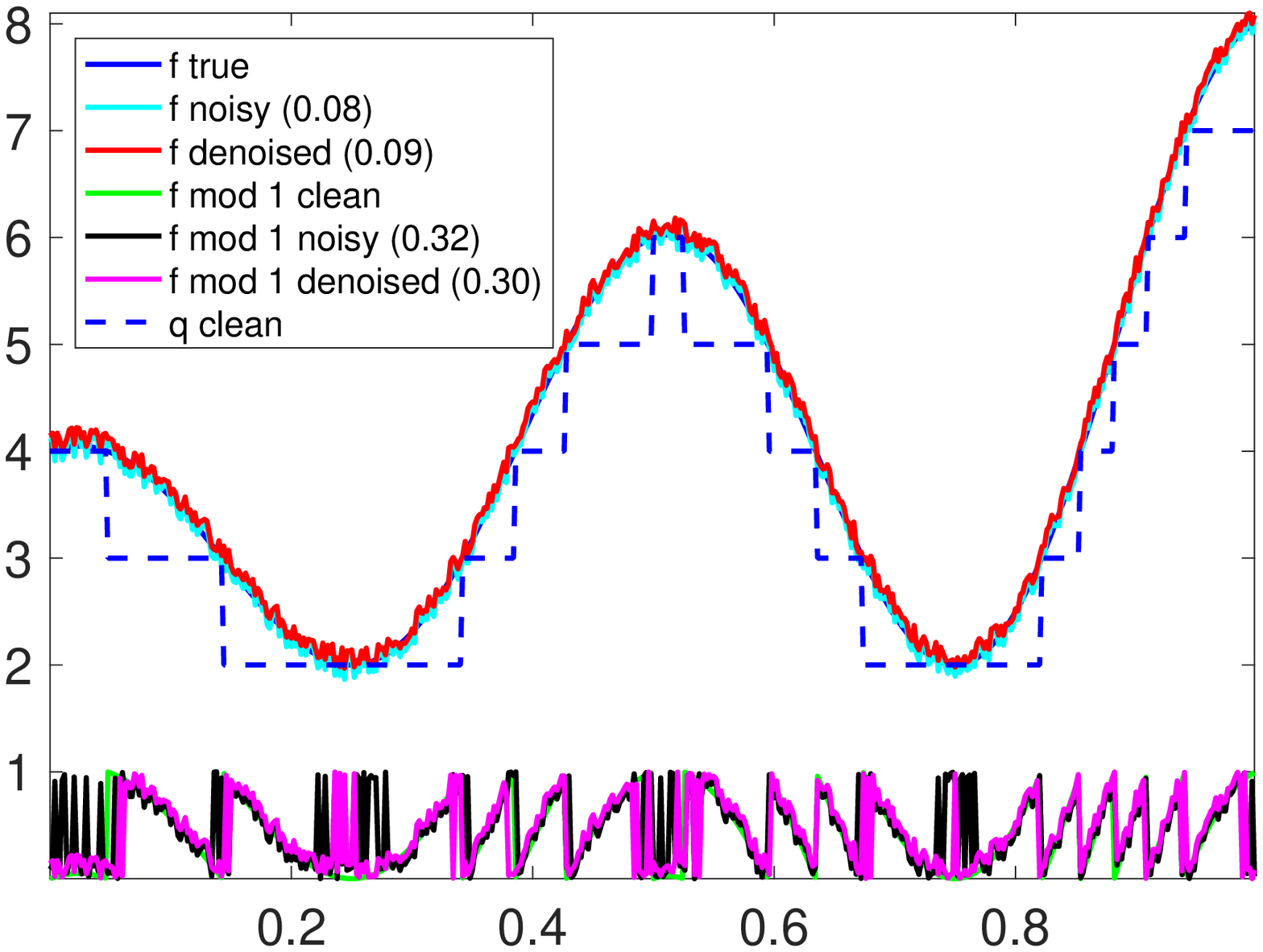} }
%
\subcaptionbox[]{  $\gamma=0.14$, \textbf{iQCQP}
}[ 0.24\textwidth ]
{\includegraphics[width=0.24\textwidth] {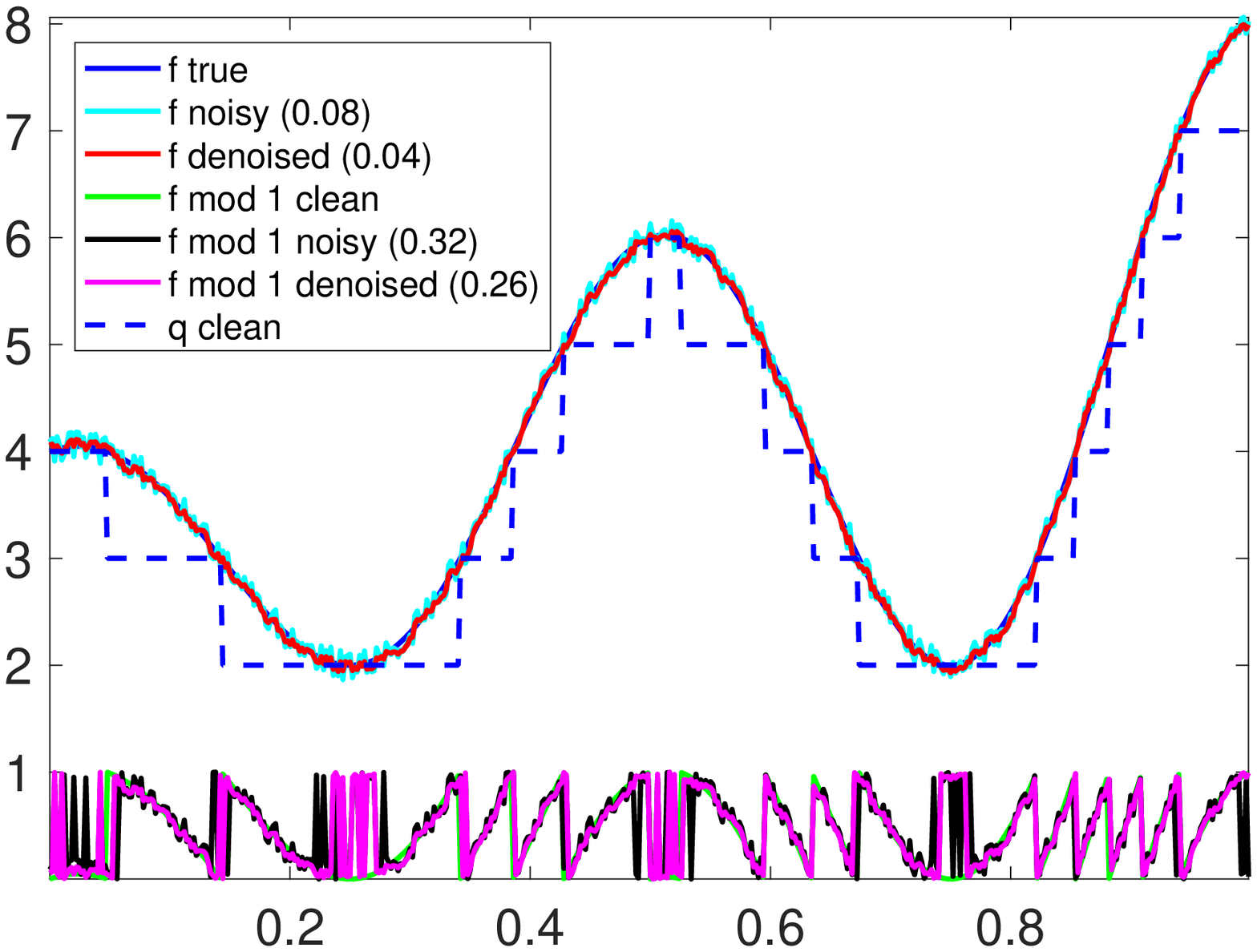} }
%
%
%
%
%
%
\subcaptionbox[]{  $\gamma=0.27$, \textbf{BKR}
}[ 0.24\textwidth ]
{\includegraphics[width=0.24\textwidth] {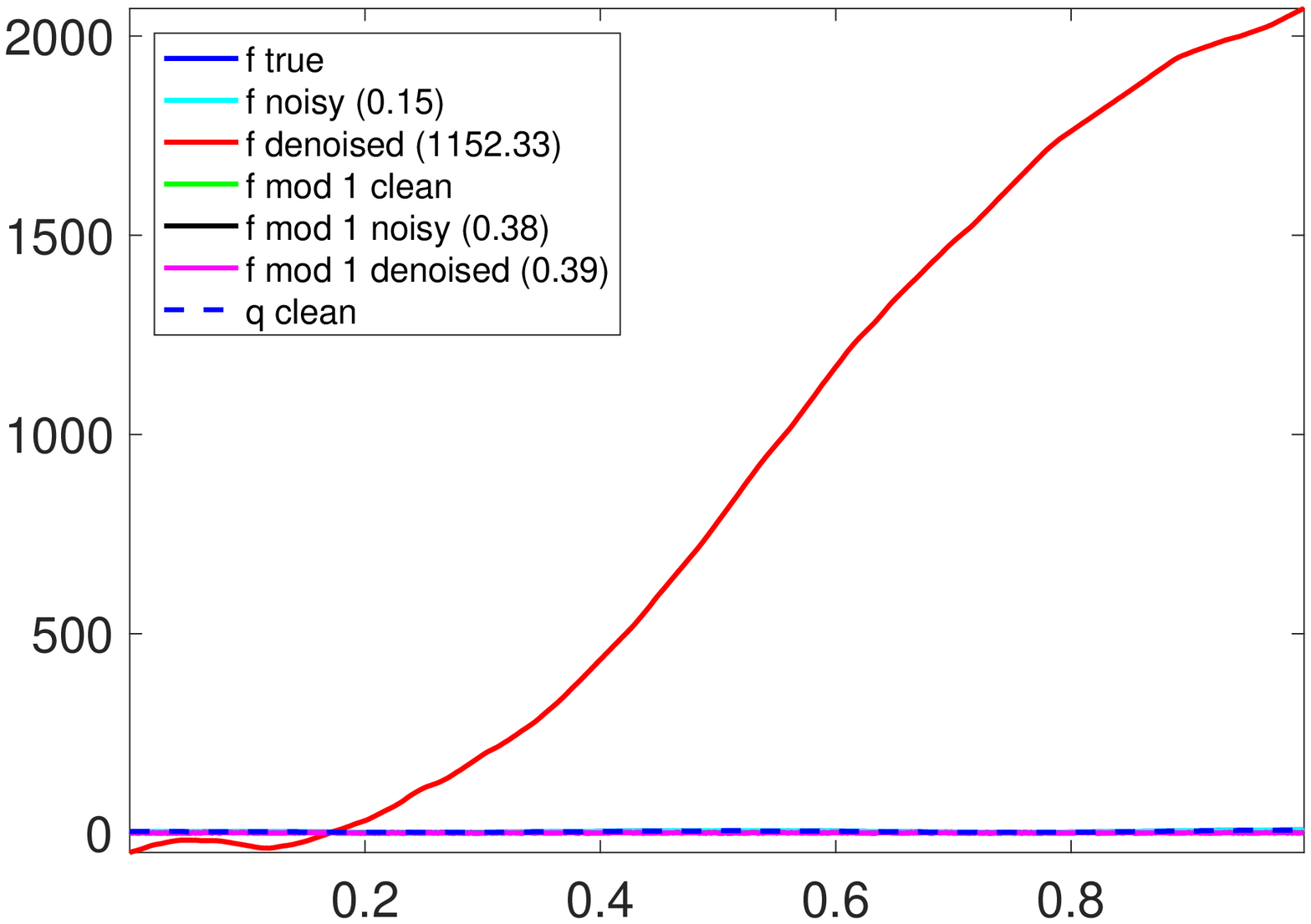} }
%
\subcaptionbox[]{  $\gamma=0.27$, \textbf{OLS}
}[ 0.24\textwidth ]
{\includegraphics[width=0.24\textwidth] {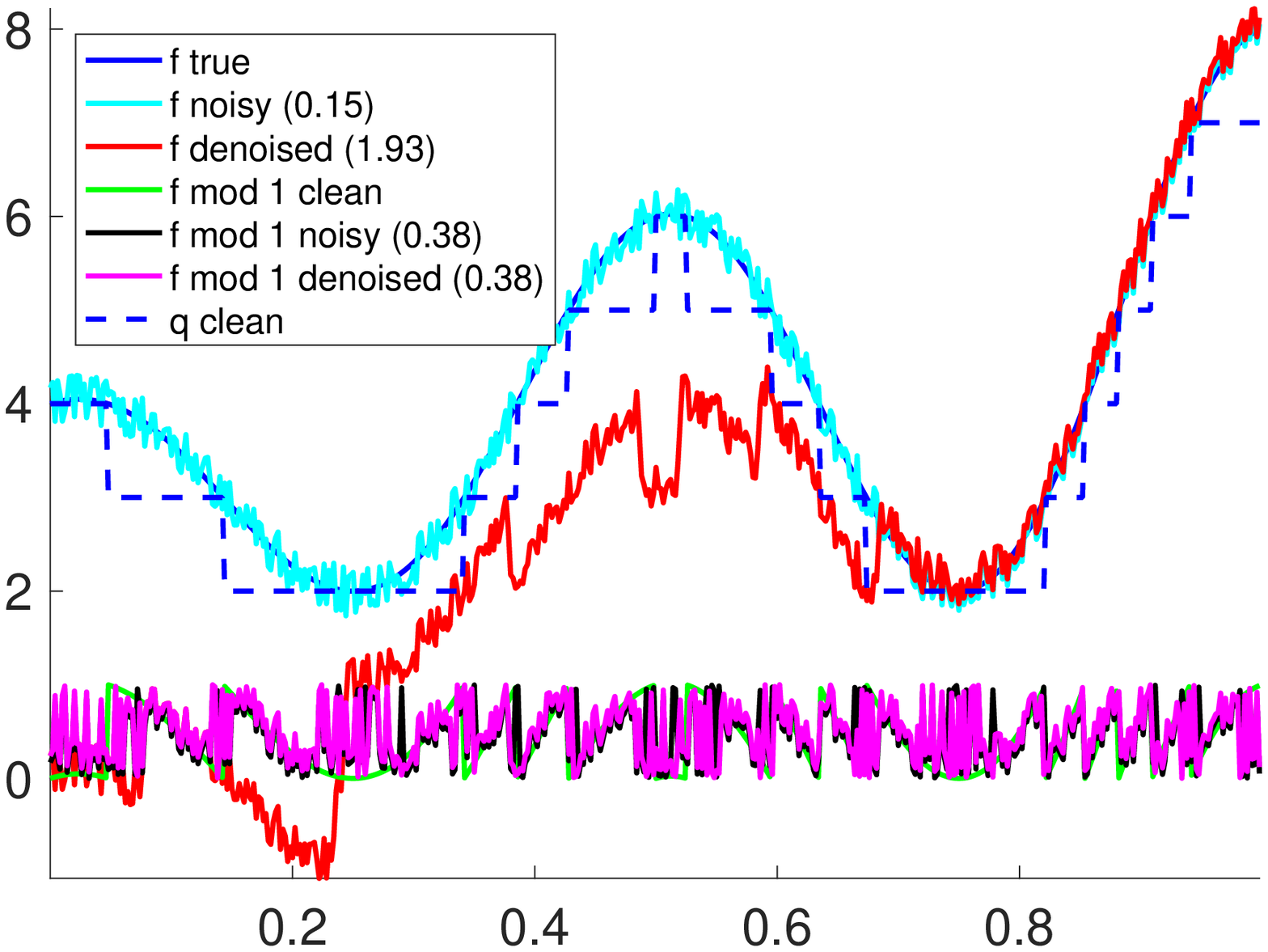} }
%
\subcaptionbox[]{  $\gamma=0.27$, \textbf{QCQP}
}[ 0.24\textwidth ]
{\includegraphics[width=0.24\textwidth] {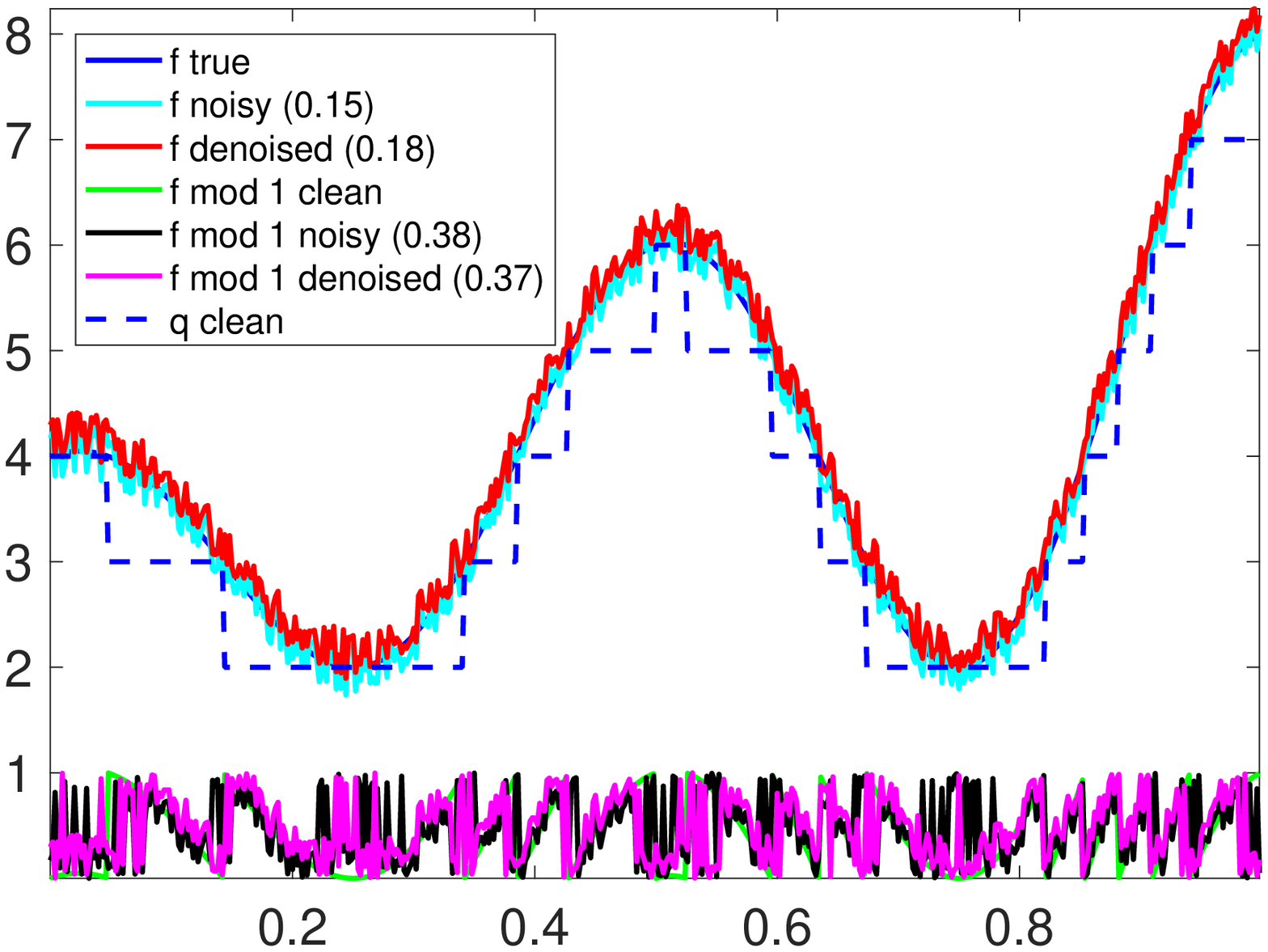} }
%
\subcaptionbox[]{  $\gamma=0.27$, \textbf{iQCQP}
}[ 0.24\textwidth ]
{\includegraphics[width=0.24\textwidth] {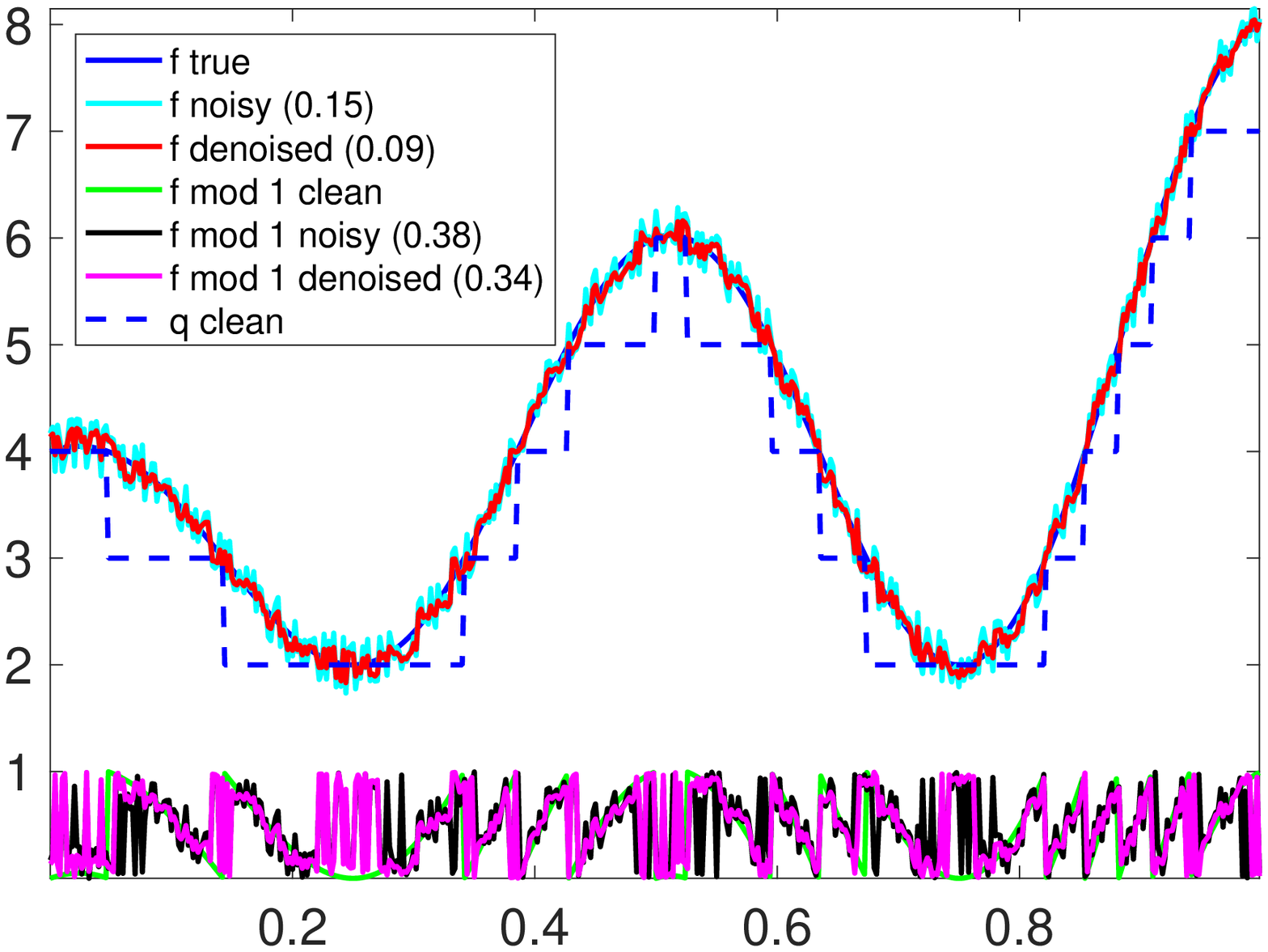} }
%
%
\vspace{-2mm}
\captionsetup{width=0.98\linewidth}
\caption[Short Caption]{Denoised instances for both the $f$ mod 1 and $f$ values, under the Bounded noise model, for  \textbf{BKR},  \textbf{OLS}, \textbf{QCQP} and \textbf{iQCQP},  as we increase the noise level $\gamma$. We keep fixed the parameters $n=500$, $k=2$, $\lambda= 0.1$. The numerical values in the legend denote the RMSE.
}
\label{fig:instances_f1_Bounded_Sampta}
\end{figure*}

In Figure \ref{fig:instances_f1_Bounded_Sampta} we show a comparison of  \textbf{OLS},  \textbf{QCQP}, and \textbf{iQCQP} with the approach introduced by Bhandari et al. in \cite{bhandari17}, whose algorithm we denote by \textbf{BKR} for brevity. We consider the Bounded noise model, and remark that at a lower level of noise $\gamma=0.13$, all methods perform  similarly well, with relative performance in the following order: 
\textbf{iQCQP} (RMSE=0.25), 
\textbf{QCQP} (RMSE=0.29), 
\textbf{OLS} (RMSE=0.30),  
\textbf{BKR} (RMSE=0.30). 
However, at higher levels of noise,  \textbf{BKR} returns  meaningless results, while \textbf{QCQP}, and especially \textbf{iQCQP}, return more accurate results.

\end{document}